\documentclass[pdflatex,sn-mathphys-num]{sn-jnl}

\usepackage{graphicx}%
\usepackage{multirow}%
\usepackage{amsmath,amssymb,amsfonts}%
\usepackage{amsthm}%
\usepackage{mathrsfs}%
\usepackage[title]{appendix}%
\usepackage{xcolor}%
\usepackage{textcomp}%
\usepackage{manyfoot}%
\usepackage{booktabs}%
\usepackage{algorithm}%
\usepackage{algorithmicx}%
\usepackage{algpseudocode}%
\usepackage{listings}%

\usepackage[dvipsnames, svgnames,table]{xcolor}
\usepackage{url}
\usepackage{subcaption}
\usepackage{multirow}
\usepackage{adjustbox}
\usepackage{float}
\usepackage{tabularx,ragged2e,booktabs}
\usepackage{pifont}
\usepackage{makecell}
\usepackage{geometry}
\theoremstyle{thmstyleone}%
\newtheorem{theorem}{Theorem}
\newtheorem{proposition}[theorem]{Proposition}%

\theoremstyle{thmstyletwo}%
\newtheorem{example}{Example}%
\newtheorem{remark}{Remark}%

\theoremstyle{thmstylethree}%
\newtheorem{definition}{Definition}%

\usepackage{scalerel}
\usepackage{tikz}
\usetikzlibrary{svg.path}

\definecolor{orcidlogocol}{HTML}{A6CE39}
\tikzset{
  orcidlogo/.pic={
    \fill[orcidlogocol] svg{M256,128c0,70.7-57.3,128-128,128C57.3,256,0,198.7,0,128C0,57.3,57.3,0,128,0C198.7,0,256,57.3,256,128z};
    \fill[white] svg{M86.3,186.2H70.9V79.1h15.4v48.4V186.2z}
                 svg{M108.9,79.1h41.6c39.6,0,57,28.3,57,53.6c0,27.5-21.5,53.6-56.8,53.6h-41.8V79.1z M124.3,172.4h24.5c34.9,0,42.9-26.5,42.9-39.7c0-21.5-13.7-39.7-43.7-39.7h-23.7V172.4z}
                 svg{M88.7,56.8c0,5.5-4.5,10.1-10.1,10.1c-5.6,0-10.1-4.6-10.1-10.1c0-5.6,4.5-10.1,10.1-10.1C84.2,46.7,88.7,51.3,88.7,56.8z};
  }
}

\definecolor{darkgreen}{RGB}{0,150,0}
\definecolor{darkred}{RGB}{220,0,0}

\newcommand{\xmark}{\textcolor{black}{\ding{55}}} 

\begin{document}

\title[Article Title]{Generative AI-Based Data Augmentation for Oral Lesion Classification: The PhotoMOCI Dataset and Benchmark}


\author*[1]{\fnm{Marco} \sur{Parola}}\email{mpar@create.aau.dk}

\author[2]{\fnm{Mario G.C.A.} \sur{Cimino}}\email{mario.cimino@unipi.it}

\author[3]{\fnm{Sabrina} \sur{Senatore}}\email{ssenatore@unisa.it}

\affil*[1]{\orgname{Aalborg University}, \country{Denmark}}

\affil[2]{\orgname{University of Pisa}, \country{Italy}}

\affil[3]{ \orgname{University of Salerno}, \country{Italy}}


\abstract{
Early detection of oral cancer via photographic imaging presents a promising avenue for large-scale oral cavity screening. However, the development of robust deep learning models is frequently hampered by the scarcity of high-quality, annotated datasets. 
To address this limitation, a novel and well-curated resource, the Photographic Multi-purpose Oral Cancer Imaging (PhotoMOCI) dataset, is introduced for developing models across multiple diagnostic tasks in oral oncology.
Then, a comprehensive benchmark study was conducted to investigate how various data augmentation strategies influence the performance of image classifiers.
Our analysis spans different generative AI frameworks, evaluating the efficacy of traditional methods against advanced generative approaches, including Generative Adversarial Networks (GANs) and Diffusion Models (DMs).
Additionally, we propose the Synthetic Image Filter (SIF), a mechanism to select specific samples based on two auxiliary models: Synthetic Proxy Classifier to ensure samples are representative of the target class and Synthetic Image Detector to verify they appear realistic, thereby selecting only the high-utility images that contribute to improving downstream performance.
Across the evaluated datasets and classifiers, the best SIF-filtered setup improves accuracy over traditional augmentation in all cases, with gains of +1.73\% and +2.35\% on PhotoMOCI and +2.38\% and +2.08\% on KOCD for ResNet50 and ViT, respectively.
Our findings reveal that while the direct application of generative data augmentation may yield performance drops, the integration of SIF, considering (i) how synthetic data looks real and (ii) how it reflects the discriminative features of the belonging class, provides a simple yet effective mechanism to filter out synthetic samples that confuse the classifier during training.
The PhotoMOCI dataset is publicly available on Kaggle at \cite{marco_parola_2025}.
}




\maketitle

\section{Introduction}\label{sec:intro}


Oral cancer (OC) remains a common and aggressive malignancy of the oral cavity, posing a substantial global burden of morbidity and mortality \cite{kim2020increasing}. This underscores the need for improved early recognition strategies \cite{kim2020increasing, di2024automated}, since, despite advances in treatment such as surgical resection and chemotherapy, the five-year survival rate remains around 60\% \cite{zhou2022ru}. Consequently, photographic screening may support the early recognition and referral of suspicious oral lesions, while definitive diagnosis remains based on clinical assessment and, when indicated, histopathological examination \cite{bramati2021early, ribeiro2022assessment}.
An effective screening must balance accessibility, speed, and cost-efficiency\cite{ribeiro2022assessment}. In this regard, recent research has explored the integration of Deep Learning (DL) with photographic imaging as a decision-support tool for oral lesion recognition and screening, rather than as a replacement for clinical or histopathological diagnosis \cite{CIMINO2025100538, PAROLA2024102433}.

Since DL has shown considerable results in many medical image analyses, coupled with photographic imaging, it offers the potential to support clinical decision-making by providing diagnostic insights while reducing the cost related to other more expensive medical imaging techniques, such as X-ray, CT-scan, etc. \cite{rajpurkar2022ai, lee2021survey}. The resulting performance therefore should not be interpreted as evidence of autonomous oral cancer diagnosis or as a substitute for the corresponding clinical and histopathological reference assessment.
However, the development of robust DL-based systems is significantly constrained by the lack of large-scale, high-quality, and well-annotated datasets that are essential for training DL architectures \cite{alzubaiDataScarsity, whang2022datacollectionqualitychallenges}. 
Indeed, as highlighted in Section \ref{sec:photomoci}, oral cancer datasets for medical imaging are often limited in size due to challenges such as patient privacy concerns, the high cost and complexity of expert annotation, and the relatively low prevalence of early-stage lesions in clinical populations \cite{alzubaiDataScarsity, whang2022datacollectionqualitychallenges, 
 di2024automated}.

In this context, synthetic data generation emerges as a viable solution. By producing realistic images that increase visual variability within the photographic lesion domain, synthetic data can potentially augment existing datasets and improve the performance of DL models trained on them \cite{tang2023deep, alzubaiDataScarsity}.
The effectiveness of synthetic image generation as a data augmentation technique varies across different medical domains. While it has shown clear benefits in small-data scenarios, where real data are scarce, and model generalization is limited, it is not always consistently effective in broader or more data-rich applications \cite{tortora2025gan}. In contrast, traditional transformation-based augmentation techniques often prove more effective and robust in large-data settings \cite{tortora2025gan}, offering simpler and more computationally efficient improvements without the complexities of generative modeling. 

As discussed in the literature review (Section \ref{sec:data-augmentation-review}), synthetic data augmentation has been widely adopted in medical imaging and oral cancer. However, to the best of our knowledge, a systematic and comprehensive evaluation of synthetic data for photographic oral cancer imaging remains lacking. Existing studies typically focus on a single generative model, most often GAN-based approaches, or rely on a single dataset, which is frequently not publicly available, as highlighted in the dataset review in Section \ref{sec:photomoci}. Moreover, many contributions address alternative imaging modalities, such as histopathological images, limiting their applicability to photographic screening.

To address both the limited availability of datasets and the absence of a comprehensive benchmark, we introduce a novel dataset designed to support future benchmarks and enable a more systematic evaluation of synthetic data generation for photographic oral cancer imaging. 
Specifically, the contributions of this work are:
 
\begin{itemize}
    \item Introducing a novel dataset in this scenario of data scarcity, the \textbf{Photographic Multi-purpose Oral Cancer Imaging (PhotoMOCI)} dataset, a well-curated collection for oral cancer that enables the development of models for multiple tasks.

    \item Investigating the impact of different data augmentation strategies on image classification performance, and presenting a comprehensive benchmark across multiple datasets, image classifiers, generative models, and augmentation techniques.
    
    \item Proposing the \textbf{Synthetic Image Filter (SIF)}, a simple but effective strategy relying on two auxiliary classification problems to select specific synthetic images that contribute to improving the final classifier performance, based on how a synthetic image looks real and how it matches discriminative features of the belonging class.
    
\end{itemize}

\section{Related work}\label{sec:realtedwork}

DL has achieved strong performance in medical imaging, but adoption is often limited by small datasets caused by privacy constraints, high annotation costs, and the rarity of certain diseases, leading to class imbalance and poor case representation \cite{goodfellow2014generative, Zhang2023DLmedImages, Johnson2019}. Thus, in Section \ref{sec:data-augmentation-review}, we review traditional and synthetic data augmentation techniques that expand dataset size and diversity, including Generative Adversarial Networks and Diffusion Models, with a specific focus on conditioning mechanisms in Section \ref{sec:conditioning}. 

\subsection{Data augmentation technique review}\label{sec:data-augmentation-review}

\textbf{Traditional data augmentation}. Traditional data augmentation increases training variability through predefined image transformations \cite{krizhevsky2012imagenet}, commonly categorized as geometric or photometric \cite{tang2023deep, shorten2019survey}. Geometric transformations modify spatial structure (e.g., rotation, flipping, scaling), whereas photometric transformations alter pixel intensities (e.g., brightness, contrast) or add noise.
These techniques, widely used in computer vision, have also proven effective in medical imaging applications \cite{shin2016deep,ronneberger2015u,alosaimi2022efficient,oya2023diagnosis}.   

Due to their computational efficiency \cite{whang2022datacollectionqualitychallenges}, traditional augmentations are broadly adopted and generally improve performance compared to training without augmentation \cite{tang2023deep, krizhevsky2012imagenet, ronneberger2015u}. 
However, they cannot generate truly novel samples or pathological features, as they only modify existing images \cite{shorten2019survey}. This limits their effectiveness in scenarios with severe data scarcity, low diversity \cite{alzubaiDataScarsity}, or class imbalance \cite{Johnson2019, kazemzadeh2025addressing}, and they lack conditional generation capabilities. 
In contrast, DL-based generative approaches show greater promise in small-data medical imaging settings \cite{tortora2025gan}, while traditional augmentation remains reliable for improving generalization in larger datasets \cite{tortora2025gan}. 


\vspace{2mm}
\textbf{Generative Adversarial Networks}. 
Generative Adversarial Networks (GANs) \cite{goodfellow2014generative} represent the next generation of generative models after autoencoders (AEs) \cite{hinton2006reducing} and later Variational Autoencoders (VAEs)\cite{kingma2013auto}. They introduced an adversarial training paradigm involving a generator and a discriminator. Through this process, the generator progressively learns to synthesize visually realistic images that deceive the discriminator \cite{goodfellow2014generative}. 
Despite their success, GANs can suffer from training instability and mode collapse, where only a limited subset of the data distribution is generated \cite{arjovsky2017wasserstein, tang2023deep}. Several variants addressed these issues, including Deep Convolutional GANs \cite{radford2015unsupervised}, Wasserstein GANs \cite{arjovsky2017wasserstein}, and StyleGAN \cite{karras2019style}, which enables high-quality image synthesis and controllable attribute manipulation \cite{karras2020analyzing}. 

GANs have been extensively applied in medical imaging for several generative tasks, including augmenting pathology contrast in mammography \cite{rofena2025lesion, rofena2025augmented}, synthetic augmentation to improve: liver classification in CT\cite{Frid_Adar_2018}, covid-19 in chest X-rays \cite{Waheed_2020, neff2017gan}, dermatology \cite{qasim2021redganattackingclassimbalance}, and brain MRI classification \cite{qi2020saggansemisupervisedattentionguidedgans}. 
In oral oncology, GANs have been used for oral cancer histology classification and segmentation \cite{alosaimi2022efficient, dossantos2023influence} and for fluorescence-based diagnosis \cite{FujimotoAutomaticOralDiagnosis}. 
Conditional GANs further extend this framework by enabling class-specific data generation, particularly useful for rare conditions \cite{mirza2014conditionalgenerativeadversarialnets,tang2023deep}. 

\vspace{2mm}
\textbf{Diffusion Models}. 
Diffusion Models (DMs) are the most recent class of generative models and have shown superior performance over previous approaches, especially in terms of image diversity \cite{ho2020denoising, dhariwal2021diffusion, tang2023deep}. 
DMs iteratively add Gaussian noise to training data and learn to reverse this process through denoising steps using UNet-based architectures \cite{sohl2015deep, ronneberger2015u, ho2020denoising}. During generation, samples are obtained by iteratively denoising random noise. A major drawback is the need for many denoising steps, resulting in longer inference times than single-pass methods \cite{ho2020denoising, tang2023deep}.
To reduce computational costs, Latent Diffusion Models (LDMs) perform diffusion in a compressed latent space learned via autoencoders \cite{rombachetal2022latent}. 
Additionally, Low-Rank Adaptation (LoRA) has been proposed as an efficient fine-tuning strategy, allowing diffusion models to adapt to new domains using minimal trainable parameters and reduced memory requirements \cite{hu2021lora, yang2024low}.
Diffusion-based methods have been applied to address class imbalance in chest X-ray datasets \cite{kazemzadeh2025addressing, pinaya2024adapted} and dermatology images \cite{sagers2023augmentingmedicalimageclassifiers}. In oral cancer, diffusion models have been explored for histopathological images \cite{uliana2025diffusion}, while their application to photographic images remains unexplored.

\subsection{Condition mechanism in image synthesis}\label{sec:conditioning}

Conditioning mechanisms guide generative models toward specific attributes and visual features \cite{bourou2024gans}. In oral cancer diagnosis, conditioning can support class-specific data generation. 

\textbf{Conditioning in GANs}. Conditioning extends GANs by incorporating auxiliary class label information $c$ (e.g., one-hot class encoding, string class name) into the generator $G$ and/or discriminator $D$. 
Given the GAN latent space $z$, basic conditioning approaches consist of concatenating $c$ to $z$, which is effective for low-dimensional inputs such as class labels \cite{bourou2024gans, liu2020diverse}. 
Alternatively, Auxiliary Classifier GANs (AC-GANs) introduce an extra classification head in the discriminator, enforcing class label semantics using an additional supervised loss term \cite{odena2017conditional, bourou2024gans}. 

\textbf{Conditioning in Diffusion Models}. 
Diffusion Models support a wide range of conditioning signals, including text, labels, images, and segmentation masks \cite{zhan2024conditional}. Conditioning can be integrated at different stages using operations such as concatenation, summation, or cross-attention. Text and class labels are typically embedded before being injected into the denoising process \cite{zhan2024conditional}. 
Images or masks can directly guide the diffusion trajectory by constraining regions of interest or initializing the latent representation \cite{ho2020denoising}. 
These mechanisms are particularly well suited to medical imaging, where precise control over generated content is critical \cite{rombachetal2022latent, kazemzadeh2025addressing}.



\section{Photographic Multi-purpose Oral Cancer Imaging Dataset}\label{sec:photomoci}




\bgroup
\def\arraystretch{1.5} 

\begin{table}[b]
\setlength{\tabcolsep}{2.5pt}
\footnotesize
\centering
\caption{Existing photographic oral cancer datasets for computer vision. The column \textit{\#Imgs (w/ lesions)} reports total images and lesion‑positive ones (within round brackets).}

\begin{tabularx}{0.67\textwidth}{
    >{\hsize=2cm}X  c >{\centering\arraybackslash\hsize=1.7cm}X 
    >{\hsize=4.9cm}X
}
    \Xhline{2.1\arrayrulewidth}
    \rowcolor{Gainsboro!120}
    \textbf{Name} & \textbf{Year} & \textbf{\#Imgs (w/ lesions)} & \textbf{Downstream tasks} \\ \hline

    Kaggle OC Lips and Tongue\cite{shivam17299_oral_cancer_2021} & 
    2020 &  
    131\hspace{0.5mm}(87) & 
    Binary image classification \\ \hline

    \rowcolor{Gainsboro!40}
    Roboflow OC\cite{roboflow-oral-cancer-data_dataset} & 
    2021 & 
    323\hspace{0.5mm}(156) & 
    Multi-label image classification, Object detection, Semantic segmentation \\ \hline

    Mendeley Oral Images \cite{diagnostics13213360} & 
    2023 & 
    373\hspace{0.5mm}(158) & 
    Binary image classification \\ \hline

    \rowcolor{Gainsboro!40}
    KOCD \hspace{1cm}\cite{MOHD_ZAID_RASHID_oral_cancer_2024} & 
    2024 & 
    950\hspace{0.5mm}(500)& 
    Binary image classification \\ \hline

    \textbf{PhotoMOCI (ours)} & 
    \textbf{2026} & 
    \textbf{700\hspace{0.5mm}(700)} & 
    Multi-label image classification, Object detection, Semantic segmentation, CBR \\

    \Xhline{2.1\arrayrulewidth}
\end{tabularx}
\label{tab:dataset-review}
\end{table}
\egroup

In this section, we review current public data resources and their limitations. Then, we introduce a novel dataset: Photographic Multi-purpose Oral Cancer Imaging.

We review publicly available oral cancer (OC) photographic datasets released between 2020 and 2025, excluding other imaging modalities, such as histopathology and radiography.
Despite the growing number of works on OC imaging, the availability of datasets remains a major limitation. Many studies rely on private or poorly documented datasets, primarily due to patient privacy concerns and institutional constraints, which significantly hinder reproducibility and fair methodological comparison.
Indeed, several datasets are accessible only upon request and require detailed justification from potential users \cite{rabinovici2024pixels, lee2023early, lin2021automatic, jubair2022novel, welikala2021clinically, heo2022deep, warin2022ai}, resulting in a time-consuming and often discouraging process. Even more restrictive are works in which datasets are not accessible at all, either because they were never intended for public release or due to legal and ethical restrictions \cite{kouketsu2024detection, vinayahalingam2024advancements}.

Table \ref{tab:dataset-review} provides an overview of publicly available photographic datasets sorted by year, detailing size and the supported computer vision downstream tasks. 
Such analysis highlights several limitations in the datasets used for oral healthcare research. First, all publicly available oral cancer datasets contain fewer than 1,000 samples, making it difficult to generalize findings. Second, the number of samples further decreases when annotations are provided for multiple downstream tasks, such as classification, detection, or segmentation. Indeed, three out of five datasets are curated only for binary classification (cancer vs.\ non-cancer). The only dataset supporting multiple downstream tasks comprises just 323 images.
Finally, note that many publicly available datasets appear larger than they are, as they include the "healthy" class. These images are easier to collect because they do not require waiting for rare pathological events. Consequently, they are often used to artificially inflate dataset sizes, potentially overselling the actual clinical contribution while the volume of high-value pathological samples remains limited.
This scarcity of well-annotated, publicly accessible datasets underscores the need for greater efforts to create and share resources that support robust and reproducible research in healthcare.


With this premise, we introduce the Photographic Multi-purpose Oral Cancer Imaging dataset, a multipurpose dataset designed for cancer recognition in oral cavity photographic imaging. It supports a wide range of downstream tasks (some of which were previously addressed), including major computer vision problems such as image classification, object detection\cite{oral_lesion_detection}, and semantic segmentation\cite{oral_lesion_segmentation}, as well as clinically oriented applications like case-based reasoning \cite{PAROLA2024102433,kenny2019twin,CIMINO2025100538,keane2019case}. 
By offering a dataset combining technical and clinical needs, PhotoMOCI supports research on photographic oral cancer screening. Nevertheless, this work specifically focuses on image classification.

The study adhered to the ethical standards outlined in the Declaration of Helsinki and its amendments for research involving human participants. Ethical approval was obtained from the local ethics committee (n.12 07/05/2024). Written informed consent was obtained from all participants following established guidelines for conducting scientific research.

Between 2021 and 2025, photographic images of the oral cavity were collected from patients attending medical examinations at the Oral Medicine Unit of the P. Giaccone University Hospital in Palermo, Italy. Dental hygienists and oral medicine consultants acquired these images using a Nikon D7200 digital camera equipped with a 105 mm lens and an SB-R200 macro flash, offering an inexpensive alternative to advanced imaging systems. The resulting image dataset is composed of high-quality resolution ranging from $\sim$2500x2000 to $\sim$6000x4000 pixels.
Additionally, histopathological analysis was performed to confirm the diagnosis and subsequent labeling. 
The final version includes \textbf{700} images belonging to three specific pathologies: \textbf{aphthous}, \textbf{traumatic}, and \textbf{neoplastic} (cancer class). The majority of images in the PhotoMOCI dataset feature a single lesion; however, 22 images contain multiple instances of lesions within the same photo for a total of 730 oral lesions (253 aphthous, 257 traumatic, and 220 neoplastic). Note that there are no instances where lesions from different classes coexist within the same image. All cases involving multiple lesions pertain to a specific scenario: the multiple aphthous lesions. The dataset contains no more than one image from the same patient.

\begin{figure}[t]
    \centering
    \begin{subfigure}[t]{0.35\linewidth}
        \centering
        \includegraphics[height=3.0cm]{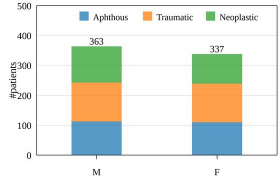}
        \caption{Sex distribution.}
        \label{fig:photomoci-gender-distribution}
    \end{subfigure}
    \hfill
    \begin{subfigure}[t]{0.60\linewidth}
        \centering
        \includegraphics[height=3.0cm]{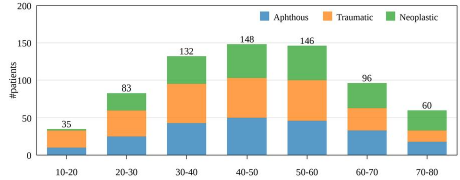}
        \caption{Age distribution.}
        \label{fig:photomoci-age-distribution}
    \end{subfigure}
    \caption{PhotoMOCI metadata distribution by sex in (a) and age group in (b), stratified by pathology class.}
    \label{fig:photomoci-metadata-distribution}
\end{figure}

Figure~\ref{fig:photomoci-metadata-distribution} summarizes the PhotoMOCI metadata distribution by sex and age group, with each bar split according to the three pathology classes. The sex distribution shows a small mismatch, with more men than women; however, across the diagnostic classes, the samples remain balanced. Regarding age, most individuals involved in the data collection are between 30 and 60 years old. In the younger segment (10--20 years), the neoplastic class is rare and traumatic lesions dominate, whereas neoplastic lesions occur more frequently among older individuals.
Other commonly reported clinical metadata, such as anatomical site and pathological stage, and visual lesion features, such as lesion size, border color, and lesion color, are not available for PhotoMOCI.

For the core computer vision tasks, the images were annotated using the open-source tool \texttt{COCO Annotator} \cite{cocoannotator}, generating a dataset in the standard COCO format for computer vision applications. 
Initial annotation of lesion regions was performed manually by a dental intern, who delineated lesion boundaries and assigned pathology labels returned by histopathological analysis. Subsequently, all annotated images were reviewed and refined by two senior clinicians with more clinical experience. The revisions focused exclusively on segmentation boundaries to improve their accuracy, as the labels are based on the histopathological results.
The bounding box is then generated by considering the minimum rectangle that encloses the lesion segment. The images and the json annotation files can be accessed on Kaggle \cite{marco_parola_2025}.

\section{Methodology}\label{sec:methodology}

\begin{figure*}[t]
\centering
    \centering
    \includegraphics[width=\linewidth]{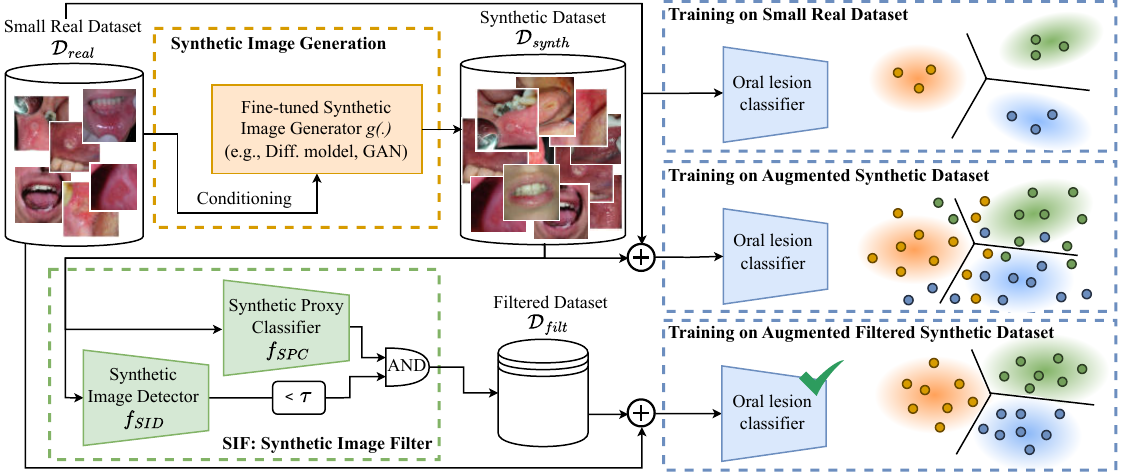}

  \caption{Overview of the proposed methodology for oral lesion classification. In the \textcolor{Apricot}{ochre} dotted box, the Synthetic Image Generation based on generative models. In the \textcolor{LimeGreen}{green} dotted box, the Synthetic Image Filter combining the Synthetic Proxy Classifier and the Synthetic Image Detector to select informative synthetic samples. In the \textcolor{RoyalBlue}{light-blue} dotted boxes, three training setups: using the original small dataset (top), the augmented synthetic dataset (middle), and the augmented synthetic filtered dataset (bottom), the latter aiming to improve generalization.}
  \label{fig:oral-synthetic-scema}
\end{figure*}

The proposed methodology follows a three-stage pipeline designed to generate, select, and exploit synthetic data for oral cancer classification, as illustrated in Figure \ref{fig:oral-synthetic-scema} and detailed in Sections \ref{sec:synthetic_image_generation}, \ref{sec:synthetic_image_selection}, and \ref{sec:training_oral_cancer_classifier}, respectively.

\begin{itemize}
    \item \textbf{Synthetic Image Generation} (\textcolor{Apricot}{ochre} dotted box). This stage relies on a generative model fine-tuned on an OC dataset (usually small) to generate synthetic images.
    
    \item \textbf{Synthetic Image Filter} (\textcolor{LimeGreen}{green} dotted box). This stage exploits the proposed SIF to select specific synthetic images that, coupled with the original dataset, improve the oral lesion classifier. 
    
    \item \textbf{Training the final oral cancer classifier} (\textcolor{RoyalBlue}{ligh-blue} dotted box). This stage is responsible for training the oral lesion classifier under different setups.
\end{itemize}

\subsection{Synthetic Image Generation}\label{sec:synthetic_image_generation}

Let $\mathcal{D}_{real}=\{(x_i, y_i)\}_{i=1}^{N}$ be the Small Real Dataset, where $x_i \in \mathcal{X}$ denotes the input real image belonging to image space $\mathcal{X}$ and $y_i \in \mathcal{Y}$ its corresponding ground‑truth label from a predefined set $\mathcal{Y}$. Let $\mathcal{D}_{synth}=\{(x_j, y_j)\}_{j=1}^{M}$ be the Synthetic Dataset, where $x_j \in \mathcal{X}$ is a synthetic image and $y_j \in \mathcal{Y}$ the corresponding label derived from the generation process. The Synthetic Image Generation stage is responsible for generating $\mathcal{D}_{synth}$, relying on a Synthetic Image Generator denoted as $g(\cdot)$ finetuned on $\mathcal{D}_{real}$.

\begin{figure*}[bp]
    \centering

    \begin{minipage}{\linewidth}
        \begin{minipage}{0.15cm}\footnotesize{0}\hspace{-0.2mm}\end{minipage}
        \begin{minipage}{\linewidth}
            \begin{subfigure}{0.048\textwidth}
                \centering \hfill 
                \adjustbox{cfbox=red 1.4pt -1.2pt}{\includegraphics[width=\linewidth]{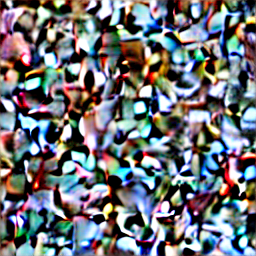}}
            \end{subfigure}\hspace{-1mm}
            \begin{subfigure}{0.048\textwidth}
                \centering \hfill
                \includegraphics[width=\linewidth]{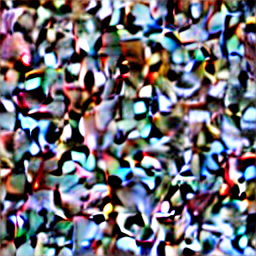}
            \end{subfigure}\hspace{-1mm}
            \begin{subfigure}{0.048\textwidth}
                \centering \hfill
                \includegraphics[width=\linewidth]{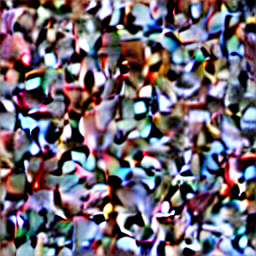}
            \end{subfigure}\hspace{-1mm}
            \begin{subfigure}{0.048\textwidth}
                \centering \hfill
                \includegraphics[width=\linewidth]{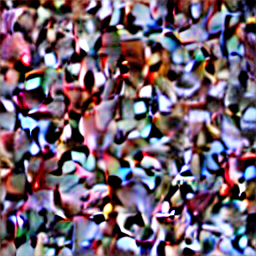}
            \end{subfigure}\hspace{-1mm}
            \begin{subfigure}{0.048\textwidth}
                \centering \hfill
                \includegraphics[width=\linewidth]{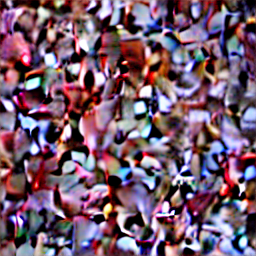}
            \end{subfigure}\hspace{-1mm}      
            \begin{subfigure}{0.048\textwidth}
                \centering \hfill
                \includegraphics[width=\linewidth]{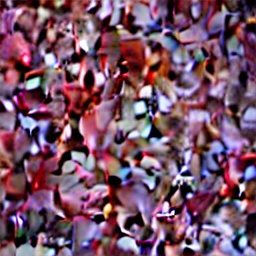}
            \end{subfigure} \hspace{-2.2mm}
            \begin{subfigure}{0.048\textwidth}
                \centering \hfill
                \includegraphics[width=\linewidth]{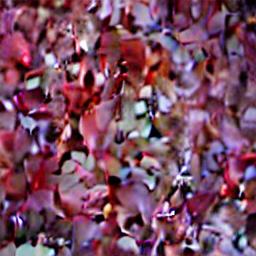}
            \end{subfigure} \hspace{-2.2mm}
            \begin{subfigure}{0.048\textwidth}
                \centering \hfill
                \includegraphics[width=\linewidth]{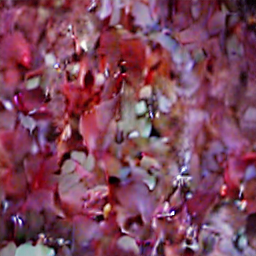}
            \end{subfigure}\hspace{-1mm}
            \begin{subfigure}{0.048\textwidth}
                \centering \hfill
                \includegraphics[width=\linewidth]{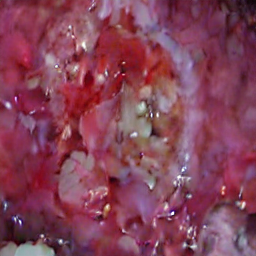}
            \end{subfigure}\hspace{-1mm}
            \begin{subfigure}{0.048\textwidth}
                \centering \hfill
                \includegraphics[width=\linewidth]{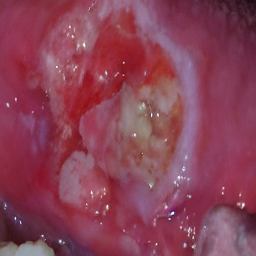}
            \end{subfigure}\hspace{-0.4mm}
            \begin{subfigure}{0.048\textwidth}
                \centering \hfill 
                \adjustbox{cfbox=red 1.4pt -1.2pt}{\includegraphics[width=\linewidth]{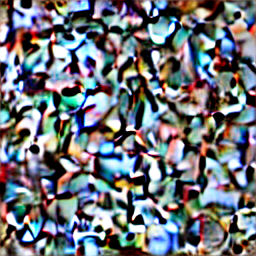}}
            \end{subfigure}\hspace{-1mm}
            \begin{subfigure}{0.048\textwidth}
                \centering \hfill
                \includegraphics[width=\linewidth]{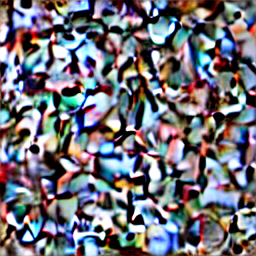}
            \end{subfigure}\hspace{-1mm}
            \begin{subfigure}{0.048\textwidth}
                \centering \hfill
                \includegraphics[width=\linewidth]{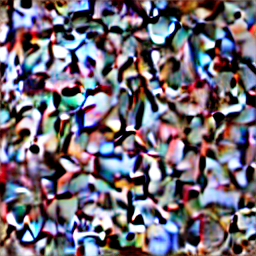}
            \end{subfigure}\hspace{-1mm}
            \begin{subfigure}{0.048\textwidth}
                \centering \hfill
                \includegraphics[width=\linewidth]{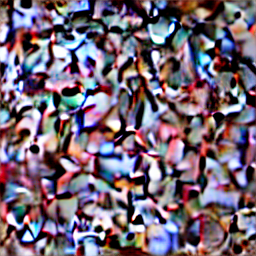}
            \end{subfigure}\hspace{-1mm}
            \begin{subfigure}{0.048\textwidth}
                \centering \hfill
                \includegraphics[width=\linewidth]{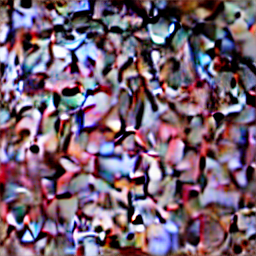}
            \end{subfigure} \hspace{-2.2mm}         
            \begin{subfigure}{0.048\textwidth}
                \centering \hfill
                \includegraphics[width=\linewidth]{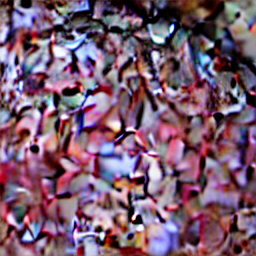}
            \end{subfigure}\hspace{-1mm}
            \begin{subfigure}{0.048\textwidth}
                \centering \hfill
                \includegraphics[width=\linewidth]{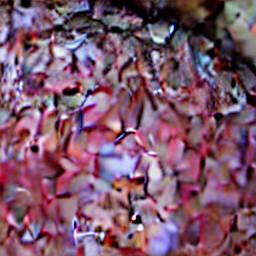}
            \end{subfigure}\hspace{-1mm}
            \begin{subfigure}{0.048\textwidth}
                \centering \hfill
                \includegraphics[width=\linewidth]{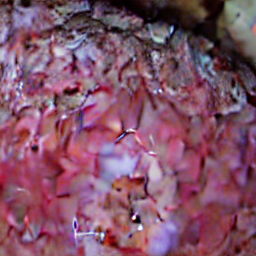}
            \end{subfigure}\hspace{-1mm}
            \begin{subfigure}{0.048\textwidth}
                \centering \hfill
                \includegraphics[width=\linewidth]{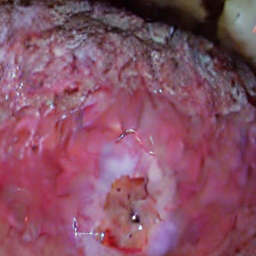}
            \end{subfigure}\hspace{-1mm}
            \begin{subfigure}{0.048\textwidth}
                \centering \hfill
                \includegraphics[width=\linewidth]{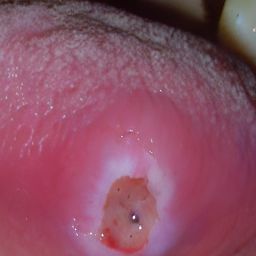}
            \end{subfigure}
        \end{minipage}
        \vspace{0.1mm}
    \end{minipage}
    \begin{minipage}{\linewidth}
         \begin{minipage}{0.15cm}\footnotesize{1}\end{minipage}
        \begin{minipage}{\linewidth}
            \begin{subfigure}{0.048\textwidth}
                \centering \hfill 
                \includegraphics[width=\linewidth]{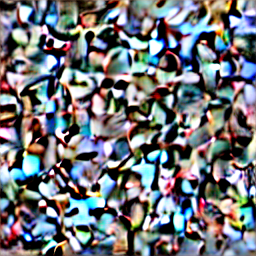}
            \end{subfigure}\hspace{-1mm}
            \begin{subfigure}{0.048\textwidth}
                \centering \hfill
                \adjustbox{cfbox=red 1.4pt -1.2pt}{\includegraphics[width=\linewidth]{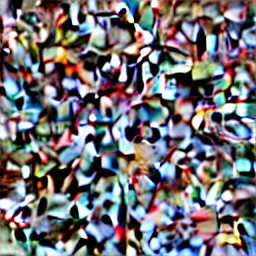}}
            \end{subfigure}\hspace{-1mm}
            \begin{subfigure}{0.048\textwidth}
                \centering \hfill
                \includegraphics[width=\linewidth]{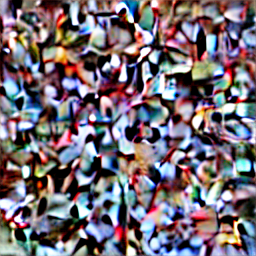}
            \end{subfigure}\hspace{-1mm}
            \begin{subfigure}{0.048\textwidth}
                \centering \hfill
                \includegraphics[width=\linewidth]{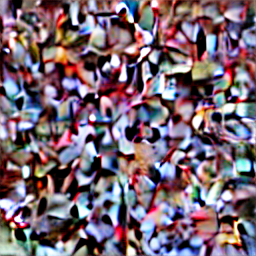}
            \end{subfigure}\hspace{-1mm}
            \begin{subfigure}{0.048\textwidth}
                \centering \hfill
                \includegraphics[width=\linewidth]{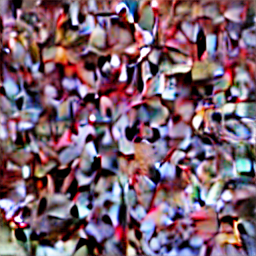}
            \end{subfigure}\hspace{-1mm}
            \begin{subfigure}{0.048\textwidth}
                \centering \hfill
                \includegraphics[width=\linewidth]{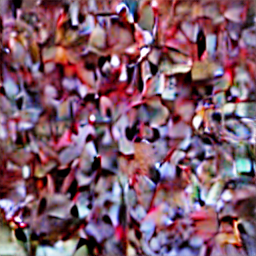}
            \end{subfigure}\hspace{-1mm}
            \begin{subfigure}{0.048\textwidth}
                \centering \hfill
                \includegraphics[width=\linewidth]{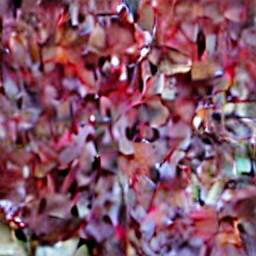}
            \end{subfigure}\hspace{-1mm}
            \begin{subfigure}{0.048\textwidth}
                \centering \hfill
                \includegraphics[width=\linewidth]{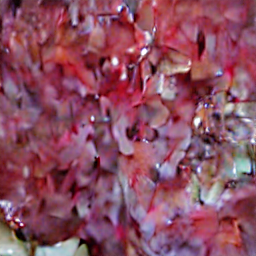}
            \end{subfigure}\hspace{-1mm}
            \begin{subfigure}{0.048\textwidth}
                \centering \hfill
                \includegraphics[width=\linewidth]{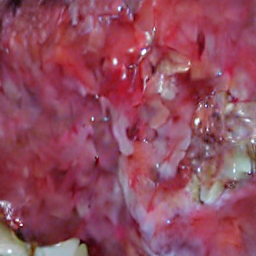}
            \end{subfigure}\hspace{-1mm}
            \begin{subfigure}{0.048\textwidth}
                \centering \hfill
                \includegraphics[width=\linewidth]{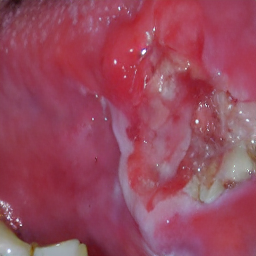}
            \end{subfigure}\hspace{-0.4mm}
            \begin{subfigure}{0.048\textwidth}
                \centering \hfill 
                \includegraphics[width=\linewidth]{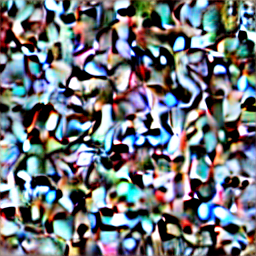}
            \end{subfigure}\hspace{-1mm}
            \begin{subfigure}{0.048\textwidth}
                \centering \hfill
                \adjustbox{cfbox=red 1.4pt -1.2pt}{\includegraphics[width=\linewidth]{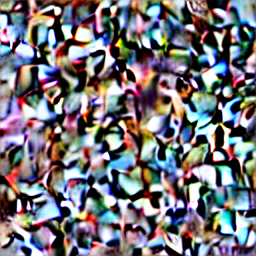}}
            \end{subfigure}\hspace{-1mm}
            \begin{subfigure}{0.048\textwidth}
                \centering \hfill
                \includegraphics[width=\linewidth]{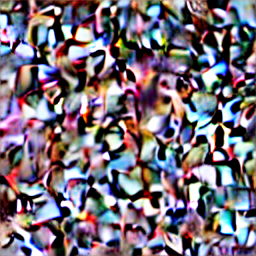}
            \end{subfigure}\hspace{-1mm}
            \begin{subfigure}{0.048\textwidth}
                \centering \hfill
                \includegraphics[width=\linewidth]{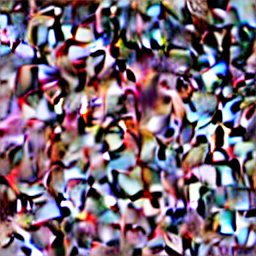}
            \end{subfigure}\hspace{-1mm}
            \begin{subfigure}{0.048\textwidth}
                \centering \hfill
                \includegraphics[width=\linewidth]{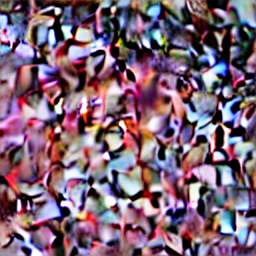}
            \end{subfigure}\hspace{-1mm}
            \begin{subfigure}{0.048\textwidth}
                \centering \hfill
                \includegraphics[width=\linewidth]{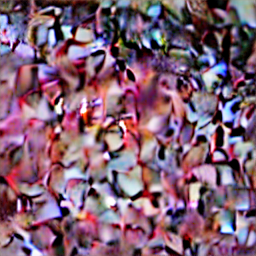}
            \end{subfigure}\hspace{-1mm}
            \begin{subfigure}{0.048\textwidth}
                \centering \hfill
                \includegraphics[width=\linewidth]{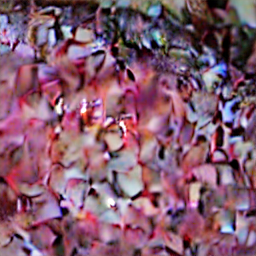}
            \end{subfigure}\hspace{-1mm}
            \begin{subfigure}{0.048\textwidth}
                \centering \hfill
                \includegraphics[width=\linewidth]{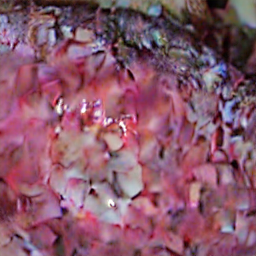}
            \end{subfigure}\hspace{-1mm}
            \begin{subfigure}{0.048\textwidth}
                \centering \hfill
                \includegraphics[width=\linewidth]{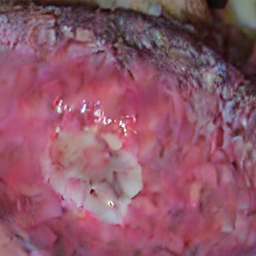}
            \end{subfigure}\hspace{-1mm}
            \begin{subfigure}{0.048\textwidth}
                \centering \hfill
                \includegraphics[width=\linewidth]{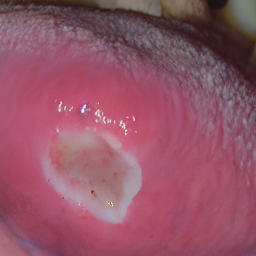}
            \end{subfigure}
        \end{minipage}
        \vspace{0.1mm}
    \end{minipage}
    \begin{minipage}{\linewidth}
         \begin{minipage}{0.15cm}\footnotesize{2}\hspace{-0.2mm}\end{minipage}
         \begin{minipage}{\linewidth}
            \begin{subfigure}{0.048\textwidth}
                \centering \hfill 
                \includegraphics[width=\linewidth]{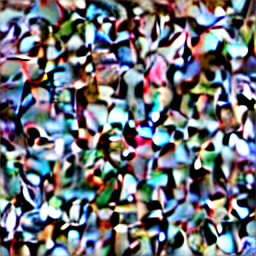}
            \end{subfigure}\hspace{-1mm}
            \begin{subfigure}{0.048\textwidth}
                \centering \hfill
                \includegraphics[width=\linewidth]{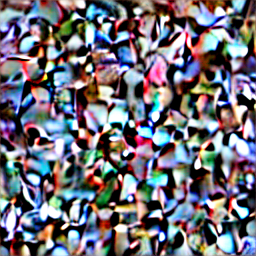}
            \end{subfigure}\hspace{-1mm}
            \begin{subfigure}{0.048\textwidth}
                \centering \hfill
                \adjustbox{cfbox=red 1.4pt -1.2pt}{\includegraphics[width=\linewidth]{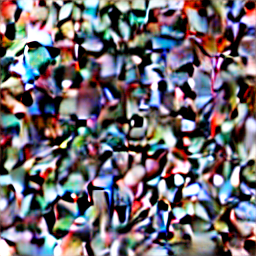}}
            \end{subfigure}\hspace{-1mm}
            \begin{subfigure}{0.048\textwidth}
                \centering \hfill
                \includegraphics[width=\linewidth]{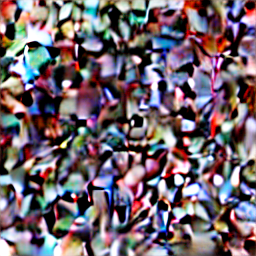}
            \end{subfigure}\hspace{-1mm}
            \begin{subfigure}{0.048\textwidth}
                \centering \hfill
                \includegraphics[width=\linewidth]{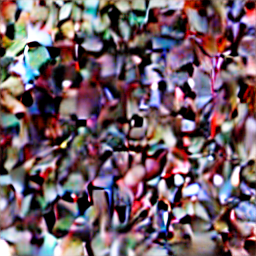}
            \end{subfigure}\hspace{-1mm}
            \begin{subfigure}{0.048\textwidth}
                \centering \hfill
                \includegraphics[width=\linewidth]{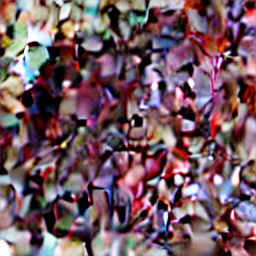}
            \end{subfigure}\hspace{-1mm}
            \begin{subfigure}{0.048\textwidth}
                \centering \hfill
                \includegraphics[width=\linewidth]{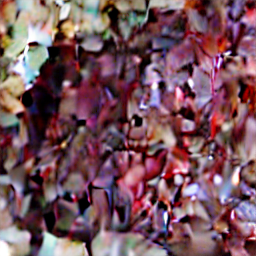}
            \end{subfigure}\hspace{-1mm}
            \begin{subfigure}{0.048\textwidth}
                \centering \hfill
                \includegraphics[width=\linewidth]{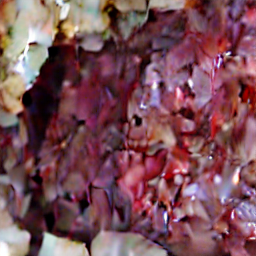}
            \end{subfigure}\hspace{-1mm}
            \begin{subfigure}{0.048\textwidth}
                \centering \hfill
                \includegraphics[width=\linewidth]{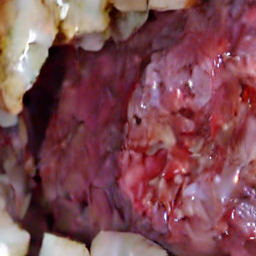}
            \end{subfigure}\hspace{-1mm}
            \begin{subfigure}{0.048\textwidth}
                \centering \hfill
                \includegraphics[width=\linewidth]{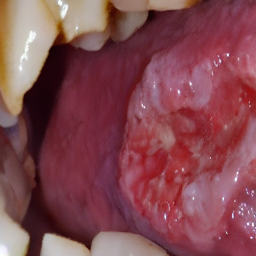}
            \end{subfigure}\hspace{-0.4mm}
            \begin{subfigure}{0.048\textwidth}
                \centering \hfill 
                \includegraphics[width=\linewidth]{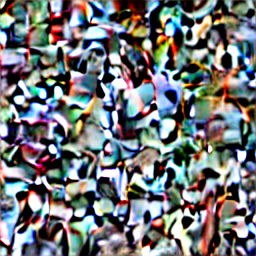}
            \end{subfigure}\hspace{-1mm}
            \begin{subfigure}{0.048\textwidth}
                \centering \hfill
                \includegraphics[width=\linewidth]{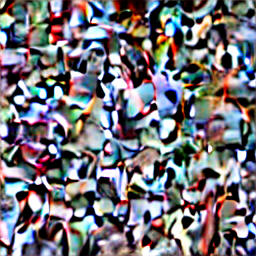}
            \end{subfigure}\hspace{-1mm}
            \begin{subfigure}{0.048\textwidth}
                \centering \hfill
                \adjustbox{cfbox=red 1.4pt -1.2pt}{\includegraphics[width=\linewidth]{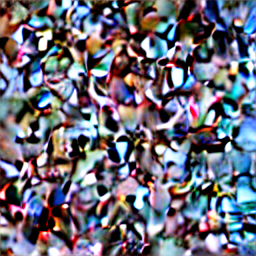}}
            \end{subfigure}\hspace{-1mm}
            \begin{subfigure}{0.048\textwidth}
                \centering \hfill
                \includegraphics[width=\linewidth]{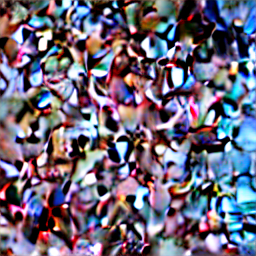}
            \end{subfigure}\hspace{-1mm}
            \begin{subfigure}{0.048\textwidth}
                \centering \hfill
                \includegraphics[width=\linewidth]{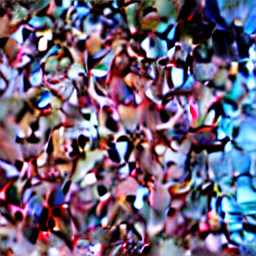}
            \end{subfigure}\hspace{-1mm}
            \begin{subfigure}{0.048\textwidth}
                \centering \hfill
                \includegraphics[width=\linewidth]{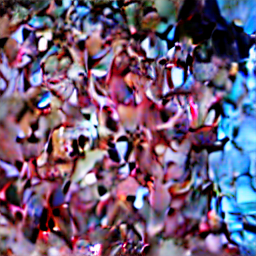}
            \end{subfigure}\hspace{-1mm}
            \begin{subfigure}{0.048\textwidth}
                \centering \hfill
                \includegraphics[width=\linewidth]{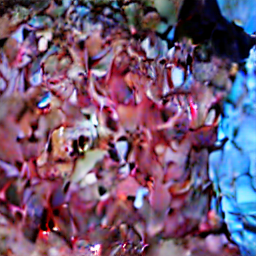}
            \end{subfigure}\hspace{-1mm}
            \begin{subfigure}{0.048\textwidth}
                \centering \hfill
                \includegraphics[width=\linewidth]{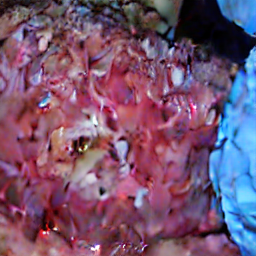}
            \end{subfigure}\hspace{-1mm}
            \begin{subfigure}{0.048\textwidth}
                \centering \hfill
                \includegraphics[width=\linewidth]{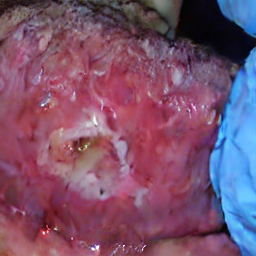}
            \end{subfigure}\hspace{-1mm}
            \begin{subfigure}{0.048\textwidth}
                \centering \hfill
                \includegraphics[width=\linewidth]{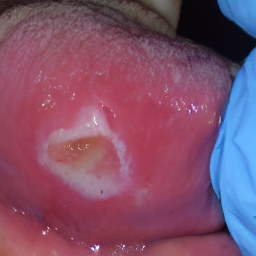}
            \end{subfigure}
        \end{minipage}
        \vspace{0.1mm}
    \end{minipage}
    \begin{minipage}{\linewidth}
        \begin{minipage}{0.15cm}\footnotesize{3}\end{minipage}
        \begin{minipage}{\linewidth}
            \begin{subfigure}{0.048\textwidth}
                \centering \hfill 
                \includegraphics[width=\linewidth]{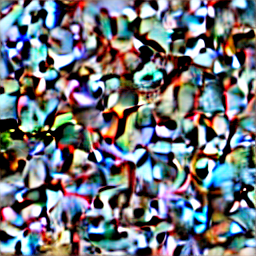}
            \end{subfigure}\hspace{-1mm}
            \begin{subfigure}{0.048\textwidth}
                \centering \hfill
                \includegraphics[width=\linewidth]{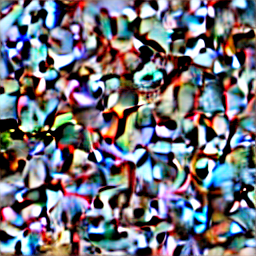}
            \end{subfigure}\hspace{-1mm}
            \begin{subfigure}{0.048\textwidth}
                \centering \hfill
                \includegraphics[width=\linewidth]{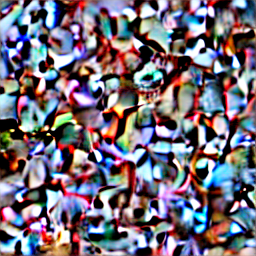}
            \end{subfigure}\hspace{-1mm}
            \begin{subfigure}{0.048\textwidth}
                \centering \hfill
                \adjustbox{cfbox=red 1.4pt -1.2pt}{\includegraphics[width=\linewidth]{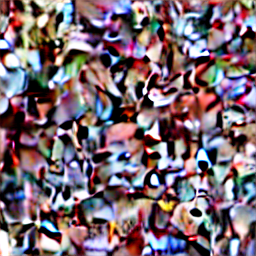}}
            \end{subfigure}\hspace{-1mm}
            \begin{subfigure}{0.048\textwidth}
                \centering \hfill
                \includegraphics[width=\linewidth]{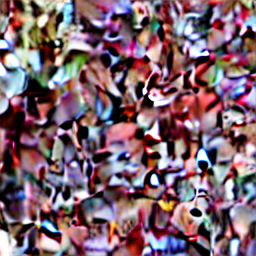}
            \end{subfigure}\hspace{-1mm}
            \begin{subfigure}{0.048\textwidth}
                \centering \hfill
                \includegraphics[width=\linewidth]{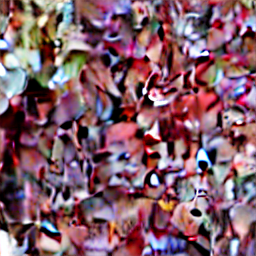}
            \end{subfigure}\hspace{-1mm}
            \begin{subfigure}{0.048\textwidth}
                \centering \hfill
                \includegraphics[width=\linewidth]{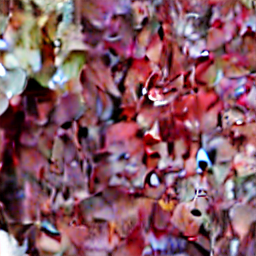}
            \end{subfigure}\hspace{-1mm}
            \begin{subfigure}{0.048\textwidth}
                \centering \hfill
                \includegraphics[width=\linewidth]{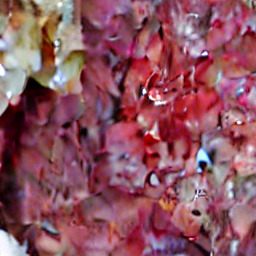}
            \end{subfigure}\hspace{-1mm}
            \begin{subfigure}{0.048\textwidth}
                \centering \hfill
                \includegraphics[width=\linewidth]{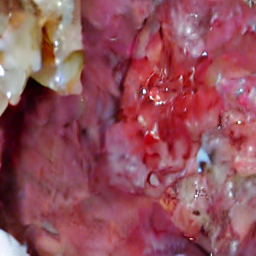}
            \end{subfigure}\hspace{-1mm}
            \begin{subfigure}{0.048\textwidth}
                \centering \hfill
                \includegraphics[width=\linewidth]{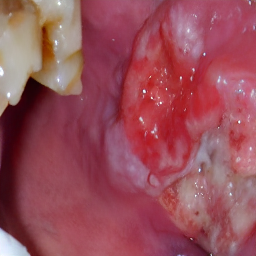}
            \end{subfigure}\hspace{-0.4mm}
            \begin{subfigure}{0.048\textwidth}
                \centering \hfill 
                \includegraphics[width=\linewidth]{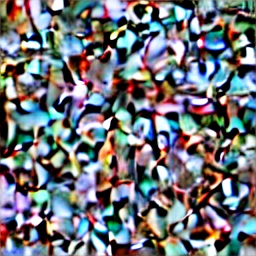}
            \end{subfigure}\hspace{-1mm}
            \begin{subfigure}{0.048\textwidth}
                \centering \hfill
                \includegraphics[width=\linewidth]{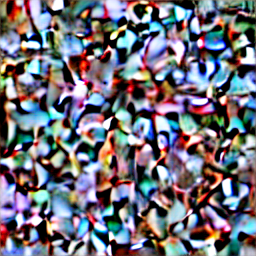}
            \end{subfigure}\hspace{-1mm}
            \begin{subfigure}{0.048\textwidth}
                \centering \hfill
                \includegraphics[width=\linewidth]{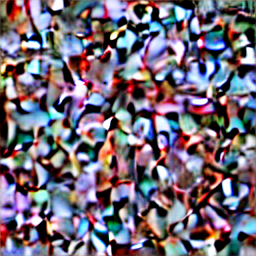}
            \end{subfigure}\hspace{-1mm}
            \begin{subfigure}{0.048\textwidth}
                \centering \hfill
                \adjustbox{cfbox=red 1.4pt -1.2pt}{\includegraphics[width=\linewidth]{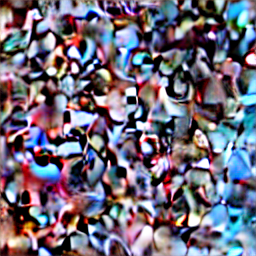}}
            \end{subfigure}\hspace{-1mm}
            \begin{subfigure}{0.048\textwidth}
                \centering \hfill
                \includegraphics[width=\linewidth]{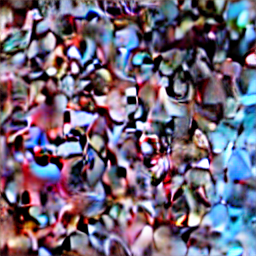}
            \end{subfigure}\hspace{-1mm}
            \begin{subfigure}{0.048\textwidth}
                \centering \hfill
                \includegraphics[width=\linewidth]{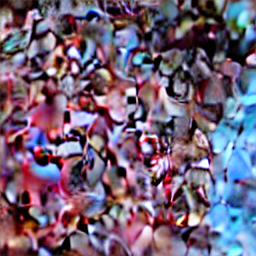}
            \end{subfigure}\hspace{-1mm}
            \begin{subfigure}{0.048\textwidth}
                \centering \hfill
                \includegraphics[width=\linewidth]{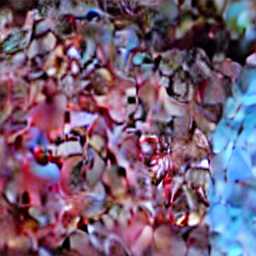}
            \end{subfigure}\hspace{-1mm}
            \begin{subfigure}{0.048\textwidth}
                \centering \hfill
                \includegraphics[width=\linewidth]{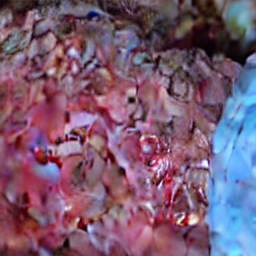}
            \end{subfigure}\hspace{-1mm}
            \begin{subfigure}{0.048\textwidth}
                \centering \hfill
                \includegraphics[width=\linewidth]{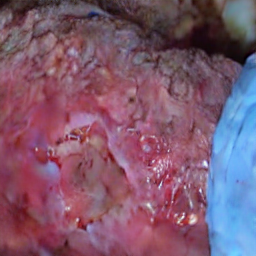}
            \end{subfigure}\hspace{-1mm}
            \begin{subfigure}{0.048\textwidth}
                \centering \hfill
                \includegraphics[width=\linewidth]{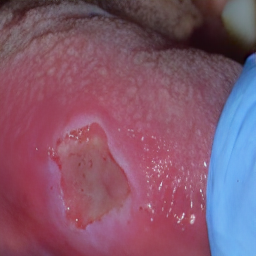}
            \end{subfigure}
            \vspace{0.4mm}
        \end{minipage}
        \begin{minipage}{\linewidth}
            \begin{minipage}{0.15cm}\footnotesize{4}\end{minipage}
            \begin{minipage}{\linewidth}
                \begin{subfigure}{0.048\textwidth}
                    \centering \hfill 
                    \includegraphics[width=\linewidth]{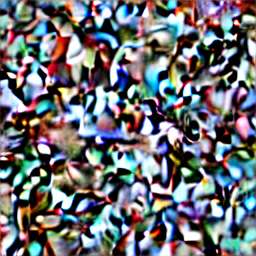}
                \end{subfigure}\hspace{-1mm}
                \begin{subfigure}{0.048\textwidth}
                    \centering \hfill
                    \includegraphics[width=\linewidth]{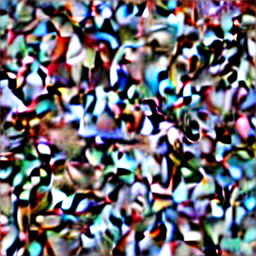}
                \end{subfigure}\hspace{-1mm}
                \begin{subfigure}{0.048\textwidth}
                    \centering \hfill
                    \includegraphics[width=\linewidth]{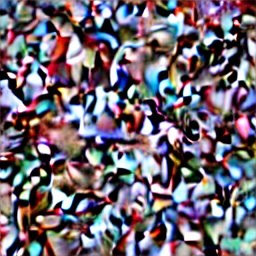}
                \end{subfigure}\hspace{-1mm}
                \begin{subfigure}{0.048\textwidth}
                    \centering \hfill
                    \includegraphics[width=\linewidth]{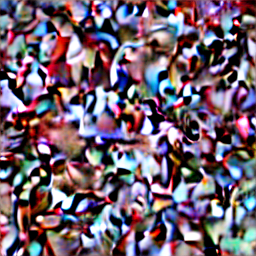}
                \end{subfigure}\hspace{-1mm}
                \begin{subfigure}{0.048\textwidth}
                    \centering \hfill
                    \adjustbox{cfbox=red 1.4pt -1.2pt}{\includegraphics[width=\linewidth]{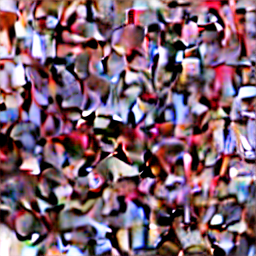}}
                \end{subfigure}\hspace{-1mm}
                \begin{subfigure}{0.048\textwidth}
                    \centering \hfill
                    \includegraphics[width=\linewidth]{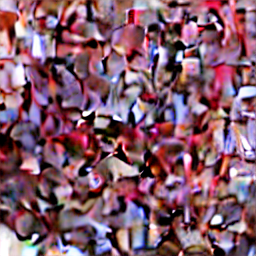}
                \end{subfigure}\hspace{-1mm}
                \begin{subfigure}{0.048\textwidth}
                    \centering \hfill
                    \includegraphics[width=\linewidth]{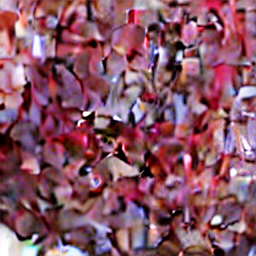}
                \end{subfigure}\hspace{-1mm}
                \begin{subfigure}{0.048\textwidth}
                    \centering \hfill
                    \includegraphics[width=\linewidth]{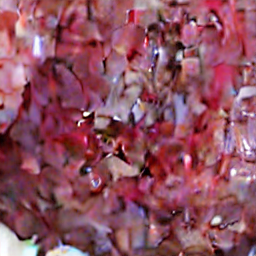}
                \end{subfigure}\hspace{-1mm}
                \begin{subfigure}{0.048\textwidth}
                    \centering \hfill
                    \includegraphics[width=\linewidth]{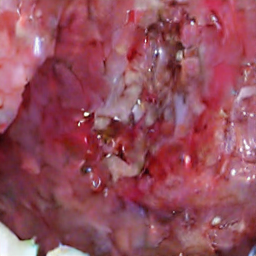}
                \end{subfigure}\hspace{-1mm}
                \begin{subfigure}{0.048\textwidth}
                    \centering \hfill
                    \includegraphics[width=\linewidth]{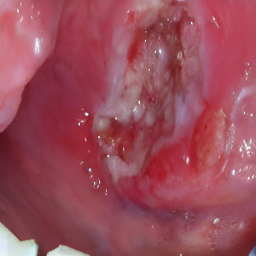}
                \end{subfigure}\hspace{-0.4mm}
                \begin{subfigure}{0.048\textwidth}
                    \centering \hfill 
                    \includegraphics[width=\linewidth]{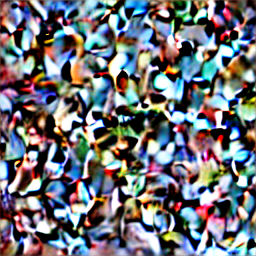}
                \end{subfigure}\hspace{-1mm}
                \begin{subfigure}{0.048\textwidth}
                    \centering \hfill
                    \includegraphics[width=\linewidth]{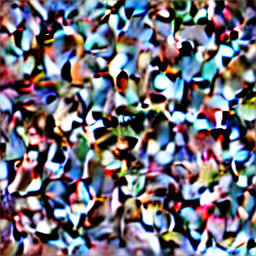}
                \end{subfigure}\hspace{-1mm}
                \begin{subfigure}{0.048\textwidth}
                    \centering \hfill
                    \includegraphics[width=\linewidth]{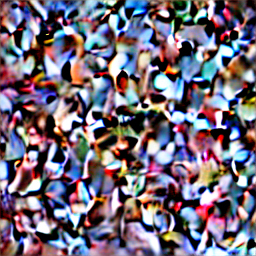}
                \end{subfigure}\hspace{-1mm}
                \begin{subfigure}{0.048\textwidth}
                    \centering \hfill
                    \includegraphics[width=\linewidth]{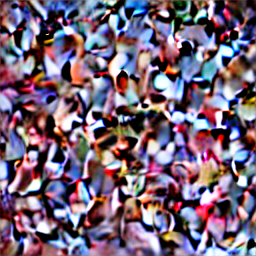}
                \end{subfigure}\hspace{-1mm}
                \begin{subfigure}{0.048\textwidth}
                    \centering \hfill
                    \adjustbox{cfbox=red 1.4pt -1.2pt}{\includegraphics[width=\linewidth]{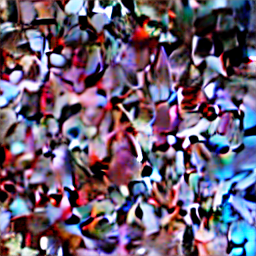}}
                \end{subfigure}\hspace{-1mm}
                \begin{subfigure}{0.048\textwidth}
                    \centering \hfill
                    \includegraphics[width=\linewidth]{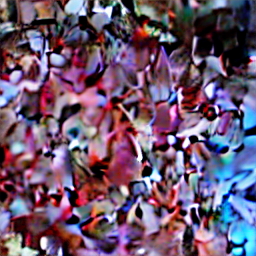}
                \end{subfigure}\hspace{-1mm}
                \begin{subfigure}{0.048\textwidth}
                    \centering \hfill
                    \includegraphics[width=\linewidth]{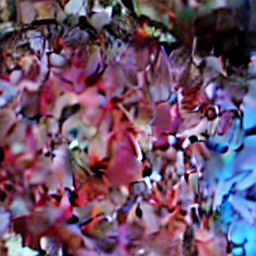}
                \end{subfigure}\hspace{-1mm}
                \begin{subfigure}{0.048\textwidth}
                    \centering \hfill
                    \includegraphics[width=\linewidth]{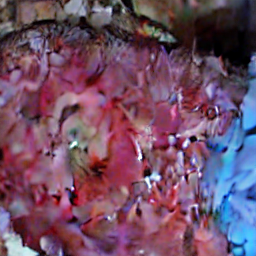}
                \end{subfigure}\hspace{-1mm}
                \begin{subfigure}{0.048\textwidth}
                    \centering \hfill
                    \includegraphics[width=\linewidth]{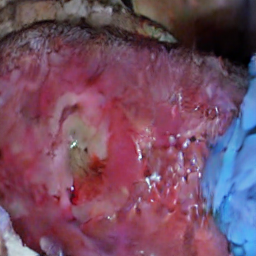}
                \end{subfigure}\hspace{-1mm}
                \begin{subfigure}{0.048\textwidth}
                    \centering \hfill
                    \includegraphics[width=\linewidth]{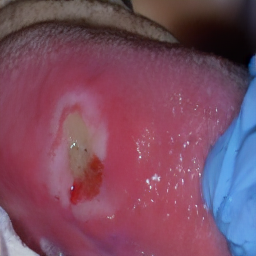}
                \end{subfigure}    
            \end{minipage}
            \vspace{0.1mm}
        \end{minipage}
        \begin{minipage}{\linewidth}
            \begin{minipage}{0.15cm}\footnotesize{5}\end{minipage}
            \begin{minipage}{\linewidth}
                \begin{subfigure}{0.048\textwidth}
                    \centering \hfill 
                    \includegraphics[width=\linewidth]{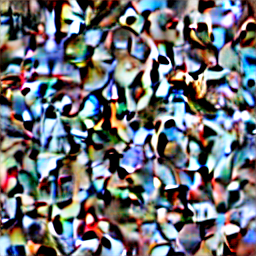}
                \end{subfigure}\hspace{-1mm}
                \begin{subfigure}{0.048\textwidth}
                    \centering \hfill
                    \includegraphics[width=\linewidth]{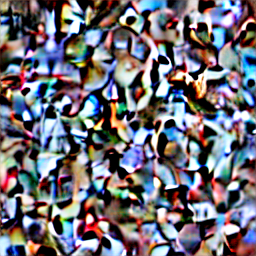}
                \end{subfigure}\hspace{-1mm}
                \begin{subfigure}{0.048\textwidth}
                    \centering \hfill
                    \includegraphics[width=\linewidth]{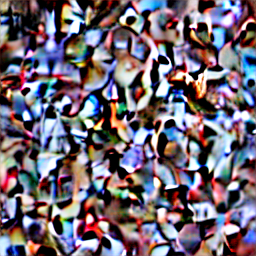}
                \end{subfigure}\hspace{-1mm}
                \begin{subfigure}{0.048\textwidth}
                    \centering \hfill
                    \includegraphics[width=\linewidth]{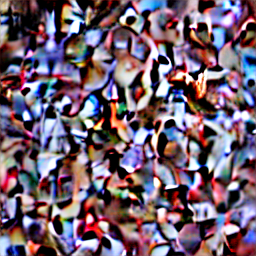}
                \end{subfigure}\hspace{-1mm}
                \begin{subfigure}{0.048\textwidth}
                    \centering \hfill
                    \includegraphics[width=\linewidth]{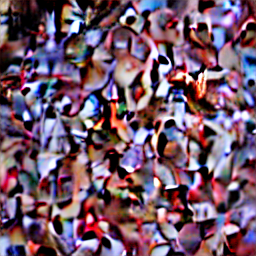}
                \end{subfigure}\hspace{-1mm}
                \begin{subfigure}{0.048\textwidth}
                    \centering \hfill
                    \adjustbox{cfbox=red 1.4pt -1.2pt}{\includegraphics[width=\linewidth]{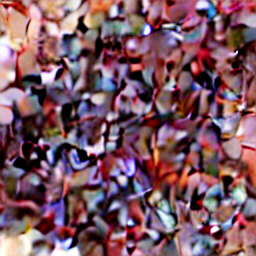}}
                \end{subfigure}\hspace{-1mm}
                \begin{subfigure}{0.048\textwidth}
                    \centering \hfill
                    \includegraphics[width=\linewidth]{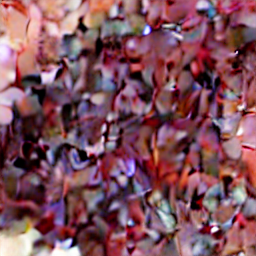}
                \end{subfigure}\hspace{-1mm}
                \begin{subfigure}{0.048\textwidth}
                    \centering \hfill
                    \includegraphics[width=\linewidth]{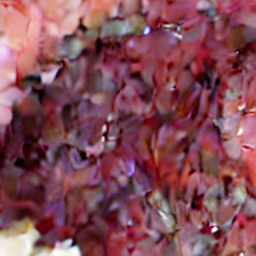}
                \end{subfigure}\hspace{-1mm}
                \begin{subfigure}{0.048\textwidth}
                    \centering \hfill
                    \includegraphics[width=\linewidth]{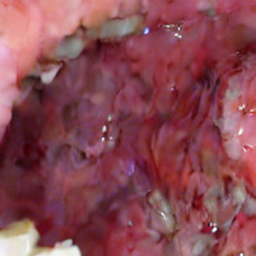}
                \end{subfigure}\hspace{-1mm}
                \begin{subfigure}{0.048\textwidth}
                    \centering \hfill
                    \includegraphics[width=\linewidth]{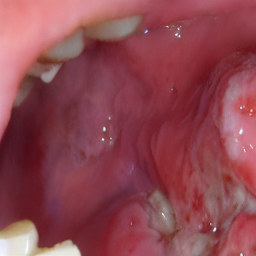}
                \end{subfigure}\hspace{-0.4mm}
                \begin{subfigure}{0.048\textwidth}
                    \centering \hfill 
                    \includegraphics[width=\linewidth]{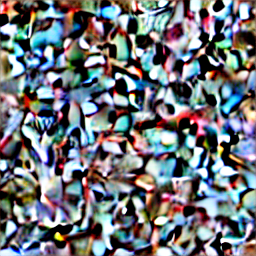}
                \end{subfigure}\hspace{-1mm}
                \begin{subfigure}{0.048\textwidth}
                    \centering \hfill
                    \includegraphics[width=\linewidth]{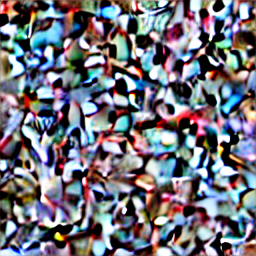}
                \end{subfigure}\hspace{-1mm}
                \begin{subfigure}{0.048\textwidth}
                    \centering \hfill
                    \includegraphics[width=\linewidth]{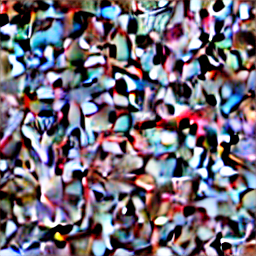}
                \end{subfigure}\hspace{-1mm}
                \begin{subfigure}{0.048\textwidth}
                    \centering \hfill
                    \includegraphics[width=\linewidth]{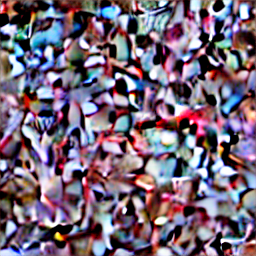}
                \end{subfigure}\hspace{-1mm}
                \begin{subfigure}{0.048\textwidth}
                    \centering \hfill
                    \includegraphics[width=\linewidth]{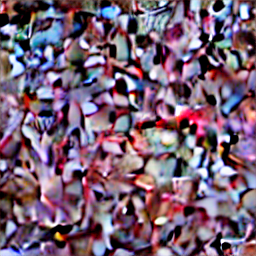}
                \end{subfigure}\hspace{-1mm}
                \begin{subfigure}{0.048\textwidth}
                    \centering \hfill
                    \adjustbox{cfbox=red 1.4pt -1.2pt}{\includegraphics[width=\linewidth]{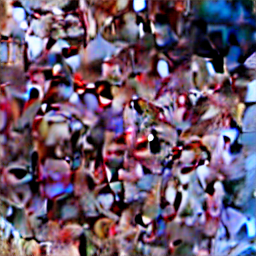}}
                \end{subfigure}\hspace{-1mm}
                \begin{subfigure}{0.048\textwidth}
                    \centering \hfill
                    \includegraphics[width=\linewidth]{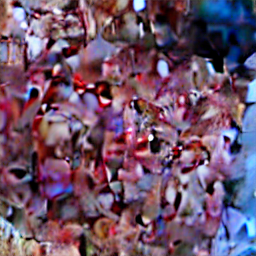}
                \end{subfigure}\hspace{-1mm}
                \begin{subfigure}{0.048\textwidth}
                    \centering \hfill
                    \includegraphics[width=\linewidth]{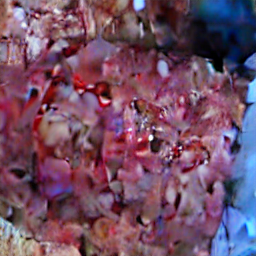}
                \end{subfigure}\hspace{-1mm}
                \begin{subfigure}{0.048\textwidth}
                    \centering \hfill
                    \includegraphics[width=\linewidth]{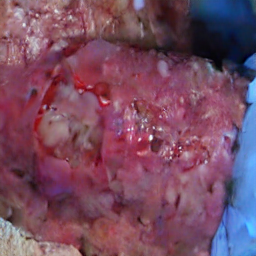}
                \end{subfigure}\hspace{-1mm}
                \begin{subfigure}{0.048\textwidth}
                    \centering \hfill
                    \includegraphics[width=\linewidth]{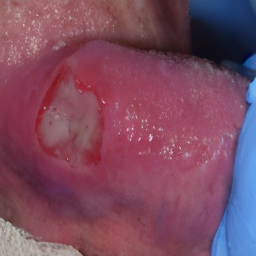}
                \end{subfigure}
            \end{minipage}
            \vspace{0.1mm}
        \end{minipage}
        \begin{minipage}{\linewidth}
         \begin{minipage}{0.15cm}\footnotesize{6}\end{minipage}
        \begin{minipage}{\linewidth}
            \begin{subfigure}{0.048\textwidth}
                \centering \hfill 
                \includegraphics[width=\linewidth]{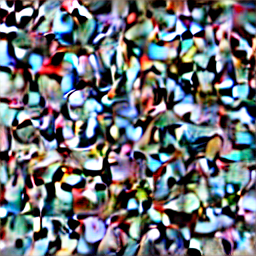}
            \end{subfigure}\hspace{-1mm}
            \begin{subfigure}{0.048\textwidth}
                \centering \hfill
                \includegraphics[width=\linewidth]{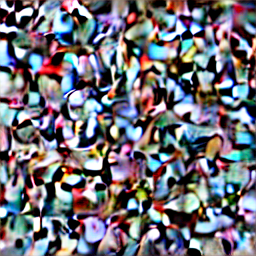}
            \end{subfigure}\hspace{-1mm}
            \begin{subfigure}{0.048\textwidth}
                \centering \hfill
                \includegraphics[width=\linewidth]{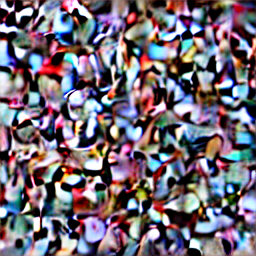}
            \end{subfigure}\hspace{-1mm}
            \begin{subfigure}{0.048\textwidth}
                \centering \hfill
                \includegraphics[width=\linewidth]{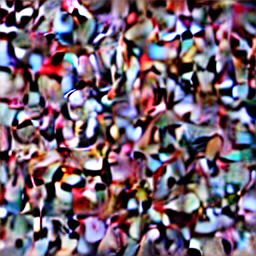}
            \end{subfigure}\hspace{-1mm}
            \begin{subfigure}{0.048\textwidth}
                \centering \hfill
                \includegraphics[width=\linewidth]{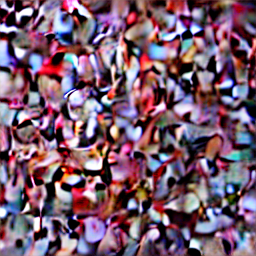}
            \end{subfigure}\hspace{-1mm}
            \begin{subfigure}{0.048\textwidth}
                \centering \hfill
                \includegraphics[width=\linewidth]{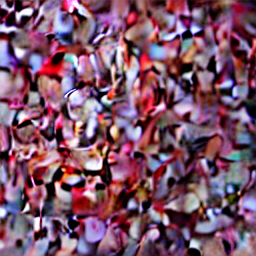}
            \end{subfigure}\hspace{-1mm}
            \begin{subfigure}{0.048\textwidth}
                \centering \hfill
                \adjustbox{cfbox=red 1.4pt -1.2pt}{\includegraphics[width=\linewidth]{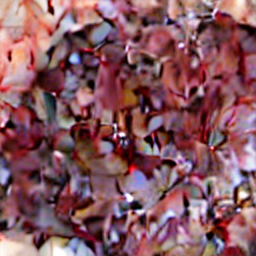}}
            \end{subfigure}\hspace{-1mm}
            \begin{subfigure}{0.048\textwidth}
                \centering \hfill
                \includegraphics[width=\linewidth]{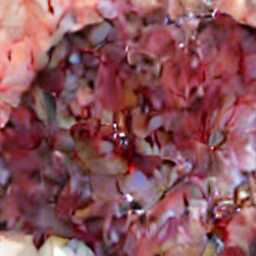}
            \end{subfigure}\hspace{-1mm}
            \begin{subfigure}{0.048\textwidth}
                \centering \hfill
                \includegraphics[width=\linewidth]{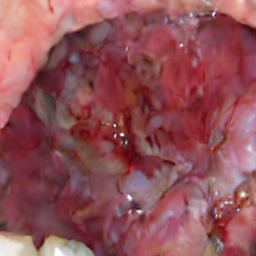}
            \end{subfigure}\hspace{-1mm}
            \begin{subfigure}{0.048\textwidth}
                \centering \hfill
                \includegraphics[width=\linewidth]{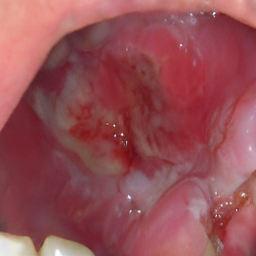}
            \end{subfigure}\hspace{-0.4mm}
            \begin{subfigure}{0.048\textwidth}
                \centering \hfill 
                \includegraphics[width=\linewidth]{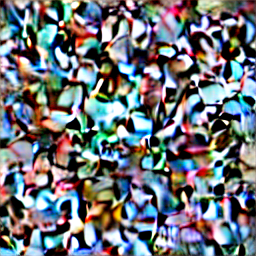}
            \end{subfigure}\hspace{-1mm}
            \begin{subfigure}{0.048\textwidth}
                \centering \hfill
                \includegraphics[width=\linewidth]{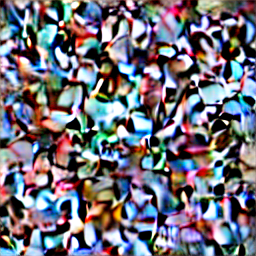}
            \end{subfigure}\hspace{-1mm}
            \begin{subfigure}{0.048\textwidth}
                \centering \hfill
                \includegraphics[width=\linewidth]{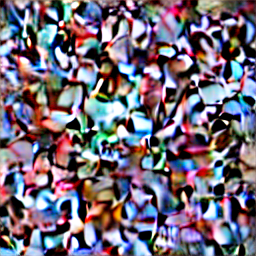}
            \end{subfigure}\hspace{-1mm}
            \begin{subfigure}{0.048\textwidth}
                \centering \hfill
                \includegraphics[width=\linewidth]{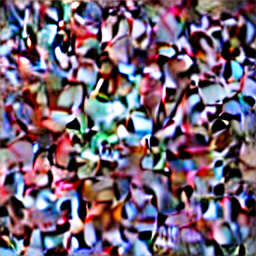}
            \end{subfigure}\hspace{-1mm}
            \begin{subfigure}{0.048\textwidth}
                \centering \hfill
                \includegraphics[width=\linewidth]{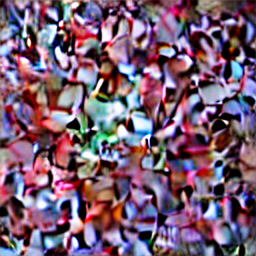}
            \end{subfigure}\hspace{-1mm}
            \begin{subfigure}{0.048\textwidth}
                \centering \hfill
                \includegraphics[width=\linewidth]{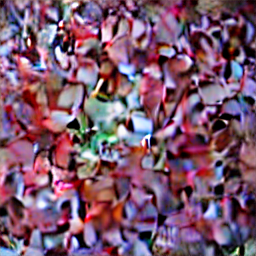}
            \end{subfigure}\hspace{-1mm}
            \begin{subfigure}{0.048\textwidth}
                \centering \hfill
                \adjustbox{cfbox=red 1.4pt -1.2pt}{\includegraphics[width=\linewidth]{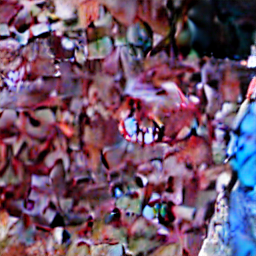}}
            \end{subfigure}\hspace{-1mm}
            \begin{subfigure}{0.048\textwidth}
                \centering \hfill
                \includegraphics[width=\linewidth]{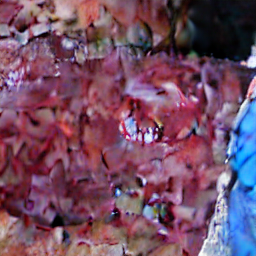}
            \end{subfigure}\hspace{-1mm}
            \begin{subfigure}{0.048\textwidth}
                \centering \hfill
                \includegraphics[width=\linewidth]{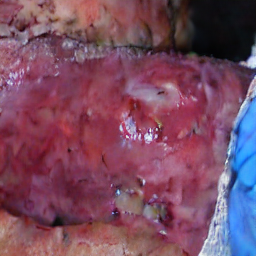}
            \end{subfigure}\hspace{-1mm}
            \begin{subfigure}{0.048\textwidth}
                \centering \hfill
                \includegraphics[width=\linewidth]{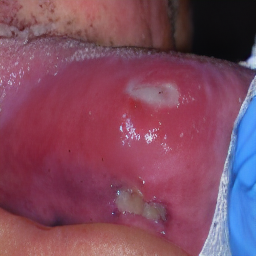}
            \end{subfigure}
        \end{minipage}
        \vspace{0.1mm}
    \end{minipage}
    \begin{minipage}{\linewidth}
         \begin{minipage}{0.15cm}\footnotesize{7}\end{minipage}
        \begin{minipage}{\linewidth}
            \begin{subfigure}{0.048\textwidth}
                \centering \hfill
                \includegraphics[width=\linewidth]{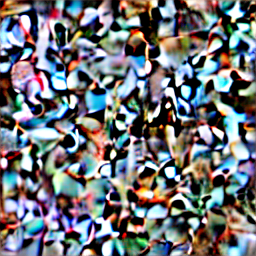}
            \end{subfigure}\hspace{-1mm}
            \begin{subfigure}{0.048\textwidth}
                \centering \hfill
                \includegraphics[width=\linewidth]{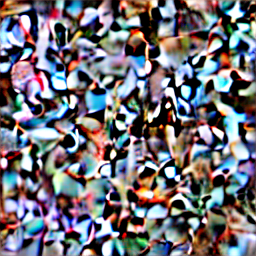}
            \end{subfigure}\hspace{-1mm}
            \begin{subfigure}{0.048\textwidth}
                \centering \hfill
                \includegraphics[width=\linewidth]{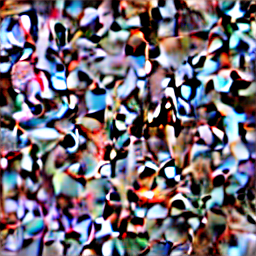}
            \end{subfigure}\hspace{-1mm}
            \begin{subfigure}{0.048\textwidth}
                \centering \hfill
                \includegraphics[width=\linewidth]{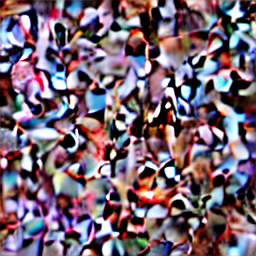}
            \end{subfigure}\hspace{-1mm}
            \begin{subfigure}{0.048\textwidth}
                \centering \hfill
                \includegraphics[width=\linewidth]{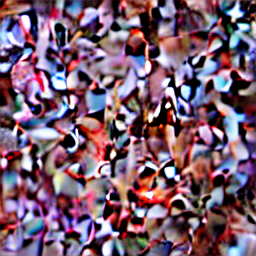}
            \end{subfigure}\hspace{-1mm}
            \begin{subfigure}{0.048\textwidth}
                \centering \hfill
                \includegraphics[width=\linewidth]{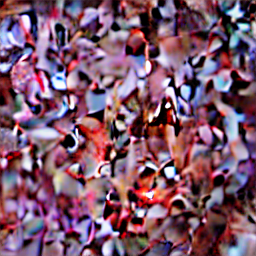}
            \end{subfigure}\hspace{-1mm}
            \begin{subfigure}{0.048\textwidth}
                \centering \hfill
                \includegraphics[width=\linewidth]{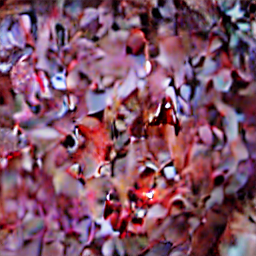}
            \end{subfigure}\hspace{-1mm}
            \begin{subfigure}{0.048\textwidth}
                \centering \hfill
                \adjustbox{cfbox=red 1.4pt -1.2pt}{\includegraphics[width=\linewidth]{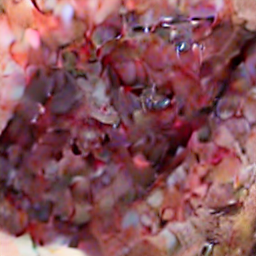}}
            \end{subfigure}\hspace{-1mm}
            \begin{subfigure}{0.048\textwidth}
                \centering \hfill
                \includegraphics[width=\linewidth]{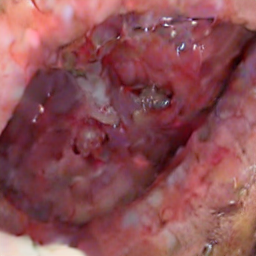}
            \end{subfigure}\hspace{-1mm}
            \begin{subfigure}{0.048\textwidth}
                \centering \hfill
                \includegraphics[width=\linewidth]{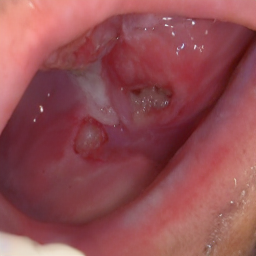}
            \end{subfigure}\hspace{-0.4mm}
            \begin{subfigure}{0.048\textwidth}
                \centering \hfill
                \includegraphics[width=\linewidth]{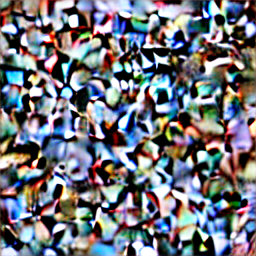}
            \end{subfigure}\hspace{-1mm}
            \begin{subfigure}{0.048\textwidth}
                \centering \hfill
                \includegraphics[width=\linewidth]{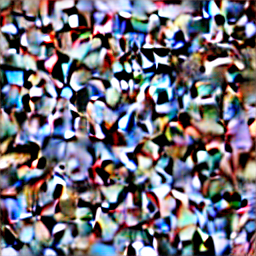}
            \end{subfigure}\hspace{-1mm}
            \begin{subfigure}{0.048\textwidth}
                \centering \hfill
                \includegraphics[width=\linewidth]{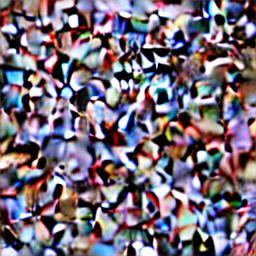}
            \end{subfigure}\hspace{-1mm}
            \begin{subfigure}{0.048\textwidth}
                \centering \hfill
                \includegraphics[width=\linewidth]{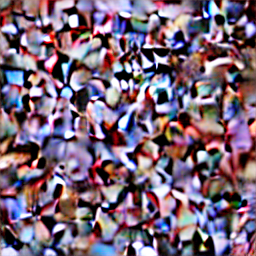}
            \end{subfigure}\hspace{-1mm}
            \begin{subfigure}{0.048\textwidth}
                \centering \hfill
                \includegraphics[width=\linewidth]{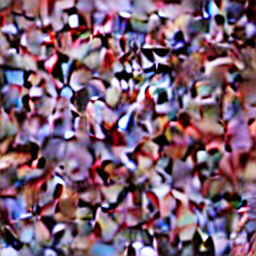}
            \end{subfigure}\hspace{-1mm}
            \begin{subfigure}{0.048\textwidth}
                \centering \hfill
                \includegraphics[width=\linewidth]{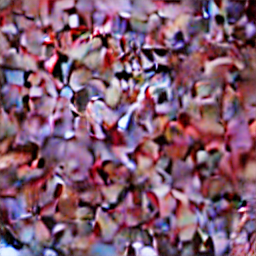}
            \end{subfigure}\hspace{-1mm}
            \begin{subfigure}{0.048\textwidth}
                \centering \hfill
                \includegraphics[width=\linewidth]{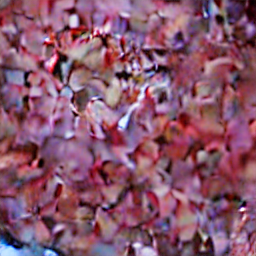}
            \end{subfigure}\hspace{-1mm}
            \begin{subfigure}{0.048\textwidth}
                \centering \hfill
                \adjustbox{cfbox=red 1.4pt -1.2pt}{\includegraphics[width=\linewidth]{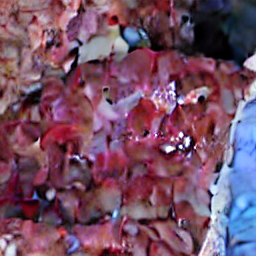}}
            \end{subfigure}\hspace{-1mm}
            \begin{subfigure}{0.048\textwidth}
                \centering \hfill
                \includegraphics[width=\linewidth]{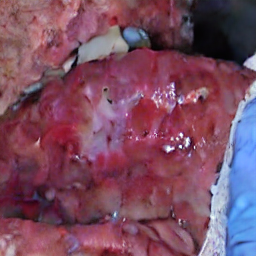}
            \end{subfigure}\hspace{-1mm}
            \begin{subfigure}{0.048\textwidth}
                \centering \hfill
                \includegraphics[width=\linewidth]{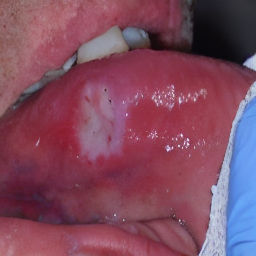}
            \end{subfigure}
        \end{minipage}
        \vspace{0.1mm}
    \end{minipage}
    \begin{minipage}{0.15cm}\footnotesize{8}\end{minipage}
    \begin{minipage}{\linewidth}
        \begin{subfigure}{0.048\textwidth}
            \centering \hfill
            \includegraphics[width=\linewidth]{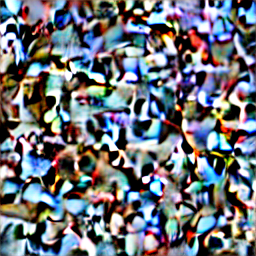}
        \end{subfigure}\hspace{-1mm}
        \begin{subfigure}{0.048\textwidth}
            \centering \hfill
            \includegraphics[width=\linewidth]{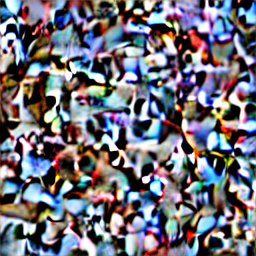}
        \end{subfigure}\hspace{-1mm}
        \begin{subfigure}{0.048\textwidth}
            \centering \hfill
            \includegraphics[width=\linewidth]{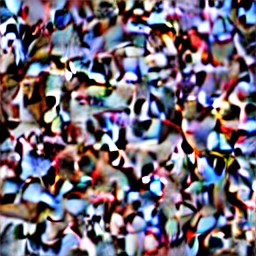}
        \end{subfigure}\hspace{-1mm}
        \begin{subfigure}{0.048\textwidth}
            \centering \hfill
            \includegraphics[width=\linewidth]{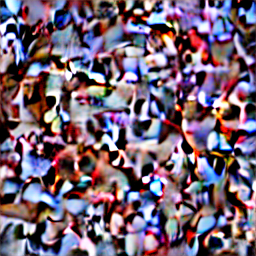}
        \end{subfigure}\hspace{-1mm}
        \begin{subfigure}{0.048\textwidth}
            \centering \hfill
            \includegraphics[width=\linewidth]{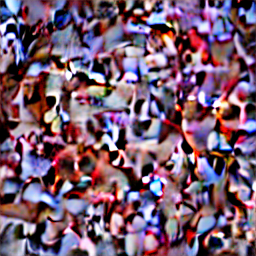}
        \end{subfigure}\hspace{-1mm}
        \begin{subfigure}{0.048\textwidth}
            \centering \hfill
            \includegraphics[width=\linewidth]{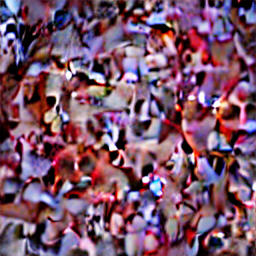}
        \end{subfigure}\hspace{-1mm}
        \begin{subfigure}{0.048\textwidth}
            \centering \hfill
            \includegraphics[width=\linewidth]{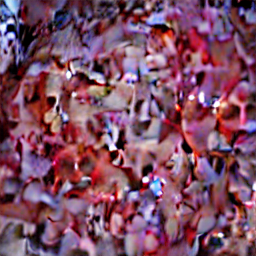}
        \end{subfigure}\hspace{-1mm}
        \begin{subfigure}{0.048\textwidth}
            \centering \hfill
            \includegraphics[width=\linewidth]{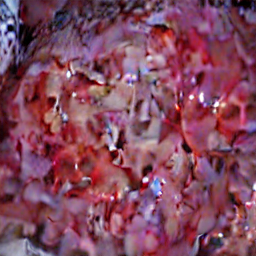}
        \end{subfigure}\hspace{-1mm}
        \begin{subfigure}{0.048\textwidth}
            \centering \hfill
            \adjustbox{cfbox=red 1.4pt -1.2pt}{\includegraphics[width=\linewidth]{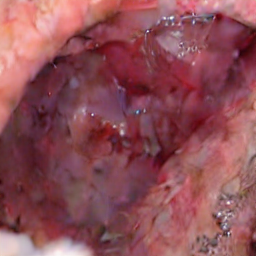}}
        \end{subfigure}\hspace{-1mm}
        \begin{subfigure}{0.048\textwidth}
            \centering \hfill
            \includegraphics[width=\linewidth]{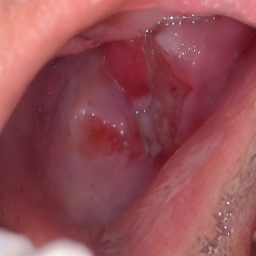}
        \end{subfigure}\hspace{-0.4mm}
        \begin{subfigure}{0.048\textwidth}
            \centering \hfill
            \includegraphics[width=\linewidth]{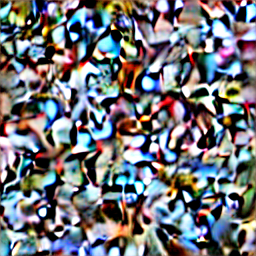}
        \end{subfigure}\hspace{-1mm}
        \begin{subfigure}{0.048\textwidth}
            \centering \hfill
            \includegraphics[width=\linewidth]{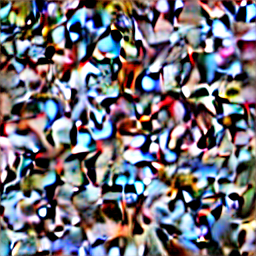}
        \end{subfigure}\hspace{-1mm}
        \begin{subfigure}{0.048\textwidth}
            \centering \hfill
            \includegraphics[width=\linewidth]{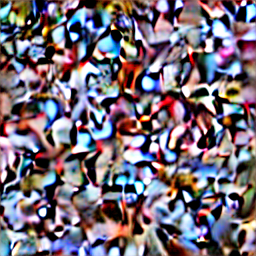}
        \end{subfigure}\hspace{-1mm}
        \begin{subfigure}{0.048\textwidth}
            \centering \hfill
            \includegraphics[width=\linewidth]{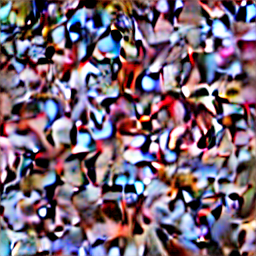}
        \end{subfigure}\hspace{-1mm}
        \begin{subfigure}{0.048\textwidth}
            \centering \hfill
            \includegraphics[width=\linewidth]{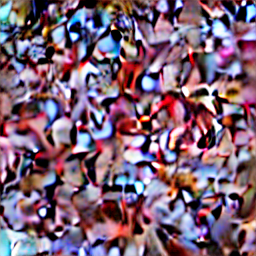}
        \end{subfigure}\hspace{-1mm}
        \begin{subfigure}{0.048\textwidth}
            \centering \hfill
            \includegraphics[width=\linewidth]{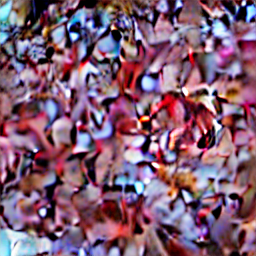}
        \end{subfigure}\hspace{-1mm}
        \begin{subfigure}{0.048\textwidth}
            \centering \hfill
            \includegraphics[width=\linewidth]{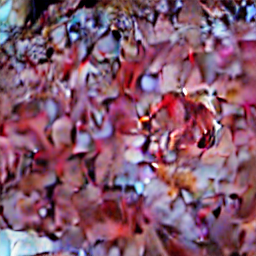}
        \end{subfigure}\hspace{-1mm}
        \begin{subfigure}{0.048\textwidth}
            \centering \hfill
            \includegraphics[width=\linewidth]{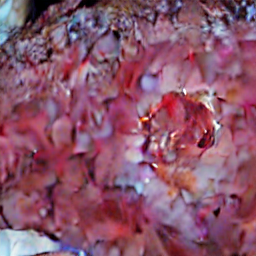}
        \end{subfigure}\hspace{-1mm}
        \begin{subfigure}{0.048\textwidth}
            \centering \hfill
            \adjustbox{cfbox=red 1.4pt -1.2pt}{\includegraphics[width=\linewidth]{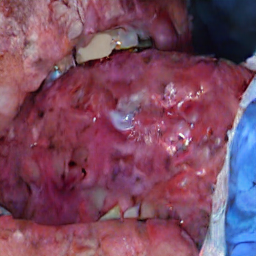}}
        \end{subfigure}\hspace{-1mm}
        \begin{subfigure}{0.048\textwidth}
            \centering \hfill
            \includegraphics[width=\linewidth]{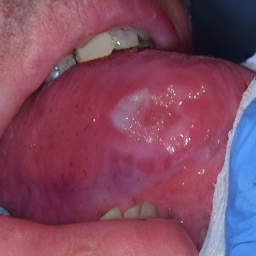}
        \end{subfigure}
        \vspace{0.4mm}
    \end{minipage}
    \begin{minipage}{0.15cm}\footnotesize{9}\end{minipage}
    \begin{minipage}{\linewidth}
        \begin{subfigure}{0.048\textwidth}
            \centering \hfill
            \includegraphics[width=\linewidth]{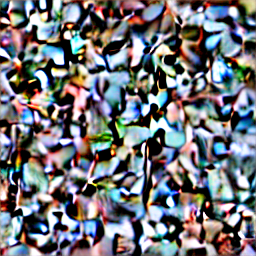}
            \centering \footnotesize{10}
        \end{subfigure}\hspace{-1mm}
        \begin{subfigure}{0.048\textwidth}
            \centering \hfill
            \includegraphics[width=\linewidth]{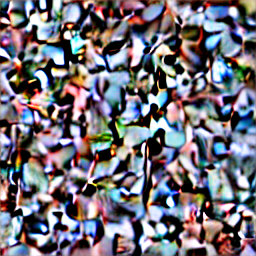}
            \centering \footnotesize{20}
        \end{subfigure}\hspace{-1mm}
        \begin{subfigure}{0.048\textwidth}
            \centering \hfill
            \includegraphics[width=\linewidth]{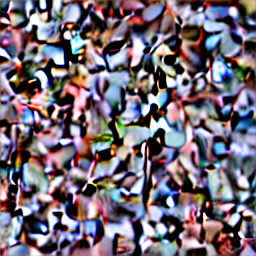}
            \centering \footnotesize{30}
        \end{subfigure}\hspace{-1mm}
        \begin{subfigure}{0.048\textwidth}
            \centering \hfill
            \includegraphics[width=\linewidth]{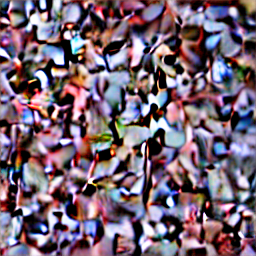}
            \centering \footnotesize{40}
        \end{subfigure}\hspace{-1mm}
        \begin{subfigure}{0.048\textwidth}
            \centering \hfill
            \includegraphics[width=\linewidth]{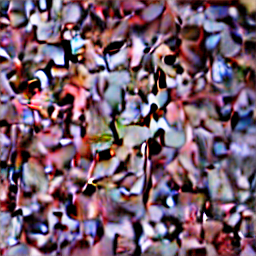}
            \centering \footnotesize{50}
        \end{subfigure}\hspace{-1mm}
        \begin{subfigure}{0.048\textwidth}
            \centering \hfill
            \includegraphics[width=\linewidth]{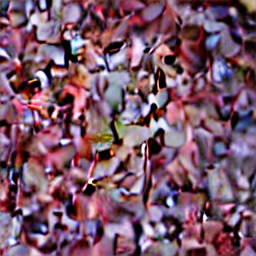}
            \centering \footnotesize{60}
        \end{subfigure}\hspace{-1mm}
        \begin{subfigure}{0.048\textwidth}
            \centering \hfill
            \includegraphics[width=\linewidth]{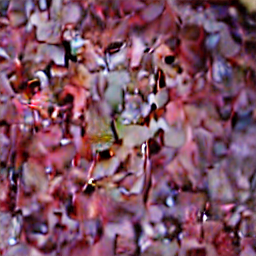}
            \centering \footnotesize{70}
        \end{subfigure}\hspace{-1mm}
        \begin{subfigure}{0.048\textwidth}
            \centering \hfill
            \includegraphics[width=\linewidth]{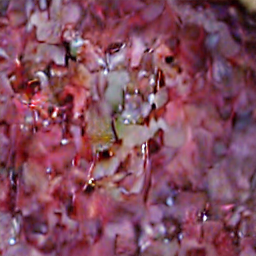}
            \centering \footnotesize{80}
        \end{subfigure}\hspace{-1mm}
        \begin{subfigure}{0.048\textwidth}
            \centering \hfill
            \includegraphics[width=\linewidth]{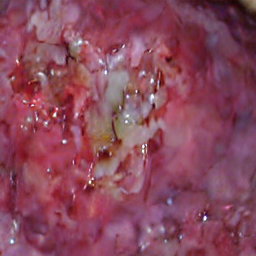}
            \centering \footnotesize{90}
        \end{subfigure}\hspace{-1mm}
        \begin{subfigure}{0.048\textwidth}
            \centering \hfill
            \adjustbox{cfbox=teal 1.4pt -1.2pt}{\includegraphics[width=\linewidth]{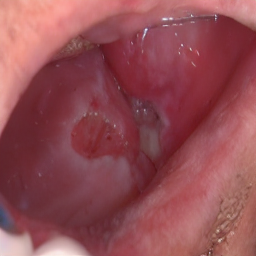}}
            \centering \footnotesize{100}
        \end{subfigure}\hspace{-0.4mm}
        \begin{subfigure}{0.048\textwidth}
            \centering \hfill
            \includegraphics[width=\linewidth]{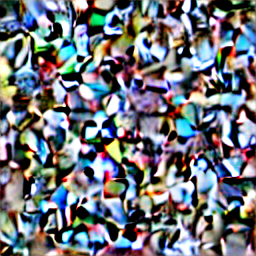}
            \centering \footnotesize{10}
        \end{subfigure}\hspace{-1mm}
        \begin{subfigure}{0.048\textwidth}
            \centering \hfill
            \includegraphics[width=\linewidth]{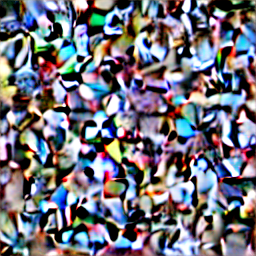}
            \centering \footnotesize{20}
        \end{subfigure}\hspace{-1mm}
        \begin{subfigure}{0.048\textwidth}
            \centering \hfill
            \includegraphics[width=\linewidth]{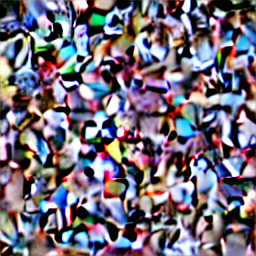}
            \centering \footnotesize{30}
        \end{subfigure}\hspace{-1mm}
        \begin{subfigure}{0.048\textwidth}
            \centering \hfill
            \includegraphics[width=\linewidth]{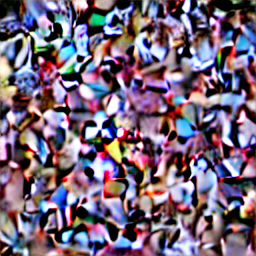}
            \centering \footnotesize{40}
        \end{subfigure}\hspace{-1mm}
        \begin{subfigure}{0.048\textwidth}
            \centering \hfill
            \includegraphics[width=\linewidth]{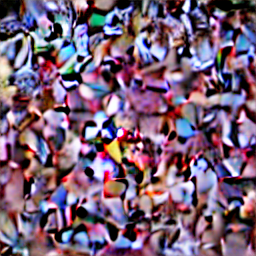}
            \centering \footnotesize{50}
        \end{subfigure}\hspace{-1mm}
        \begin{subfigure}{0.048\textwidth}
            \centering \hfill
            \includegraphics[width=\linewidth]{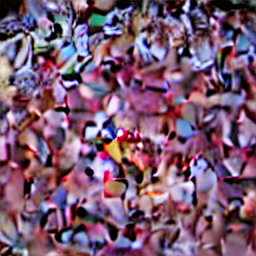}
            \centering \footnotesize{60}
        \end{subfigure}\hspace{-1mm}
        \begin{subfigure}{0.048\textwidth}
            \centering \hfill
            \includegraphics[width=\linewidth]{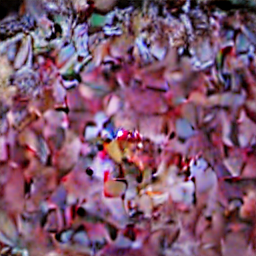}
            \centering \footnotesize{70}
        \end{subfigure}\hspace{-1mm}
        \begin{subfigure}{0.048\textwidth}
            \centering \hfill
            \includegraphics[width=\linewidth]{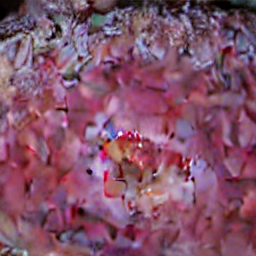}
            \centering \footnotesize{80}
        \end{subfigure}\hspace{-1mm}
        \begin{subfigure}{0.048\textwidth}
            \centering \hfill
            \includegraphics[width=\linewidth]{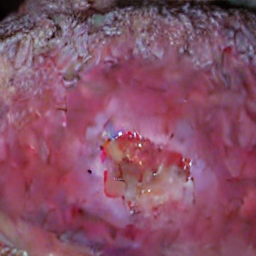}
            \centering \footnotesize{90}
        \end{subfigure}\hspace{-1mm}
        \begin{subfigure}{0.048\textwidth}
            \centering \hfill
            \adjustbox{cfbox=teal 1.4pt -1.2pt}{\includegraphics[width=\linewidth]{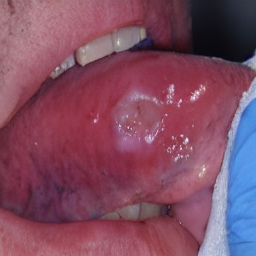}}
            \centering \footnotesize{100}
        \end{subfigure}
        
    \end{minipage} 
    \end{minipage}
    
    \vspace{1.5mm}
    \centering \footnotesize{\hspace{0.5cm} Step \hspace{8.3cm} Step}
    \vspace{1.5mm}
    
    \centering \small{\hspace{0.5cm} (a) \hspace{8.3cm} (b)}

    \caption{ 
    Two examples of diffusion processes using SD(txt+img) with conditioning mechanisms based on text and image for \textit{neoplastic} and \textit{traumatic} lesions in (a) and (b) from PhotoMOCI, respectively. In each example, ten denoising processes with image conditioning performed at different steps. The \textcolor{red}{red}-bordered images indicate where the conditioning image is introduced. In the last row, with \textcolor{teal}{water-green} borders, the original conditioning image injected at the 100th step, overriding earlier ones.}
    \label{fig:diffusion-process-neoplastic}
\end{figure*}

For the generation of oral lesions, all synthetic samples $(x_j,y_j) \in \mathcal{D}_{synth}$ are generated from $g(.)$ using conditioning information $c$ (whether a string, a vector, or a single value) referring to class labels $y_i$. Therefore, both $x_j$ and $y_j$ result as a function of $c$.
Specifically, we employ two state-of-the-art generative models. The first is Stable Diffusion \cite{rombachetal2022latent}, a latent diffusion framework fine-tuned through the LoRA domain adaptation technique. The second is StyleGAN3 \cite{karras2021alias}, which introduces an alias-free generator replacing conventional upsampling and filtering operations with low-pass filters.
For each generative model, we explore two conditioning strategies, leading to four distinct synthetic generation setups:

\begin{itemize}
    \item \textbf{SD(txt)}: Stable Diffusion conditioned on text. Lesion descriptions are first encoded into embeddings using CLIP \cite{clip}. These embeddings are then combined with the latent representations during the denoising process by summing them, allowing the model to generate images that closely align with the provided textual descriptions.

    \item \textbf{SD(txt+img)}: Stable Diffusion conditioned on both text and images. Similar to the previous setup, textual embeddings are fused with image embeddings. However, instead of executing the full denoising trajectory from pure noise, the process is initialized from a conditioning image and proceeds for a limited number of denoising steps. This approach leverages the input image as a structural prior. Figure~\ref{fig:diffusion-process-neoplastic} illustrates this text–image conditioning scheme for a neoplastic lesion, where the conditioning image is injected at different stages of the denoising process.
    
    \item \textbf{SG3(lbl)}: StyleGAN3 conditioned on class labels. Here, labels are one-hot encoded and projected into a learned embedding space. The resulting embeddings are then concatenated with the generator’s latent codes, enabling label-aware synthesis.
    
    \item \textbf{AC-SG3}: Auxiliary Classifier StyleGAN3 conditioned on class labels. In this configuration, conditioning is enforced through the discriminator, which is augmented with an auxiliary classifier head that predicts the class label in addition to the adversarial objective, thereby encouraging the generator to generate real images that are not only indistinguishable from real images but also belong to the specified class.
\end{itemize}

Figure~\ref{fig:diffusion-process-neoplastic} presents two examples of diffusion processes in the SD(txt+img) setup. The model is a Stable Diffusion architecture finetuned on the PhotoMOCI dataset using LoRA over 100 denoising steps. In each case, the same conditioning image, belonging to the neoplastic Figure~\ref{fig:diffusion-process-neoplastic}(a) and traumatic Figure~\ref{fig:diffusion-process-neoplastic}(b) classes, is injected at different stages of the denoising process to guide generation. The specific step at which the conditioning image is introduced is highlighted with a \textcolor{red}{red} border. The final image in the last row, marked with a \textcolor{teal}{water-green} border, displays the original conditioning image injected at the final denoising step, which overrides the preceding steps and fully dominates the output. 

\subsection{Synthetic Image Filter}\label{sec:synthetic_image_selection}


While generative models can synthesize large volumes of data, the utility of such samples is not always significant. Poorly generated samples, characterized by artifacts or semantic misalignment with the target class, introduce label noise that can degrade the performance of downstream classifiers. To mitigate this, we propose the \textbf{Synthetic Image Filter (SIF)}, a mechanism designed to select only those samples $(x_j,y_j)$ from $\mathcal{D}_{synth}$ that enhance the generalization of the final oral cancer classifier.
As illustrated in Figure~\ref{fig:oral-synthetic-scema} (\textcolor{LimeGreen}{green} dotted box), SIF implements a filtering mechanism based on two auxiliary models trained for complementary purposes: a \textit{Synthetic Proxy Classifier} (SPC), denoted as $f_{SPC}$, and a \textit{Synthetic Image Detector} (SID), denoted as $f_{SID}$.


The $f_{SPC}$ is a classification model trained exclusively on synthetic images to solve the same downstream classification task as the oral lesion classifier detailed in the next section.  Intuitively, this model evaluates whether a synthetic image $x_j$ exhibits visual patterns coherent with its intended label $y_j$. 
On the other hand, SID ($f_{SID}$) is a binary classifier trained to discriminate real images $x_i \in \mathcal{D}_{real}$ from synthetic images $x_j \in \mathcal{D}_{synth}$.
Its objective is to capture the distributional shift between synthetic and real images, providing a machine-based measure of realism for new images $\hat{x}$:
\begin{equation}
    f_{SID}(\hat{x}) \in [0, 1]
\end{equation}
where a value closer to $1$ indicates high confidence that the sample is synthetic.
In this context, ``realism'' refers to machine indistinguishability from the real-image distribution as estimated by SID, and should not be interpreted as clinical realism or diagnostic plausibility assessed by clinicians. Although these notions may be related, since clinically implausible artifacts can make synthetic samples easier to detect, SID evaluates realism only from the perspective of the trained discriminator.
Finally, SIF combines the predictions of these two auxiliary models to select synthetic images for data augmentation. Specifically, a sample is retained only if it is \textit{correctly classified by SPC}, indicating class label semantic consistency, and \textit{misclassified by SID}, suggesting strong visual similarity to real images. The resulting Filtered Dataset $\mathcal{D}_{filt}$ is formally defined as $\mathcal{D}_{filt} = \{ (x_j, y_j) \in \mathcal{D}_{synth} \mid f_{SPC}(x_j) = y_j \land f_{SID}(x_j) < \tau \}$, where $\tau$ is a decision threshold (we adopt $\tau$=$0.5$). The selected samples are subsequently combined with the original small real dataset to form the final training set $\mathcal{D}_{real} \cup \mathcal{D}_{filt}$.

\subsection{Training the oral lesion classifier}\label{sec:training_oral_cancer_classifier}

In the last stage, the oral lesion classifier is trained under different data configurations to assess the impact of synthetic data and the proposed filtering strategy. Specifically, the classifier is trained using one of the following input datasets: 

\begin{itemize}
    \item the original small real dataset $\mathcal{D}_{real}$, which may generate a classifier that generalizes poorly because it is trained on a limited amount of data; 
    \item the original dataset augmented with the full Synthetic Dataset $\mathcal{D}_{real} \cup \mathcal{D}_{synth}$, which may produce a classifier with performance drops due to the uncontrolled integration of noisy synthetic images;
    \item the original dataset augmented with the Filtered Dataset $\mathcal{D}_{real} \cup \mathcal{D}_{filt}$, which produce a more robust classifier.
\end{itemize}

\section{Experiments and results}\label{sec:experiments}

The experiment section is organized into two parts:  
The first part, presented in Section \ref{sec:experiment-synthetic-generation}, describes how we finetune different generative architectures and reports experiments on synthetic image generation. Then, we describe how we develop the auxiliary classifiers composing SIF and how it is used to train oral lesion classifiers under different synthetic data augmentation setups in Sections \ref{sec:SIF_experiments} and \ref{sec:experiment-recognition}, respectively. Finally, we conduct an ablation study to understand the impact of the different SIF components on the overall filter. 
The experiments are implemented in PyTorch 1.10.2 (CUDA 11.3) and are publicly available on GitHub at \cite{github-repo}. They are run on NVIDIA A100 GPUs with 32GB of memory. 

\textbf{Datasets}. We evaluate our approach on PhotoMOCI for the multi-class problem, and on KOCD (version 1, released under Apache 2.0)~\cite{MOHD_ZAID_RASHID_oral_cancer_2024} for the binary problem, since they represent the largest and most recent datasets available in this field.
For experimental consistency and reproducibility, all experiments on PhotoMOCI follow the official dataset partition, consisting of 490, 105, and 105 images for the training, validation, and test sets, available at \cite{marco_parola_2025}. 
The KOCD dataset consists of 950 images labeled according to a binary classification setup: 450 cancer and 500 non-cancer. The images are collected from 950 patients, with each patient contributing one image; therefore, no near-duplicate images are present in the dataset.
Since KOCD is released without predefined partitions, we define a hold-out 70\%-15\%-15\% split, resulting in 665 (321 cancer and 344 non-cancer), 142 (64 cancer and 78 non-cancer), and 143 (66 cancer and 77 non-cancer) images for the training, validation, and test sets, respectively. The split is generated with a fixed random seed and is therefore reproducible across all experiments \cite{github-repo}. For both PhotoMOCI and KOCD, the same splits are consistently used throughout all the experiments.

\subsection{Synthetic Image Generation}\label{sec:experiment-synthetic-generation}

\textbf{Configuration Setup}. 
We evaluated four distinct synthetic image generation setups on 100 samples per class for each epoch against the real data.
Alongside such generation setups, we established a traditional augmentation baseline consisting of both geometric and photometric transformations: random flipping, color jittering, and affine transformations (rotations $\pm10^{\circ}$, translations $\pm15\%$, scaling 85-115\%, and shearing $\pm5^{\circ}$). The quality of all synthetic sets was benchmarked against the original, non-augmented data. Generative models were finetuned on the training set and conditioned with dataset-specific prompts: "\texttt{a medical photograph of an 'x' lesion}" for PhotoMOCI, where \texttt{'x'} is the class label (aphthous, traumatic, neoplastic), and "\texttt{a medical photograph of a 'x' tissue}" for KOCD, where \texttt{'x'} was either \texttt{cancerous} or \texttt{healthy} (cancer and non-cancer classes). 
Image-conditional setups in the case of SD(txt+img), used images from the training set. Since image conditioning can occur at different stages of the diffusion process, as shown in Figure \ref{fig:diffusion-process-neoplastic}, we empirically identified a range of steps that balance fidelity to the conditioning image with sufficient diversity in the generated output. Specifically, it is applied at a random step between steps 55 and 75 of the 100-step diffusion process.

For the training of our conditional and AC-StyleGAN3 variants, we adopted the `stylegan3-t` configuration and trained the models from scratch. With a batch size of 16 images, the training process stabilized with R1 regularization and a $\gamma$ coefficient of 0.6. Distinct learning rates were set for the generator ($2.5 \times 10^{-3}$) and the discriminator ($2.0 \times 10^{-3}$). To prevent discriminator overfitting, we employed Adaptive Discriminator Augmentation (ADA) alongside real-time horizontal mirroring.
For the training of our latent diffusion models, we fine-tuned the Stable Diffusion v1.5 model using Low-Rank Adaptation (LoRA). The model was trained for a maximum of 200 epochs with a batch size of 32. We monitored the validation loss, `val/loss`, to save the top three checkpoints. An 8-bit Adam optimizer was employed with a learning rate of $1.0 \times 10^{-4}$ for the U-Net and $0.5 \times 10^{-4}$ for the text encoder, using a weight decay of $1.0 \times 10^{-2}$. The LoRA configuration featured a rank of 16, an alpha of 32, and a dropout of 0.15, with adapters applied to the attention and projection modules of the U-Net and text encoder, respectively. Training was optimized using `bf16-mixed` precision, 4-bit quantization.

\textbf{Metrics}. To evaluate the performance of generative models, we employ different established state-of-the-art metrics \cite{NEURIPS2023_1f5c5cd0}. Such evaluation metrics can be categorized into two main groups. The first comprises distance-based metrics, including the Fréchet Inception Distance (FID) \cite{heusel2017gans} and Kernel Inception Distance (KID) \cite{binkowski2018demystifying} scores, which quantify the similarity between the distributions of generated and real data. The second group includes quality-based metrics that assess the fidelity and variability of the class labels predicted by a pre-trained classifier. This includes Precision and Recall \cite{kynkaanniemi2019improved}, and Coverage \cite{naeem2020reliable} scores.

\begin{table}[b]
\footnotesize
\centering
\caption{Synthetic Image Generation results on PhotoMOCI.}
\label{tab:metriche_generative_PhotoMOCI}
\begin{tabular}{lc @{\hspace{4.7mm}} c @{\hspace{4.7mm}}c @{\hspace{4.7mm}}c@{\hspace{4.7mm}} c@{\hspace{4.7mm}}c @{\hspace{4.7mm}} c}
\toprule
\textbf{\makecell[l]{Synthetic img.\\gen. setup}} & \textbf{FID$\downarrow$} & \textbf{KID$\downarrow$} & \textbf{Prec.$\uparrow$} & \textbf{Recall$\uparrow$} & \textbf{Cov.$\uparrow$} \\
\midrule
Tr. aug.  & 230.30 & 0.126 & 0.317 & 0.183 & 0.058  \\
SD(txt)  & 223.92 & 0.079 & 0.149 & \textbf{0.660} & 0.191 \\
SD(txt+img)  & \textbf{179.24} & \textbf{0.072}  & \textbf{0.543} & 0.319 & \textbf{0.521} \\
SG3(lbl)  & 195.77 & 0.099 &  0.404 & 0.234 & 0.287 \\
AC-SG3  & 290.42 & 0.230 & 0.426 & 0.011 & 0.032 \\
\bottomrule
\end{tabular}%
\end{table}

\begin{table}[t]
\footnotesize
\centering
\caption{Synthetic Image Generation results on KOCD.}
\label{tab:metriche_generative_KODC}
\begin{tabular}{lc @{\hspace{4.7mm}} c @{\hspace{4.7mm}}c @{\hspace{4.7mm}}c@{\hspace{4.7mm}} c@{\hspace{4.7mm}}c @{\hspace{4.7mm}} c}
\toprule
\textbf{\makecell[l]{Synthetic img.\\gen. setup}} & \textbf{FID$\downarrow$} & \textbf{KID$\downarrow$} & \textbf{Prec.$\uparrow$} & \textbf{Recall$\uparrow$} & \textbf{Cov.$\uparrow$} \\
\midrule
Tr. aug. & 262.91 & 0.126 & 0.284 & 0.632 & 0.242 \\
SD(txt)  & 225.38 & 0.074  & 0.273 & \textbf{0.762} & 0.231 \\
SD(txt+img)  & \textbf{143.87} & \textbf{0.034} & 0.867 & 0.573 & \textbf{0.888}\\
SG3(lbl)  & 169.57 & 0.062 & \textbf{0.888} & 0.112 & 0.748 \\
AC-SG3     & 319.78 & 0.245 & 0.035 & 0.035 & 0.042 \\
\bottomrule
\end{tabular}%
\end{table}

\textbf{Results}. Tables \ref{tab:metriche_generative_PhotoMOCI} and \ref{tab:metriche_generative_KODC} report the synthetic image generation task results on PhotoMOCI and KOCD, respectively. Across both datasets, consistent trends emerge: SD(txt-img) generally outperforms the other approaches in most metrics, except Recall on PhotoMOCI and both Precision and Recall on KOCD. On PhotoMOCI, SD(txt) achieves higher Recall, indicating that, in the context of diffusion models, text-only conditioning leads to greater diversity in the generated samples, effectively acting as a “general-purpose” generator. However, this diversity negatively affects the quality, as indicated by its low precision value, meaning that many of the generated images are not realistic. On KOCD, SD(txt) again surpasses SD(txt-img) in Recall, while SG3(lbl) achieves the best Precision, with its capacity to generate images that, from the perspective of a pre-trained classifier, closely reflect the label distributions observed in real samples.

\begin{figure*}[t]
    \centering
    \begin{minipage}{\textwidth}

    \centering \footnotesize{\hspace{0.1cm} StableDiffusion (txt) \hspace{4.9cm} StableDiffusion (txt+img)} \vspace{0.5mm}

    \begin{minipage}{\linewidth}
        \begin{minipage}{0.15cm}\rotatebox{90}{\footnotesize{traumatic}}\end{minipage}
        \begin{minipage}{\linewidth}
            \begin{subfigure}{0.091\textwidth}
                \centering \hfill 
                \includegraphics[width=\linewidth]{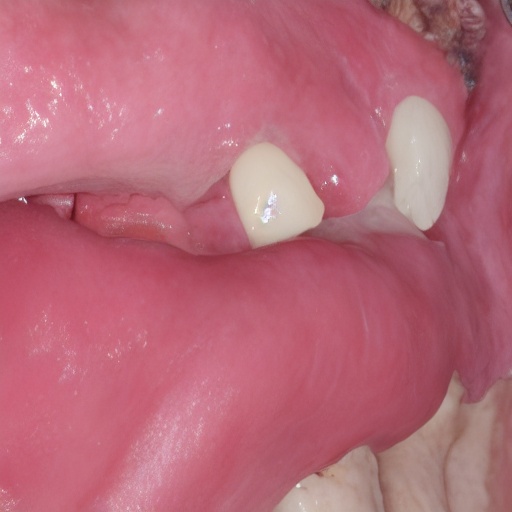}
            \end{subfigure}
            \begin{subfigure}{0.091\textwidth}
                \centering \hfill
                \includegraphics[width=\linewidth]{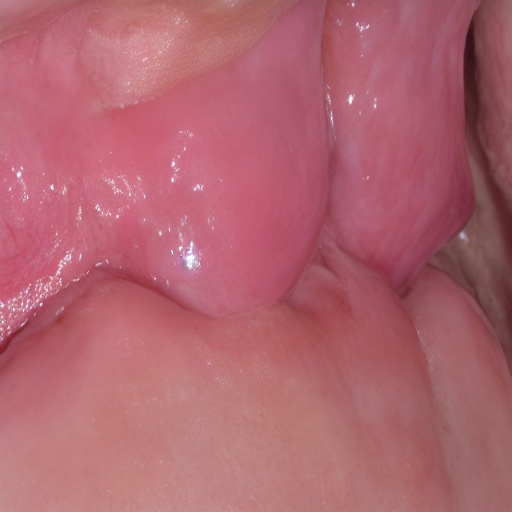}
            \end{subfigure}
            \begin{subfigure}{0.091\textwidth}
                \centering \hfill
                \includegraphics[width=\linewidth]{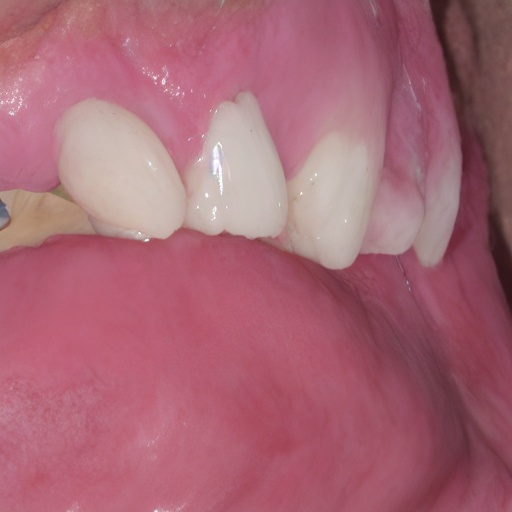}
            \end{subfigure}
            \begin{subfigure}{0.091\textwidth}
                \centering \hfill
                \includegraphics[width=\linewidth]{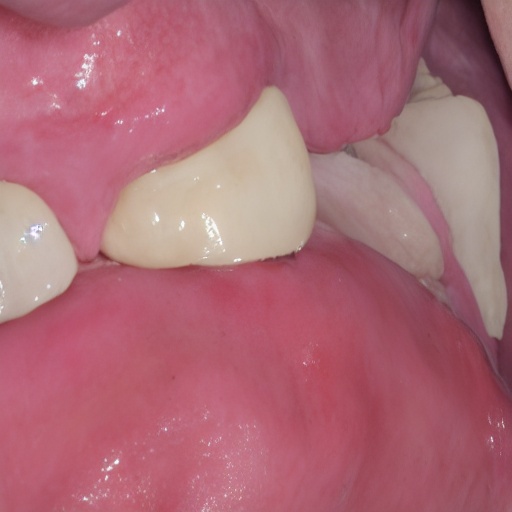}
            \end{subfigure}
            \begin{subfigure}{0.091\textwidth}
                \centering \hfill
                \includegraphics[width=\linewidth]{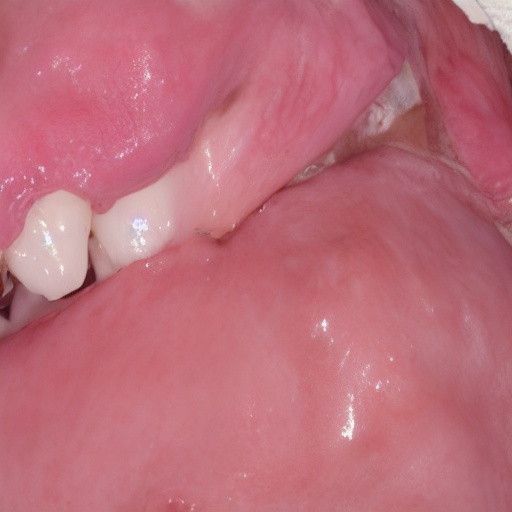}
            \end{subfigure}
            \hspace{0.1mm}            
            \begin{subfigure}{0.091\textwidth}
                \centering \hfill
                \includegraphics[width=\linewidth]{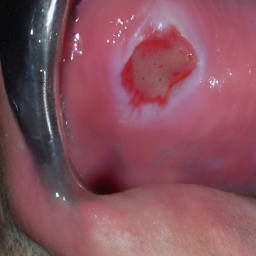}
            \end{subfigure}
            \begin{subfigure}{0.091\textwidth}
                \centering \hfill
                \includegraphics[width=\linewidth]{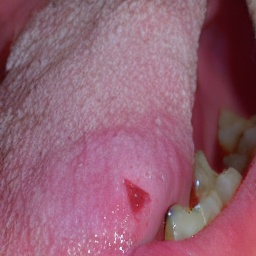}
            \end{subfigure}
            \begin{subfigure}{0.091\textwidth}
                \centering \hfill
                \includegraphics[width=\linewidth]{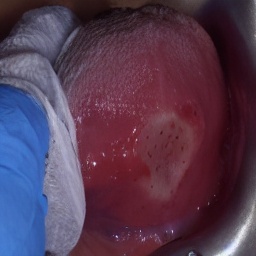}
            \end{subfigure}
            \begin{subfigure}{0.091\textwidth}
                \centering \hfill
                \includegraphics[width=\linewidth]{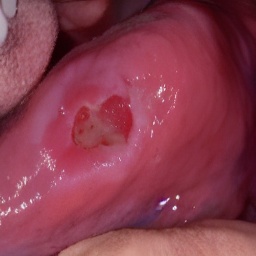}
            \end{subfigure}
            \begin{subfigure}{0.091\textwidth}
                \centering \hfill
                \includegraphics[width=\linewidth]{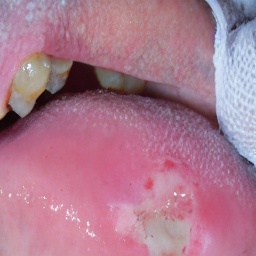}
            \end{subfigure}
        \end{minipage}
    \end{minipage}
    \vspace{0.5mm}

    \begin{minipage}{\linewidth}
         \begin{minipage}{0.15cm}\rotatebox{90}{\footnotesize{aphthous}}\end{minipage}
        \begin{minipage}{\linewidth}
            \begin{subfigure}{0.091\textwidth}
                \centering \hfill 
                \includegraphics[width=\linewidth]{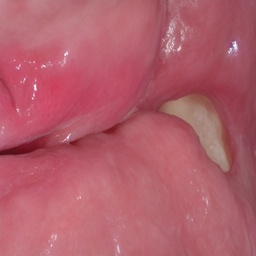}
            \end{subfigure}
            \begin{subfigure}{0.091\textwidth}
                \centering \hfill
                \includegraphics[width=\linewidth]{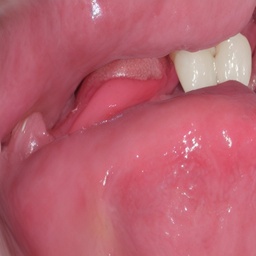}
            \end{subfigure}
            \begin{subfigure}{0.091\textwidth}
                \centering \hfill
                \includegraphics[width=\linewidth]{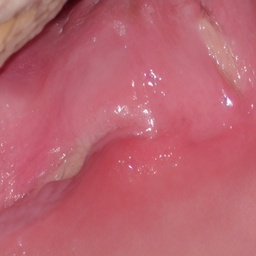}
            \end{subfigure}
            \begin{subfigure}{0.091\textwidth}
                \centering \hfill
                \includegraphics[width=\linewidth]{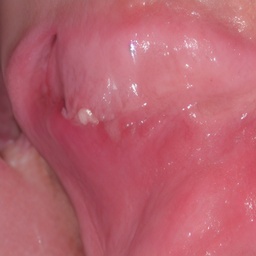}
            \end{subfigure}
            \begin{subfigure}{0.091\textwidth}
                \centering \hfill
                \includegraphics[width=\linewidth]{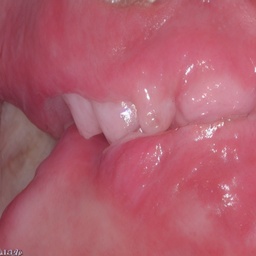}
            \end{subfigure}
            \hspace{0.1mm}
            \begin{subfigure}{0.091\textwidth}
                \centering \hfill
                \includegraphics[width=\linewidth]{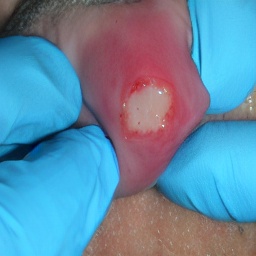}
            \end{subfigure}
            \begin{subfigure}{0.091\textwidth}
                \centering \hfill
                \includegraphics[width=\linewidth]{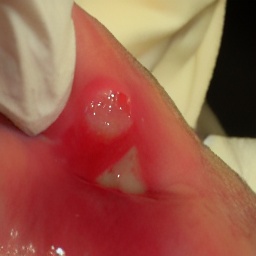}
            \end{subfigure}
            \begin{subfigure}{0.091\textwidth}
                \centering \hfill
                \includegraphics[width=\linewidth]{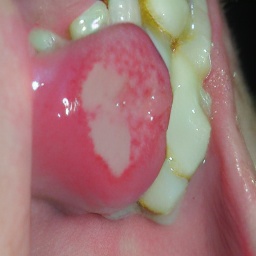}
            \end{subfigure}
            \begin{subfigure}{0.091\textwidth}
                \centering \hfill
                \includegraphics[width=\linewidth]{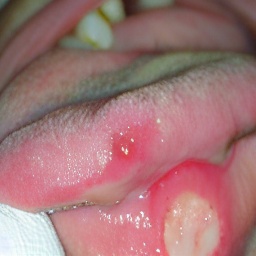}
            \end{subfigure}
            \begin{subfigure}{0.091\textwidth}
                \centering \hfill
                \includegraphics[width=\linewidth]{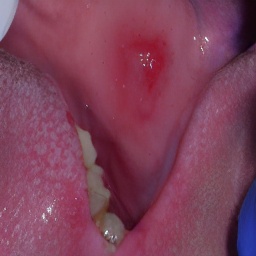}
            \end{subfigure}
        \end{minipage}
    \end{minipage}
    \vspace{0.5mm}

    \begin{minipage}{\linewidth}
         \begin{minipage}{0.15cm}\rotatebox{90}{\footnotesize{neoplastic}}\end{minipage}
         \begin{minipage}{\linewidth}
            \begin{subfigure}{0.091\textwidth}
                \centering \hfill 
                \includegraphics[width=\linewidth]{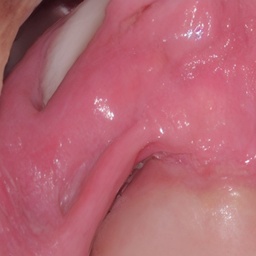}
            \end{subfigure}
            \begin{subfigure}{0.091\textwidth}
                \centering \hfill
                \includegraphics[width=\linewidth]{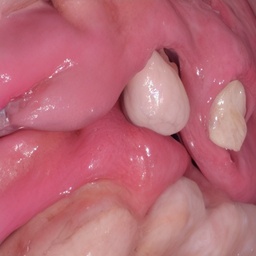}
            \end{subfigure}
            \begin{subfigure}{0.091\textwidth}
                \centering \hfill
                \includegraphics[width=\linewidth]{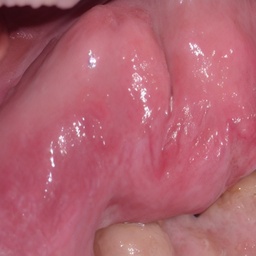}
            \end{subfigure}
            \begin{subfigure}{0.091\textwidth}
                \centering \hfill
                \includegraphics[width=\linewidth]{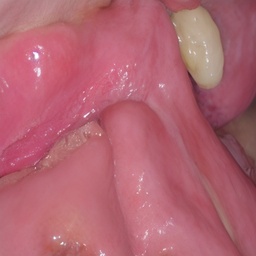}
            \end{subfigure}
            \begin{subfigure}{0.091\textwidth}
                \centering \hfill
                \includegraphics[width=\linewidth]{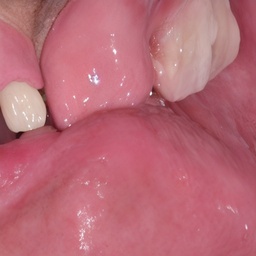}
            \end{subfigure}
            \hspace{0.1mm}
            \begin{subfigure}{0.091\textwidth}
                \centering \hfill
                \includegraphics[width=\linewidth]{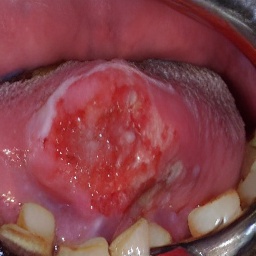}
            \end{subfigure}
            \begin{subfigure}{0.091\textwidth}
                \centering \hfill
                \includegraphics[width=\linewidth]{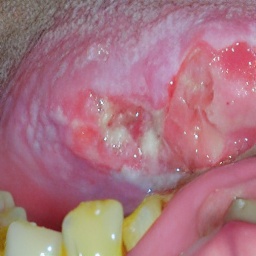}
            \end{subfigure}
            \begin{subfigure}{0.091\textwidth}
                \centering \hfill
                \includegraphics[width=\linewidth]{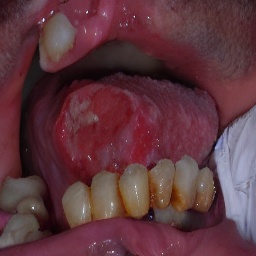}
            \end{subfigure}
            \begin{subfigure}{0.091\textwidth}
                \centering \hfill
                \includegraphics[width=\linewidth]{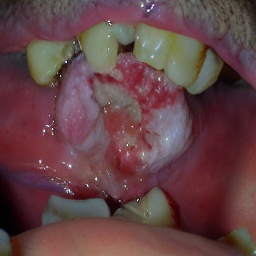}
            \end{subfigure}
            \begin{subfigure}{0.091\textwidth}
                \centering \hfill
                \includegraphics[width=\linewidth]{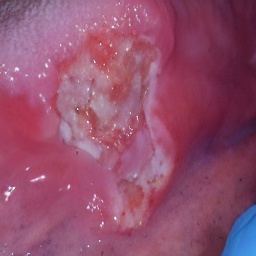}
            \end{subfigure}
        \end{minipage}
    \end{minipage}

    \vspace{2mm}

    \centering \hspace{1.cm}\footnotesize{StyleGAN3 (lbl) \hspace{4.4cm} 
    AuxiliaryClassifier-StyleGAN3}  \vspace{0.5mm}
    
    \begin{minipage}{\linewidth}
        \begin{minipage}{0.15cm}\rotatebox{90}{\footnotesize{traumatic}}\end{minipage}
        \begin{minipage}{\linewidth}
            \begin{subfigure}{0.091\textwidth}
                \centering \hfill 
                \includegraphics[width=\linewidth]{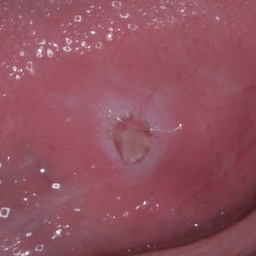}
            \end{subfigure}
            \begin{subfigure}{0.091\textwidth}
                \centering \hfill
                \includegraphics[width=\linewidth]{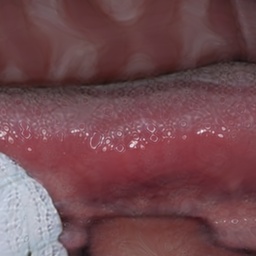}
            \end{subfigure}
            \begin{subfigure}{0.091\textwidth}
                \centering \hfill
                \includegraphics[width=\linewidth]{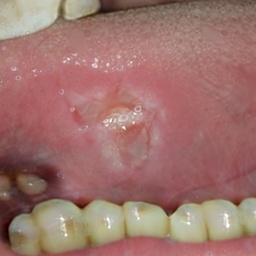}
            \end{subfigure}
            \begin{subfigure}{0.091\textwidth}
                \centering \hfill
                \includegraphics[width=\linewidth]{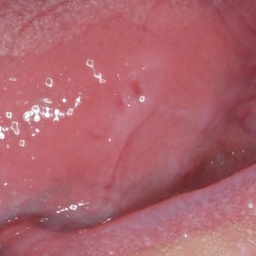}
            \end{subfigure}
            \begin{subfigure}{0.091\textwidth}
                \centering \hfill
                \includegraphics[width=\linewidth]{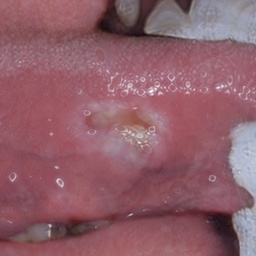}
            \end{subfigure}
            \hspace{0.1mm}
            \begin{subfigure}{0.091\textwidth}
                \centering \hfill
                \includegraphics[width=\linewidth]{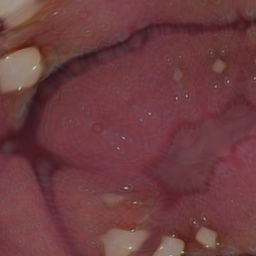}
            \end{subfigure}
            \begin{subfigure}{0.091\textwidth}
                \centering \hfill
                \includegraphics[width=\linewidth]{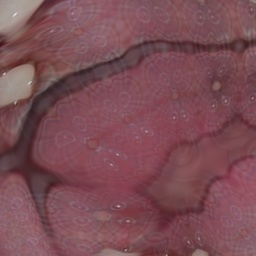}
            \end{subfigure}
            \begin{subfigure}{0.091\textwidth}
                \centering \hfill
                \includegraphics[width=\linewidth]{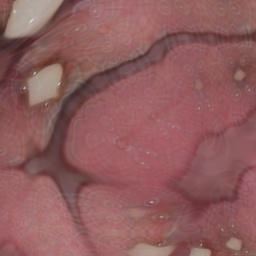}
            \end{subfigure}
            \begin{subfigure}{0.091\textwidth}
                \centering \hfill
                \includegraphics[width=\linewidth]{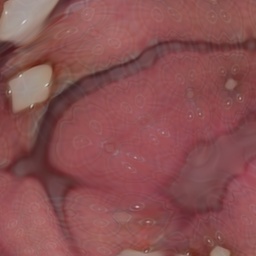}
            \end{subfigure}
            \begin{subfigure}{0.091\textwidth}
                \centering \hfill
                \includegraphics[width=\linewidth]{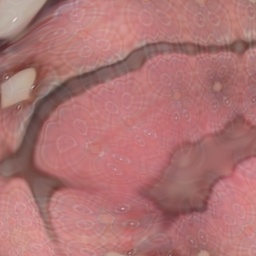}
            \end{subfigure}
        \end{minipage}
    \end{minipage}

    \begin{minipage}{\linewidth}
         \begin{minipage}{0.15cm}\rotatebox{90}{\footnotesize{aphthous}}\end{minipage}
        \begin{minipage}{\linewidth}
            \begin{subfigure}{0.091\textwidth}
                \centering \hfill 
                \includegraphics[width=\linewidth]{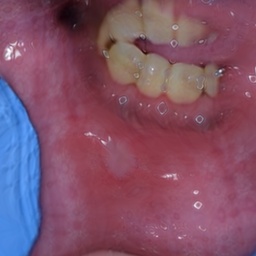}
            \end{subfigure}
            \begin{subfigure}{0.091\textwidth}
                \centering \hfill
                \includegraphics[width=\linewidth]{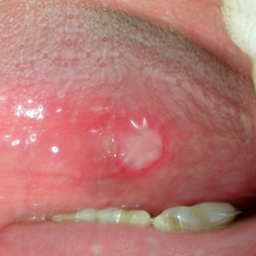}
            \end{subfigure}
            \begin{subfigure}{0.091\textwidth}
                \centering \hfill
                \includegraphics[width=\linewidth]{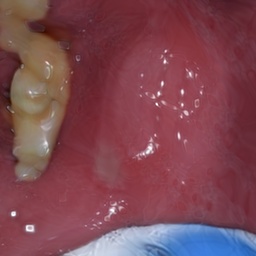}
            \end{subfigure}
            \begin{subfigure}{0.091\textwidth}
                \centering \hfill
                \includegraphics[width=\linewidth]{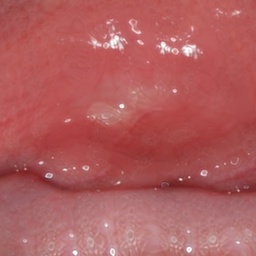}
            \end{subfigure}
            \begin{subfigure}{0.091\textwidth}
                \centering \hfill
                \includegraphics[width=\linewidth]{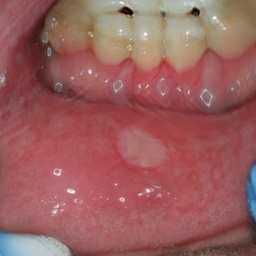}
            \end{subfigure}
            \hspace{0.1mm}
            \begin{subfigure}{0.091\textwidth}
                \centering \hfill
                \includegraphics[width=\linewidth]{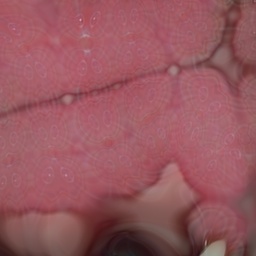}
            \end{subfigure}
            \begin{subfigure}{0.091\textwidth}
                \centering \hfill
                \includegraphics[width=\linewidth]{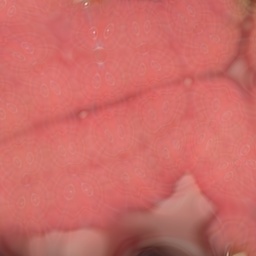}
            \end{subfigure}
            \begin{subfigure}{0.091\textwidth}
                \centering \hfill
                \includegraphics[width=\linewidth]{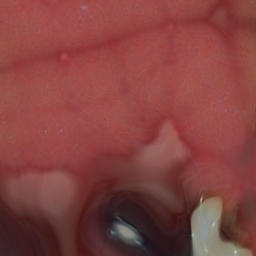}
            \end{subfigure}
            \begin{subfigure}{0.091\textwidth}
                \centering \hfill
                \includegraphics[width=\linewidth]{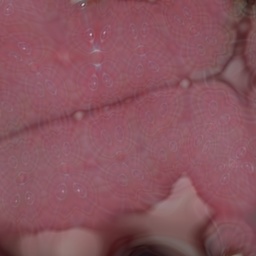}
            \end{subfigure}
            \begin{subfigure}{0.091\textwidth}
                \centering \hfill
                \includegraphics[width=\linewidth]{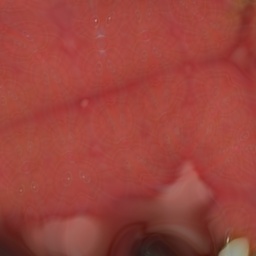}
            \end{subfigure}
        \end{minipage}
    \end{minipage}
    \vspace{0.5mm}

    \begin{minipage}{\linewidth}
         \begin{minipage}{0.15cm}\rotatebox{90}{\footnotesize{neoplastic}}\end{minipage}
         \begin{minipage}{\linewidth}
            \begin{subfigure}{0.091\textwidth}
                \centering \hfill 
                \includegraphics[width=\linewidth]{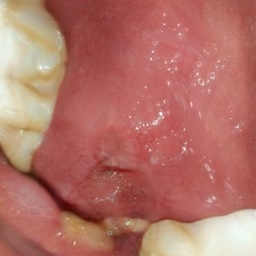}
            \end{subfigure}
            \begin{subfigure}{0.091\textwidth}
                \centering \hfill
                \includegraphics[width=\linewidth]{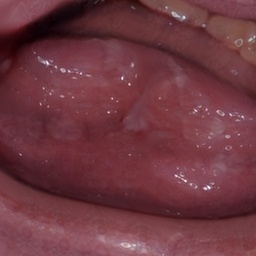}
            \end{subfigure}
            \begin{subfigure}{0.091\textwidth}
                \centering \hfill
                \includegraphics[width=\linewidth]{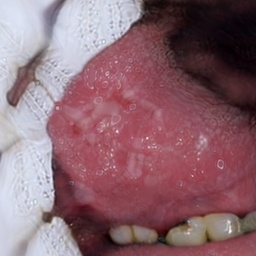}
            \end{subfigure}
            \begin{subfigure}{0.091\textwidth}
                \centering \hfill
                \includegraphics[width=\linewidth]{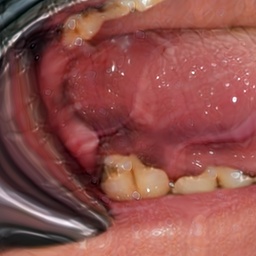}
            \end{subfigure}
            \begin{subfigure}{0.091\textwidth}
                \centering \hfill
                \includegraphics[width=\linewidth]{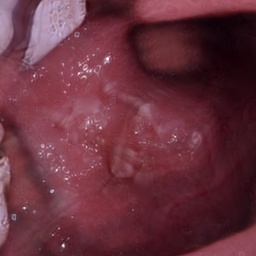}
            \end{subfigure}
            \hspace{0.1mm}
            \begin{subfigure}{0.091\textwidth}
                \centering \hfill
                \includegraphics[width=\linewidth]{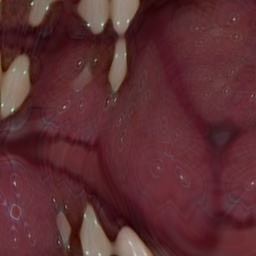}
            \end{subfigure}
            \begin{subfigure}{0.091\textwidth}
                \centering \hfill
                \includegraphics[width=\linewidth]{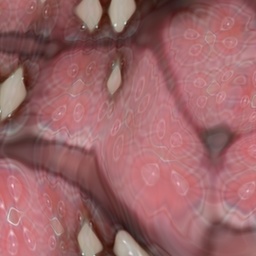}
            \end{subfigure}
            \begin{subfigure}{0.091\textwidth}
                \centering \hfill
                \includegraphics[width=\linewidth]{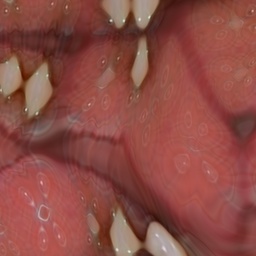}
            \end{subfigure}
            \begin{subfigure}{0.091\textwidth}
                \centering \hfill
                \includegraphics[width=\linewidth]{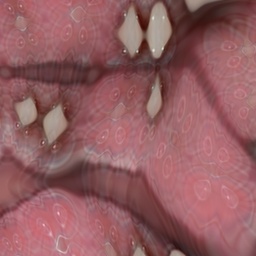}
            \end{subfigure}
            \begin{subfigure}{0.091\textwidth}
                \centering \hfill
                \includegraphics[width=\linewidth]{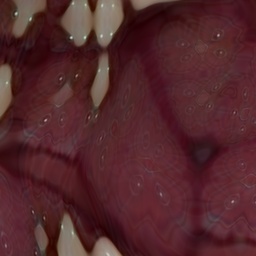}
            \end{subfigure}
        \end{minipage}
    \end{minipage}

    \end{minipage}
    
    \caption{Synthetic image examples generated from the PhotoMOCI dataset using four synthetic image generation setups: Stable Diffusion conditioned on text (top-left), Stable Diffusion conditioned on text and image (top-right), StyleGAN3 conditioned on text (bottom-left), and AC-SG3 (bottom-right). Within each block, the first, second, and third rows correspond to traumatic, aphthous, and neoplastic lesions, respectively.}
    \label{fig:SynthPhotoMOCI-examples}
\end{figure*}

\begin{figure*}[h]
    \centering
    \begin{minipage}{\textwidth}

    \centering \footnotesize{\hspace{0.1cm} StableDiffusion (txt) \hspace{4.9cm} StableDiffusion (txt+img)} \vspace{0.5mm}
    
    \begin{minipage}{\linewidth}
       \begin{minipage}{0.15cm}\rotatebox{90}{\footnotesize{cancer}}\end{minipage}
        \begin{minipage}{\linewidth}
            \begin{subfigure}{0.091\textwidth}
                \centering \hfill 
                \includegraphics[width=\linewidth]{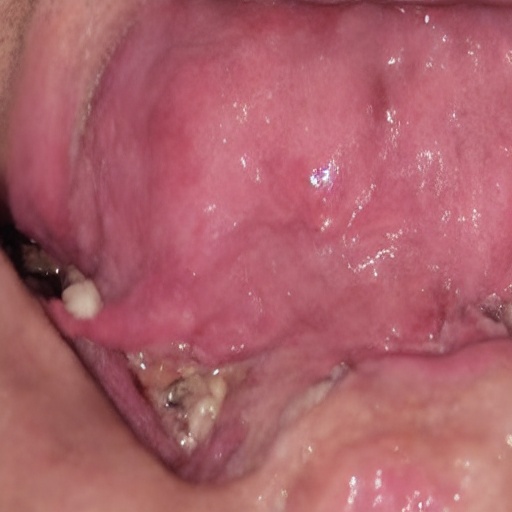}
            \end{subfigure}
            \begin{subfigure}{0.091\textwidth}
                \centering \hfill
                \includegraphics[width=\linewidth]{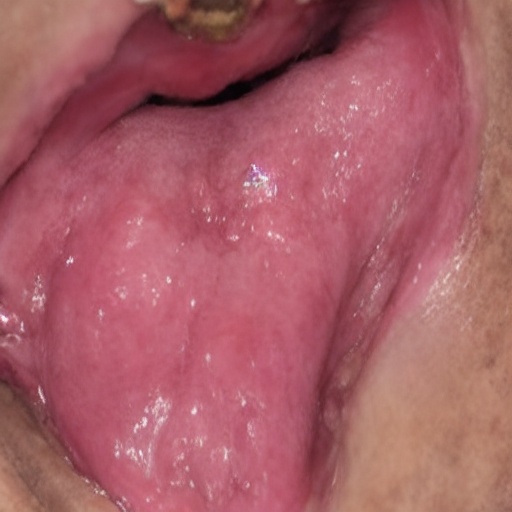}
            \end{subfigure}
            \begin{subfigure}{0.091\textwidth}
                \centering \hfill
                \includegraphics[width=\linewidth]{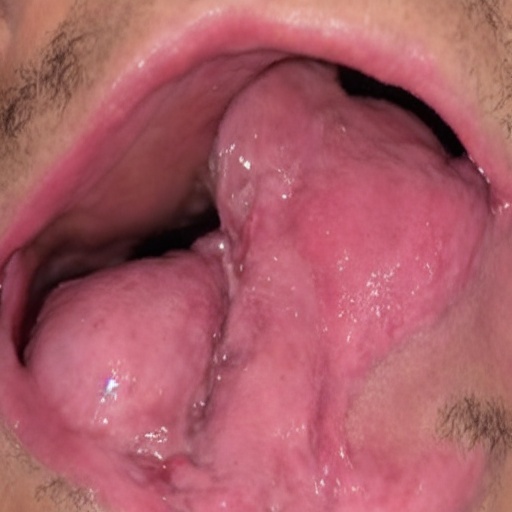}
            \end{subfigure}
            \begin{subfigure}{0.091\textwidth}
                \centering \hfill
                \includegraphics[width=\linewidth]{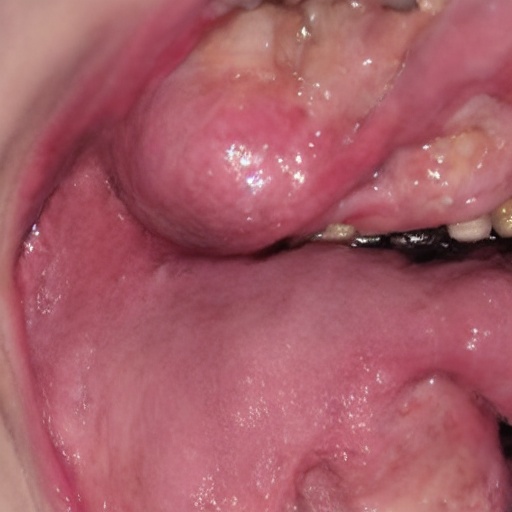}
            \end{subfigure}
            \begin{subfigure}{0.091\textwidth}
                \centering \hfill
                \includegraphics[width=\linewidth]{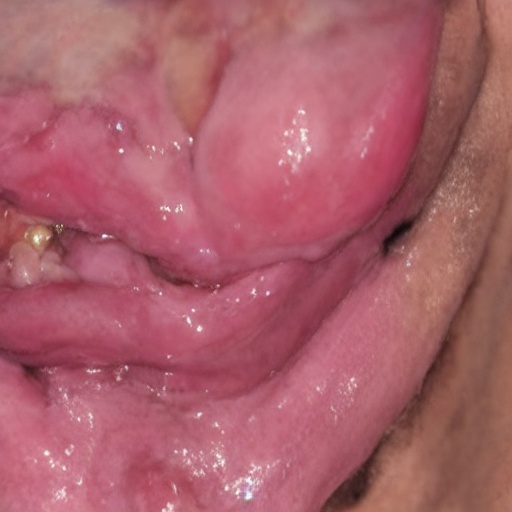}
            \end{subfigure}
            \hspace{0.1mm} 
            \begin{subfigure}{0.091\textwidth}
                \centering \hfill
                \includegraphics[width=\linewidth]{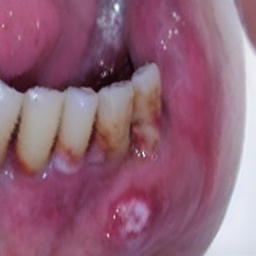}
            \end{subfigure}
            \begin{subfigure}{0.091\textwidth}
                \centering \hfill
                \includegraphics[width=\linewidth]{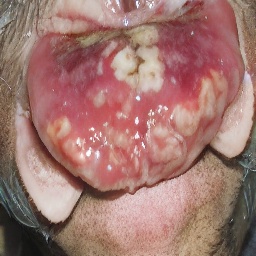}
            \end{subfigure}
            \begin{subfigure}{0.091\textwidth}
                \centering \hfill
                \includegraphics[width=\linewidth]{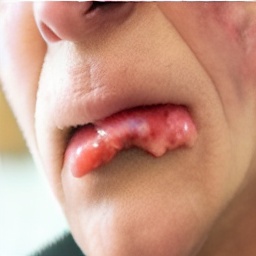}
            \end{subfigure}
            \begin{subfigure}{0.091\textwidth}
                \centering \hfill
                \includegraphics[width=\linewidth]{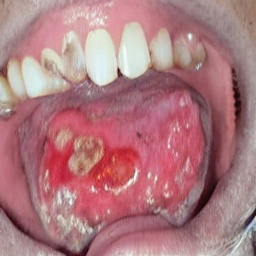}
            \end{subfigure}
            \begin{subfigure}{0.091\textwidth}
                \centering \hfill
                \includegraphics[width=\linewidth]{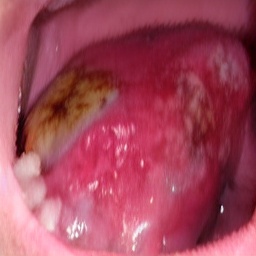}
            \end{subfigure}
        \end{minipage}
    \end{minipage}

    \begin{minipage}{\linewidth}
         \begin{minipage}{0.15cm}\rotatebox{90}{\footnotesize{non-cancer}}\end{minipage}
         \begin{minipage}{\linewidth}
            \begin{subfigure}{0.091\textwidth}
                \centering \hfill 
                \includegraphics[width=\linewidth]{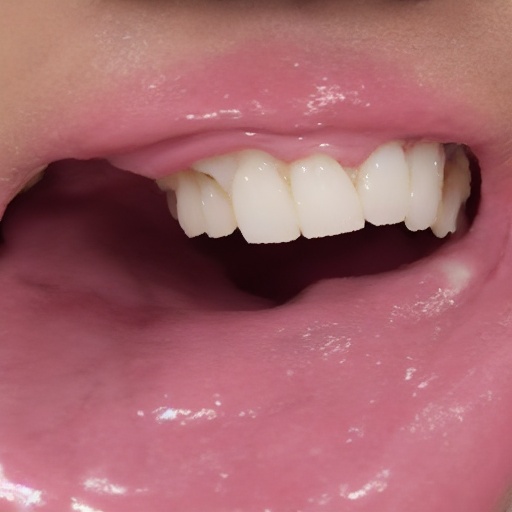}
            \end{subfigure}
            \begin{subfigure}{0.091\textwidth}
                \centering \hfill
                \includegraphics[width=\linewidth]{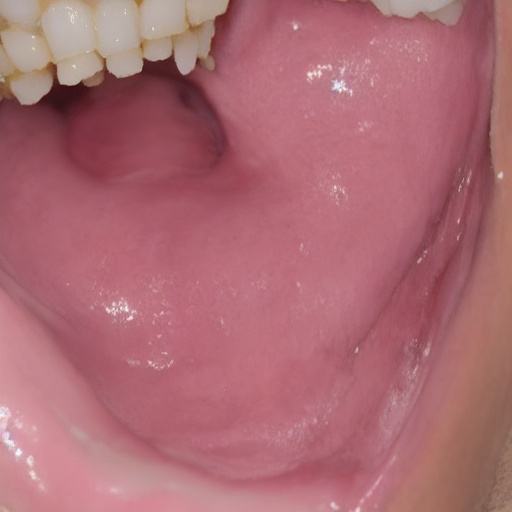}
            \end{subfigure}
            \begin{subfigure}{0.091\textwidth}
                \centering \hfill
                \includegraphics[width=\linewidth]{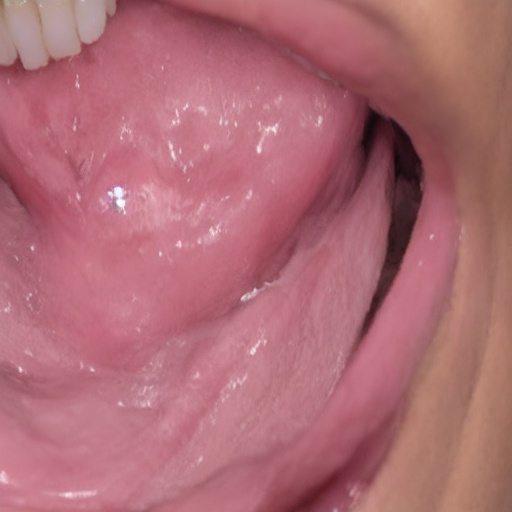}
            \end{subfigure}
            \begin{subfigure}{0.091\textwidth}
                \centering \hfill
                \includegraphics[width=\linewidth]{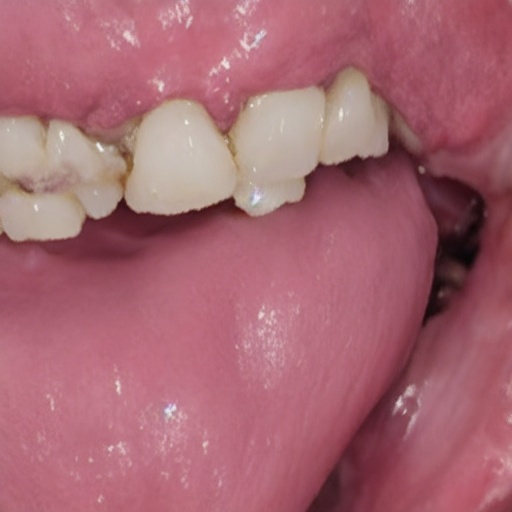}
            \end{subfigure}
            \begin{subfigure}{0.091\textwidth}
                \centering \hfill
                \includegraphics[width=\linewidth]{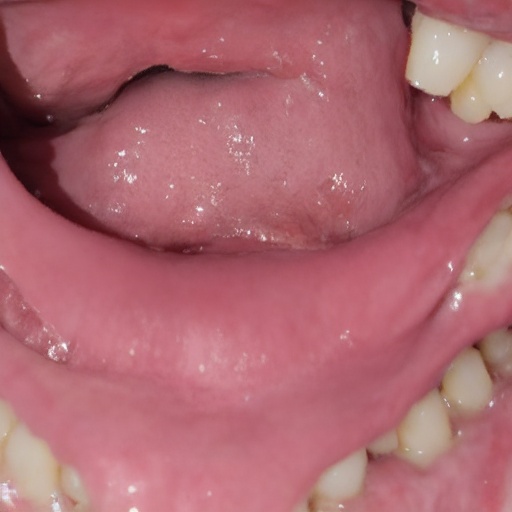}
            \end{subfigure}
            \hspace{0.1mm} 
            \begin{subfigure}{0.091\textwidth}
                \centering \hfill
                \includegraphics[width=\linewidth]{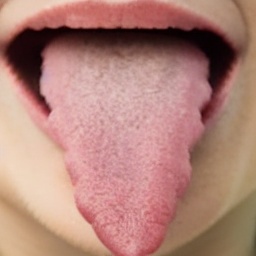}
            \end{subfigure}
            \begin{subfigure}{0.091\textwidth}
                \centering \hfill
                \includegraphics[width=\linewidth]{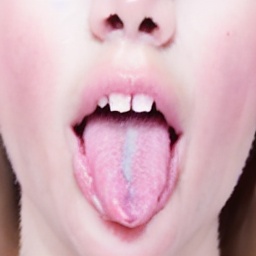}
            \end{subfigure}
            \begin{subfigure}{0.091\textwidth}
                \centering \hfill
                \includegraphics[width=\linewidth]{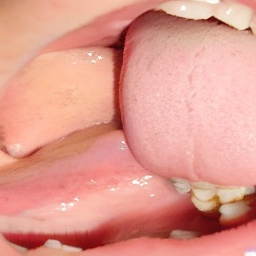}
            \end{subfigure}
            \begin{subfigure}{0.091\textwidth}
                \centering \hfill
                \includegraphics[width=\linewidth]{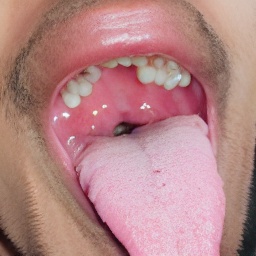}
            \end{subfigure}
            \begin{subfigure}{0.091\textwidth}
                \centering \hfill
                \includegraphics[width=\linewidth]{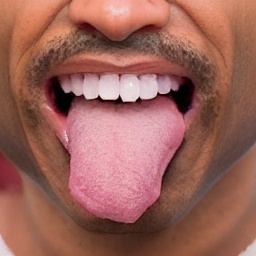}
            \end{subfigure}
        \end{minipage}
    \end{minipage}

    \vspace{2mm}

    \centering \hspace{1.cm}\footnotesize{StyleGAN3 (lbl) \hspace{4.4cm} 
    AuxiliaryClassifier-StyleGAN3}  \vspace{0.5mm}

    \begin{minipage}{\linewidth}
        \begin{minipage}{0.15cm}\rotatebox{90}{\footnotesize{cencer}}\end{minipage}
        \begin{minipage}{\linewidth}
            \begin{subfigure}{0.091\textwidth}
                \centering \hfill 
                \includegraphics[width=\linewidth]{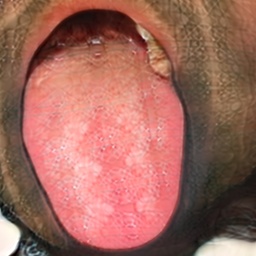}
            \end{subfigure}
            \begin{subfigure}{0.091\textwidth}
                \centering \hfill
                \includegraphics[width=\linewidth]{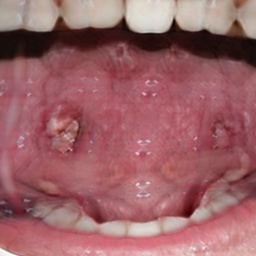}
            \end{subfigure}
            \begin{subfigure}{0.091\textwidth}
                \centering \hfill
                \includegraphics[width=\linewidth]{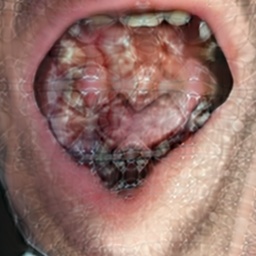}
            \end{subfigure}
            \begin{subfigure}{0.091\textwidth}
                \centering \hfill
                \includegraphics[width=\linewidth]{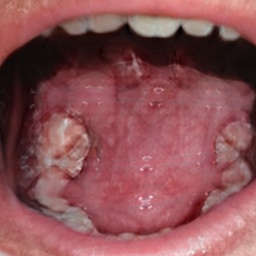}
            \end{subfigure}
            \begin{subfigure}{0.091\textwidth}
                \centering \hfill
                \includegraphics[width=\linewidth]{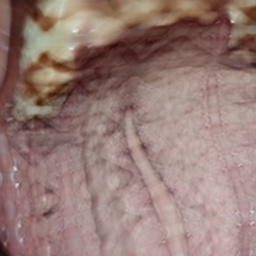}
            \end{subfigure}
            \hspace{0.1mm} 
            \begin{subfigure}{0.091\textwidth}
                \centering \hfill
                \includegraphics[width=\linewidth]{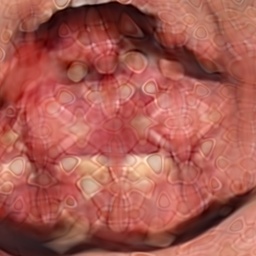}
            \end{subfigure}
            \begin{subfigure}{0.091\textwidth}
                \centering \hfill
                \includegraphics[width=\linewidth]{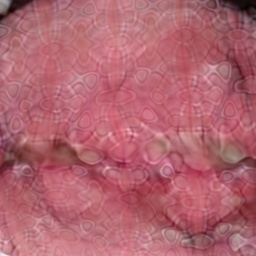}
            \end{subfigure}
            \begin{subfigure}{0.091\textwidth}
                \centering \hfill
                \includegraphics[width=\linewidth]{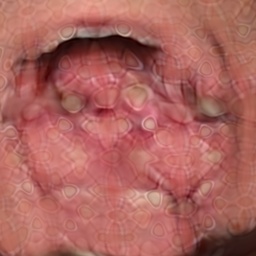}
            \end{subfigure}
            \begin{subfigure}{0.091\textwidth}
                \centering \hfill
                \includegraphics[width=\linewidth]{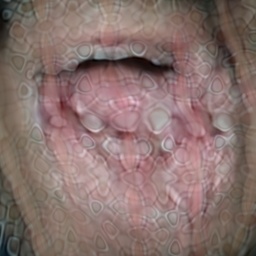}
            \end{subfigure}
            \begin{subfigure}{0.091\textwidth}
                \centering \hfill
                \includegraphics[width=\linewidth]{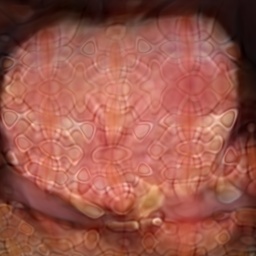}
            \end{subfigure}
        \end{minipage}
    \end{minipage}
    \vspace{0.5mm}

    \begin{minipage}{\linewidth}
         \begin{minipage}{0.15cm}\rotatebox{90}{\footnotesize{non-cancer}}\end{minipage}
        \begin{minipage}{\linewidth}
            \begin{subfigure}{0.091\textwidth}
                \centering \hfill 
                \includegraphics[width=\linewidth]{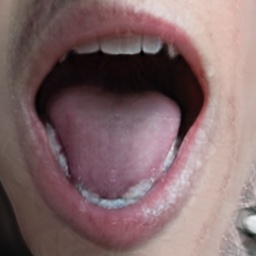}
            \end{subfigure}
            \begin{subfigure}{0.091\textwidth}
                \centering \hfill
                \includegraphics[width=\linewidth]{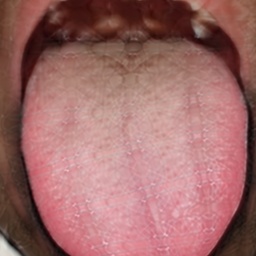}
            \end{subfigure}
            \begin{subfigure}{0.091\textwidth}
                \centering \hfill
                \includegraphics[width=\linewidth]{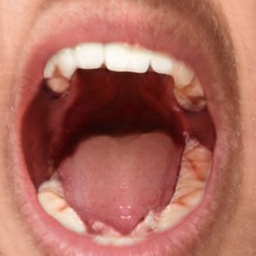}
            \end{subfigure}
            \begin{subfigure}{0.091\textwidth}
                \centering \hfill
                \includegraphics[width=\linewidth]{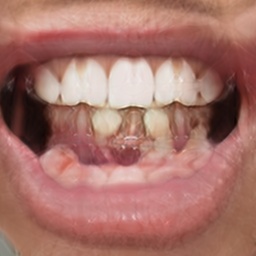}
            \end{subfigure}
            \begin{subfigure}{0.091\textwidth}
                \centering \hfill
                \includegraphics[width=\linewidth]{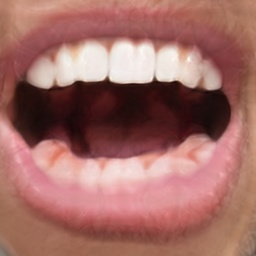}
            \end{subfigure}
            \hspace{0.1mm} 
            \begin{subfigure}{0.091\textwidth}
                \centering \hfill
                \includegraphics[width=\linewidth]{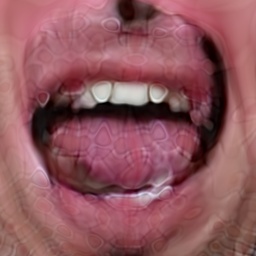}
            \end{subfigure}
            \begin{subfigure}{0.091\textwidth}
                \centering \hfill
                \includegraphics[width=\linewidth]{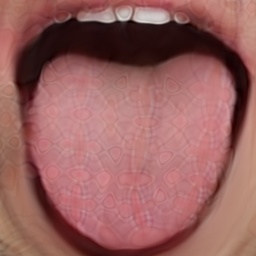}
            \end{subfigure}
            \begin{subfigure}{0.091\textwidth}
                \centering \hfill
                \includegraphics[width=\linewidth]{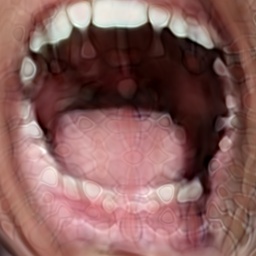}
            \end{subfigure}
            \begin{subfigure}{0.091\textwidth}
                \centering \hfill
                \includegraphics[width=\linewidth]{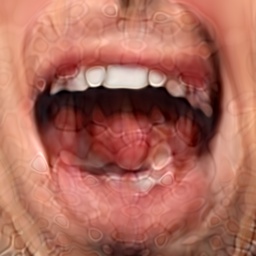}
            \end{subfigure}
            \begin{subfigure}{0.091\textwidth}
                \centering \hfill
                \includegraphics[width=\linewidth]{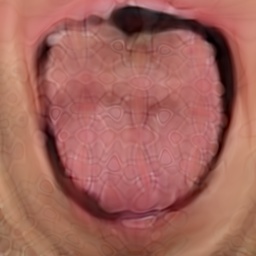}
            \end{subfigure}
        \end{minipage}
    \end{minipage}

    \end{minipage}
    
    \caption{Synthetic image examples generated from the KOCD  using four synthetic image generation setups: Stable Diffusion conditioned on text (top-left), Stable Diffusion conditioned on text and image (top-right), StyleGAN3 conditioned on text (bottom-left), and AC-SG3 (bottom-right). Within each block, the first row corresponds to the cancer class, and the second row corresponds to the non-cancer class.}
    \label{fig:SynthKOCD-examples}
\end{figure*}

This insight is also supported by Figures \ref{fig:SynthPhotoMOCI-examples} and \ref{fig:SynthKOCD-examples}, which present examples of generated synthetic images by class with the considered generation strategies on PhotoMOCI and KOCD, respectively. Qualitatively, the images produced by SD(txt) are clearly recognizable as synthetic to the human eye, showing anatomical features of the oral cavity that do not exist in either dataset and exhibiting almost no variability within the same class. By contrast, SD(txt-img) and SG3(lbl) generate synthetic images that appear visually plausible and capture the distinctive features of the classes (e.g., necrotic tissue in the case of neoplastic lesions) in both datasets. Finally, a discrepancy can be observed in the quality of the images produced by AC-SG3 across datasets: on PhotoMOCI, the generated samples lack intra-class variability, displaying a recurring visual pattern within the same class, whereas this phenomenon is not present on KOCD. This suggests that AC-SG3 may struggle to generalize across settings, such as PhotoMOCI, where the class granularity is higher.

Finally, we generate the synthetic samples that will be used in the next stages of the methodology. 
Specifically, for each synthetic image generation setup trained on KOCD, we generate 3000 images (1500 per class). While we generate 2100 images for PhotoMOCI (700 per class).

We report the wall-clock time required by each generative setup in Table~\ref{tab:generation_runtime}, referring to the same hardware, an NVIDIA A100 GPU with 32GB of memory. For each setup, the generation of a single image was repeated 10 times and we report the mean generation time with the 95\% confidence interval. Additionally, for each dataset, we report the actual time for the entire synthetic image generation process. GAN-based models are computationally more efficient than diffusion-based models; however, synthetic image generation is performed only once as an offline preprocessing step and therefore does not affect the downstream classifier training and inference phases.

\begin{table}[t]
\footnotesize
\centering
\caption{Computational time for generating synthetic images on an NVIDIA A100 GPU.}
\label{tab:generation_runtime}
\begin{tabular}{l c c c}
\toprule
\textbf{\makecell[l]{Synthetic img.\\gen. setup}} &
\textbf{\makecell[l]{Single image \\ time (s)}} &
\textbf{\makecell[l]{PhotoMOCI\\ 2100 images}} &
\textbf{\makecell[l]{KOCD\\ 3000 images}} \\
\midrule
SD(txt)     & 1.86 $\pm$ 0.05 & 1 h 05 min 06 s & 1 h 33 min 00 s \\
SD(txt+img) & 0.82 $\pm$ 0.06 & 28 min 42 s & 41 min 00 s \\
SG3(lbl)    & 0.045 $\pm$ 0.003 & 1 min 35 s & 2 min 15 s \\
AC-SG3      & 0.048 $\pm$ 0.004 & 1 min 41 s & 2 min 24 s \\
\bottomrule
\end{tabular}
\end{table}

\begin{table}[t]
\centering
\caption{SIF modules performance: classification results of SPC and SID on PhotoMOCI and KOCD considering the four synthetic data generation.}
\label{tab:SIF_metrics}
\begin{tabular}{l @{\hspace{2.2mm}}l @{\hspace{2.2mm}} c @{\hspace{2.2mm}} c}
\toprule
\multirow{2}{*}{Dataset} & \multirow{2}{*}{\makecell[l]{Synthetic img.\\gen. setup}} & \multicolumn{2}{c}{Accuracy}\\
\cmidrule{3-4}
& & \textbf{SPC $\uparrow$} & \textbf{SID$\downarrow$} \\
\midrule
\multirow{4}{*}{PhotoMOCI (ours)} 
& SD(txt)       & 0.368 $\pm$ 0.045 & 0.788 $\pm$ 0.032 \\
& SD(txt+img)   & 0.771 $\pm$ 0.037 & 0.643 $\pm$ 0.040 \\
& SG3(lbl)      & 0.740 $\pm$ 0.029 & 0.596 $\pm$ 0.038 \\
& AC-SG3        & 0.893 $\pm$ 0.036 & 0.921 $\pm$ 0.036 \\

\midrule

\multirow{4}{*}{KOCD\cite{MOHD_ZAID_RASHID_oral_cancer_2024}} 
& SD(txt)       & 0.617 $\pm$ 0.025 & 0.709 $\pm$ 0.030 \\
& SD(txt+img)   & 0.914 $\pm$ 0.029 & 0.677 $\pm$ 0.021\\
& SG3(lbl)      & 0.857 $\pm$ 0.033 & 0.780 $\pm$ 0.024 \\
& AC-SG3        & 0.783 $\pm$ 0.026  & 0.893 $\pm$ 0.019 \\
\bottomrule
\end{tabular}%
\end{table}


\subsection{Synthetic Image Filter}\label{sec:SIF_experiments}

\textbf{Configuration Setup}.
For the implementation of the SIF auxiliary classifiers, SPC and SID, we employ two ResNet50 architectures.
For each setup, we perform a training hyperparameter grid search. With the batch size maximized to 64 under GPU constraints, we search $lr$ in $[1e\textendash4,5e\textendash4,1e\textendash5,5e\textendash5,1e\textendash6,5e\textendash6]$ using the Adam optimizer \cite{kingma2014adam}. Given the best configuration, models were trained for 150 epochs with early stopping (patience 10). To get statistical results, we repeated each training 10 times with different seeds, reporting the mean and 95\% confidence interval. 
Regarding SPC, we use 490 and 665 synthetic images for training on PhotoMOCI and KOCD, respectively; while validation and test sets include only the original real images. For SID, we use the original train, validation, and test splits and augment each with an equal number of randomly selected synthetic images, preserving class balance to train a binary classifier able to distinguish between real and synthetic. This results in 1400 and 1900 for PhotoMOCI and KOCD, respectively.

\textbf{Results}.
Table \ref{tab:SIF_metrics} reports the results obtained on both classification tasks by SPC and SID on PhotoMOCI and KOCD. We observe that, for SPC addressing the same oral lesion classification task, the accuracy is high across all configurations, except for SD(txt) on PhotoMOCI, suggesting that the class distributions are generally well separable. In contrast, for SID, we would ideally expect performance close to random guessing; however, the observed accuracy values are higher, indicating a discrepancy between the distributions of real and synthetic data. This discrepancy is not necessarily apparent to human inspection but is captured through latent patterns.
Consistent with Tables \ref{tab:metriche_generative_KODC} and \ref{tab:metriche_generative_PhotoMOCI}, SD(txt+img) and SG3 achieve the highest performance, suggesting that the effectiveness of the two auxiliary classification tasks can be preliminarily assessed using generative evaluation metrics (FID, KID, Precision, Recall, and Coverage).

\begin{table*}[t]
\footnotesize
\centering
\caption{Number and percentage of synthetic images retained by SIF on PhotoMOCI and KOCD, reported in aggregate and by class.}
\label{tab:SIF_retained}
{
\setlength{\tabcolsep}{1.9pt}
\begin{tabular}{l c c c c c c c}
\toprule
\multirow{2}{*}{\textbf{\makecell[l]{Synthetic\\ img. gen.\\setup}}} &
\multicolumn{4}{c}{\textbf{PhotoMOCI}} &
\multicolumn{3}{c}{\textbf{KOCD}} \\
\cmidrule(lr){2-5}\cmidrule(lr){6-8}
& \textbf{Total} &
\textbf{Traumatic} &
\textbf{Aphthous} &
\textbf{Neoplastic} &
\textbf{Total} &
\textbf{Cancer} &
\textbf{Non-cancer} \\
\midrule
SD(txt)     & 139 (6.6\%) & 30 (4.3\%)  & 83 (11.9\%) & 26 (3.7\%)  & 308 (10.3\%) & 91 (6.1\%)  & 217 (14.5\%) \\
SD(txt+img) & 470 (22.4\%) & 157 (22.4\%) & 119 (17.0\%) & 194 (27.7\%) & 653 (21.8\%) & 362 (24.1\%) & 291 (19.4\%) \\
SG3(lbl)    & 367 (17.5\%) & 141 (20.1\%) & 137 (19.6\%) & 89 (12.7\%)  & 776 (25.9\%) & 415 (27.7\%) & 361 (24.1\%) \\
AC-SG3      & 70 (3.3\%)   & 12 (1.7\%)   & 19 (2.7\%)   & 39 (5.6\%)   & 152 (5.1\%)  & 24 (1.6\%)   & 128 (8.5\%) \\
\bottomrule
\end{tabular}
}
\end{table*}

After training the SIF modules, we apply the complete filtering pipeline to the synthetic images generated by each configuration and retain only the samples accepted by both auxiliary classifiers. Table~\ref{tab:SIF_retained} reports the number and percentage of synthetic images retained by the SIF for each synthetic generation setup for PhotoMOCI and KOCD (both overall and per class). Percentages are computed with respect to the generated pool of each setup: 2100 images for PhotoMOCI (700 per class) and 3000 images for KOCD (1500 per class). It provides an overview of how many generated images effectively contribute to the augmented training sets. Overall, the number of retained samples varies considerably across setups, with SD(txt+img) and SG3(lbl) generally producing the largest number of accepted samples, with 470 (22.4\%) and 367 (17.5\%) on PhotoMOCI and 653 (21.8\%) and 776 (25.9\%) on KOCD, respectively. Moreover, the retained images are reasonably distributed across disease categories rather than being concentrated in a single class. Although some variability is observed, no pronounced bias toward or against a particular pathology emerges after filtering.
The computational time required by SIF corresponds to the inference time of the auxiliary architectures used in the filtering pipeline. We average 10 inference runs for each classification architecture. ResNet50-based SIF filtering requires 10.1 ms per image, while ViT-based filtering requires 21.7 ms. This introduces a minimal overhead and does not impact the overall computational cost.

\begin{table*}[b]
\tiny
\centering
\caption{Classification performance on PhotoMOCI using ResNet and ViT under different data augmentation strategies. The evaluated setups include training without augmentation (No aug.) and with traditional augmentation (Tr. aug.), marked with \xmark\ since SIF is not applicable, as well as four generative-AI–based setups evaluated both without and with SIF, distinguished by \checkmark. Highlighted in \colorbox{Black!15!White}{gray}, the setups where SIF improves performance over non-SIF counterparts and surpasses traditional augmentation. Highlighted in \textbf{bold}, the best-performing augmentation setup for each classifier.}
\label{tab:metriche_classificazione_PhotoMOCI}
\setlength{\tabcolsep}{1.8pt}
\scalebox{1.14}[1.3]{
\begin{tabular}{llccccc}
\toprule
\textbf{Model} & \textbf{Conditioning} & \textbf{SIF} &
\textbf{Acc} & \textbf{Prec} & \textbf{Recall} & \textbf{F1-score} \\
\midrule
\multirow{11}{*}{
\begin{minipage}{0.15cm}
\rotatebox{90}{\footnotesize{ResNet\cite{resnet}}}
\end{minipage}
}
& No aug. & \xmark &
0.81050$\pm$0.00532 & 0.81538$\pm$0.00553 &
0.81076$\pm$0.00581 & 0.81414$\pm$0.00601 \\
\cmidrule{2-7}
& Tr.aug.(base) & \xmark &
0.82519$\pm$0.00600 & 0.83130$\pm$0.00572 &
0.82316$\pm$0.00603 & 0.82873$\pm$0.00612 \\
\cmidrule{2-7}
& \multirow{2}{*}{SD(txt)}
& & 0.74153$\pm$0.00590 & 0.74797$\pm$0.00619 &
0.74225$\pm$0.00528 & 0.74550$\pm$0.00493 \\
& & \checkmark &
0.80076$\pm$0.00544 & 0.82653$\pm$0.00485 &
0.80567$\pm$0.00521 & 0.81237$\pm$0.00495 \\
\cmidrule{2-7}
& \multirow{2}{*}{SD(txt+img)}
& & 0.83655$\pm$0.00520 & 0.83704$\pm$0.00446 &
0.83679$\pm$0.00499 & 0.83698$\pm$0.00524 \\
& & \checkmark &
\cellcolor[gray]{0.85}\textbf{0.84253$\pm$0.00498} &
\cellcolor[gray]{0.85}\textbf{0.85137$\pm$0.00507} &
\cellcolor[gray]{0.85}\textbf{0.84294$\pm$0.00481} &
\cellcolor[gray]{0.85}\textbf{0.85095$\pm$0.00511} \\
\cmidrule{2-7}
& \multirow{2}{*}{SG3(lbl)}
& & 0.83282$\pm$0.00488 & 0.84001$\pm$0.00519 &
0.83335$\pm$0.00535 & 0.83560$\pm$0.00515 \\
& & \checkmark &
\cellcolor[gray]{0.85}0.83603$\pm$0.00454 &
\cellcolor[gray]{0.85}0.84261$\pm$0.00478 &
\cellcolor[gray]{0.85}0.83776$\pm$0.00440 &
\cellcolor[gray]{0.85}0.83860$\pm$0.00423 \\
\cmidrule{2-7}
& \multirow{2}{*}{AC-SG3}
& & 0.79153$\pm$0.00638 & 0.80797$\pm$0.00533 &
0.79225$\pm$0.00604 & 0.79450$\pm$0.00592 \\
& & \checkmark &
0.80988$\pm$0.00545 & 0.81530$\pm$0.00630 &
0.80849$\pm$0.00577 & 0.81069$\pm$0.00563 \\
\midrule
\multirow{11}{*}{
\begin{minipage}{0.15cm}
\rotatebox{90}{\footnotesize{ViT\cite{vit}}}
\end{minipage}
}
& No aug. & \xmark &
0.82173$\pm$0.00481 & 0.82330$\pm$0.00499 &
0.82033$\pm$0.00508 & 0.82125$\pm$0.00510 \\
\cmidrule{2-7}
& Tr.aug.(base) & \xmark &
0.82950$\pm$0.00617 & 0.83176$\pm$0.00628 &
0.82984$\pm$0.00637 & 0.83014$\pm$0.00615 \\
\cmidrule{2-7}
& \multirow{2}{*}{SD(txt)}
& & 0.75800$\pm$0.00534 & 0.76146$\pm$0.00593 &
0.75724$\pm$0.00520 & 0.76031$\pm$0.00563 \\
& & \checkmark &
0.82109$\pm$0.00467 & 0.82739$\pm$0.00446 &
0.82335$\pm$0.00489 & 0.82616$\pm$0.00423 \\
\cmidrule{2-7}
& \multirow{2}{*}{SD(txt+img)}
& & 0.84500$\pm$0.00497 & 0.85190$\pm$0.00497 &
0.84892$\pm$0.00489 & 0.84876$\pm$0.00473 \\
& & \checkmark &
\cellcolor[gray]{0.85}\textbf{0.85296}$\pm$\textbf{0.00414} &
\cellcolor[gray]{0.85}\textbf{0.85948}$\pm$\textbf{0.00427} &
\cellcolor[gray]{0.85}\textbf{0.85310}$\pm$\textbf{0.00416} &
\cellcolor[gray]{0.85}\textbf{0.85761}$\pm$\textbf{0.00444} \\
\cmidrule{2-7}
& \multirow{2}{*}{SG3(lbl)}
& & 0.84174$\pm$0.00511 & 0.84655$\pm$0.00519 &
0.84281$\pm$0.00532 & 0.84502$\pm$0.00531 \\
& & \checkmark &
\cellcolor[gray]{0.85}0.84741$\pm$0.00413 &
\cellcolor[gray]{0.85}0.85269$\pm$0.00432 &
\cellcolor[gray]{0.85}0.84444$\pm$0.00429 &
\cellcolor[gray]{0.85}0.84897$\pm$0.00435 \\
\cmidrule{2-7}
& \multirow{2}{*}{AC-SG3}
& & 0.81453$\pm$0.00580 & 0.82138$\pm$0.00554 &
0.81724$\pm$0.00535 & 0.81882$\pm$0.00500 \\
& & \checkmark &
\cellcolor[gray]{0.85}0.82778$\pm$0.00440 &
\cellcolor[gray]{0.85}0.83003$\pm$0.00448 &
\cellcolor[gray]{0.85}0.82721$\pm$0.00453 &
\cellcolor[gray]{0.85}0.82959$\pm$0.00416 \\
\bottomrule
\end{tabular}%
}
\end{table*}


\subsection{Training Oral classifier}\label{sec:experiment-recognition}

\textbf{Configuration Setup}.
The recognition of lesions from photographic datasets was addressed as a classification task. We studied the impact of synthetic data on performance by training models on both original and augmented datasets (PhotoMOCI and KOCD). 
For each dataset, five augmentation setups were compared against a baseline trained only on the original data: traditional augmentation, two diffusion-based augmentations, and two GAN-based augmentations.
Except in the traditional augmentation setup, which was applied dynamically during training following common practice, synthetic images were added to the originals for training with and without applying SIF.
As image classifiers, we considered both a convolutional model and a transformer: ResNet50 \cite{resnet} and a Vision Transformer (ViT) \cite{vit}, both pre-trained on ImageNet \cite{imagenet} from torchvision
The model's evaluation is based on standard classification metrics: accuracy, precision, recall, and F1-score.
For each model and setup, we perform a hyperparameter grid search. With $batch=64$, we search the best $lr$ in $[1e\textendash4,5e\textendash4,1e\textendash5,5e\textendash5,1e\textendash6,5e\textendash6]$ using Adam \cite{kingma2014adam}. As for the experiments in the previous section, we train the models for 150 epochs with an early stopping condition, repeating the training, changing the seed, and reporting the mean and 95\% confidence interval. 

\textbf{Results}.
Tables \ref{tab:metriche_classificazione_PhotoMOCI} and \ref{tab:metriche_classificazione_KOCD} summarize classification metrics for the PhotoMOCI and KOCD datasets, respectively, showing both similarities and differences. 
It is important to note that the datasets are relatively small compared to standard benchmarks such as ImageNet\cite{imagenet} or COCO\cite{coco}.
In all experiments, the best-performing setup involves Stable Diffusion conditioned on both image and text, utilizing either ResNet or ViT when applying SIF. On the PhotoMOCI dataset, this setup improves accuracy by 1.73\% and 2.35\% over traditional augmentation (baseline) for ResNet and ViT, respectively. On KOCD, these improvements over the baseline reach 2.38\% and 2.08\%.

\begin{table*}[t]
\tiny
\centering
\caption{Classification performance on KOCD using ResNet and ViT under different data augmentation strategies. The evaluated setups include training without augmentation (No aug.) and with traditional augmentation (Tr. aug.), marked with \xmark\ since SIF is not applicable, as well as four generative-AI–based setups evaluated both without and with SIF, distinguished by \checkmark. Highlighted in \colorbox{Black!15!White}{gray}, the setups where SIF improves performance over non-SIF counterparts and surpasses traditional augmentation. Highlighted in \textbf{bold}, the best-performing augmentation setup for each classifier.}
\label{tab:metriche_classificazione_KOCD}
\setlength{\tabcolsep}{1.8pt}
\scalebox{1.14}[1.3]{
\begin{tabular}{llccccc}
\toprule
\textbf{Model} & \textbf{Conditioning} & \textbf{SIF} &
\textbf{Acc} & \textbf{Prec} & \textbf{Recall} & \textbf{F1-score} \\
\midrule
\multirow{11}{*}{
\begin{minipage}{0.15cm}
\rotatebox{90}{\footnotesize{ResNet\cite{resnet}}}
\end{minipage}
}
& No aug. & \xmark &
0.90935$\pm$0.00453 & 0.90028$\pm$0.00452 &
0.90074$\pm$0.00449 & 0.90026$\pm$0.00454 \\
\cmidrule{2-7}
& Tr.aug.(base) & \xmark &
0.93109$\pm$0.00554 & 0.93103$\pm$0.00547 &
0.93120$\pm$0.00563 & 0.93095$\pm$0.00576 \\
\cmidrule{2-7}
& \multirow{2}{*}{SD(txt)}
& & 0.88564$\pm$0.00637 & 0.89160$\pm$0.00656 &
0.88713$\pm$0.00625 & 0.89121$\pm$0.00700 \\
& & \checkmark &
0.91616$\pm$0.00410 & 0.91937$\pm$0.00474 &
0.91668$\pm$0.00422 & 0.91871$\pm$0.00390 \\
\cmidrule{2-7}
& \multirow{2}{*}{SD(txt+img)}
& & 0.95011$\pm$0.00498 & 0.95225$\pm$0.00501 &
0.94917$\pm$0.00533 & 0.95075$\pm$0.00511 \\
& & \checkmark &
\cellcolor[gray]{0.85}\textbf{0.95492}$\pm$\textbf{0.00467} &
\cellcolor[gray]{0.85}\textbf{0.95719}$\pm$\textbf{0.00501} &
\cellcolor[gray]{0.85}\textbf{0.95266}$\pm$\textbf{0.00461} &
\cellcolor[gray]{0.85}\textbf{0.95543}$\pm$\textbf{0.00521} \\
\cmidrule{2-7}
& \multirow{2}{*}{SG3(lbl)}
& & 0.94013$\pm$0.00453 & 0.94233$\pm$0.00503 &
0.93809$\pm$0.00446 & 0.93917$\pm$0.00537 \\
& & \checkmark &
\cellcolor[gray]{0.85}0.94825$\pm$0.00399 &
\cellcolor[gray]{0.85}0.94889$\pm$0.00392 &
\cellcolor[gray]{0.85}0.94778$\pm$0.00428 &
\cellcolor[gray]{0.85}0.94811$\pm$0.00474 \\
\cmidrule{2-7}
& \multirow{2}{*}{AC-SG3}
& & 0.92815$\pm$0.00595 & 0.93158$\pm$0.00645 &
0.92868$\pm$0.00583 & 0.93101$\pm$0.00604 \\
& & \checkmark &
\cellcolor[gray]{0.85}0.93551$\pm$0.00546 &
\cellcolor[gray]{0.85}0.93700$\pm$0.00566 &
\cellcolor[gray]{0.85}0.93289$\pm$0.00485 &
\cellcolor[gray]{0.85}0.93300$\pm$0.00555 \\
\midrule
\multirow{11}{*}{
\begin{minipage}{0.15cm}
\rotatebox{90}{\footnotesize{ViT\cite{vit}}}
\end{minipage}
}
& No aug. & \xmark &
0.91820$\pm$0.00304 & 0.92210$\pm$0.00323 &
0.91727$\pm$0.00321 & 0.92008$\pm$0.00318 \\
\cmidrule{2-7}
& Tr.aug.(base) & \xmark &
0.93722$\pm$0.00395 & 0.93989$\pm$0.00402 &
0.93701$\pm$0.00401 & 0.93817$\pm$0.00432 \\
\cmidrule{2-7}
& \multirow{2}{*}{SD(txt)}
& & 0.89552$\pm$0.00392 & 0.89981$\pm$0.00416 &
0.89461$\pm$0.00364 & 0.89777$\pm$0.00390 \\
& & \checkmark &
0.93133$\pm$0.00374 & 0.93172$\pm$0.00345 &
0.93175$\pm$0.00322 & 0.93156$\pm$0.00332 \\
\cmidrule{2-7}
& \multirow{2}{*}{SD(txt+img)}
& & 0.95095$\pm$0.00436 & 0.95050$\pm$0.00408 &
0.94976$\pm$0.00452 & 0.94923$\pm$0.00439 \\
& & \checkmark &
\cellcolor[gray]{0.85}\textbf{0.95804$\pm$0.00397} &
\cellcolor[gray]{0.85}\textbf{0.95812$\pm$0.00401} &
\cellcolor[gray]{0.85}\textbf{0.95787}$\pm$\textbf{0.00389} &
\cellcolor[gray]{0.85}\textbf{0.95893}$\pm$\textbf{0.00397} \\
\cmidrule{2-7}
& \multirow{2}{*}{SG3(lbl)}
& & 0.94265$\pm$0.00401 & 0.94819$\pm$0.00362 &
0.94678$\pm$0.00360 & 0.94735$\pm$0.00370 \\
& & \checkmark &
\cellcolor[gray]{0.85}0.948911$\pm$0.00393 &
\cellcolor[gray]{0.85}0.95293$\pm$0.00323 &
\cellcolor[gray]{0.85}0.94809$\pm$0.00300 &
\cellcolor[gray]{0.85}0.934784$\pm$0.00353 \\
\cmidrule{2-7}
& \multirow{2}{*}{AC-SG3}
& & 0.93021$\pm$0.00613 & 0.93465$\pm$0.00617 &
0.92999$\pm$0.00523 & 0.93373$\pm$0.00614 \\
& & \checkmark &
\cellcolor[gray]{0.85}0.93935$\pm$0.00404 &
\cellcolor[gray]{0.85}0.94080$\pm$0.00430 &
\cellcolor[gray]{0.85}0.93863$\pm$0.00499 &
\cellcolor[gray]{0.85}0.93901$\pm$0.00462 \\
\bottomrule
\end{tabular}%
}
\end{table*}

Several relevant trends emerge from this analysis. Traditional data augmentation consistently outperforms training without augmentation by 1–2\%, confirming its effectiveness in enhancing model generalizability.
Regarding the different synthetic data augmentation techniques, highlighted in gray, we observe setups where SIF improves performance over non-SIF counterparts and surpasses traditional augmentation. The adoption of synthetic data augmentation without SIF does not necessarily yield improvements. Indeed, on both PhotoMOCI and KOCD, integrating synthetic data produced via SD(txt) and AC-SG3 without SIF (represented in the first row of each two-row block) results in either a performance degradation or marginal gains that fail to surpass traditional augmentation, regardless of the classifier used.
However, by filtering images using SIF, a performance recovery is achieved when using AC-SG3 in three out of four cases (with ViT and ResNet on KOCD). This suggests that AC-SG3-based synthetic augmentation can generate high-quality images that, when selected by SIF, successfully drive performance gains. In contrast, this improvement is absent in the SD(txt) setup, where accuracy after applying SIF merely matches the results of training without augmentation. This is further validated by the composition of the filtered dataset, which consists of only a few dozen images compared to the hundreds of synthetic images generated. This indicates that SIF filters out the majority of synthetic images, effectively reverting to a dataset size comparable to the original real samples.
Finally, regarding SD(txt+img) and SG3(lbl), we observe an improvement even when SIF is not applied (around 1–2\%) over the baseline, suggesting that both generative techniques produce high-quality images that improve generalization. Furthermore, these improvements become even more pronounced when SIF is adopted. Specifically, we observe an additional improvement of 0.520\% and 0.608\% when using SD(txt+img) and SG3(lbl), respectively.

\begin{figure*}[b]
\centering
    \begin{subfigure}{0.341\textwidth}
    \centering
    {\tiny ResNet50, Tr. aug.\par}
    \includegraphics[width=\textwidth]{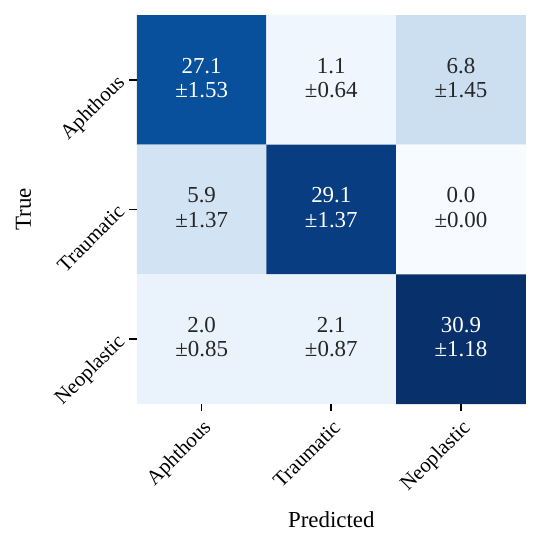}
    \end{subfigure}
    \hspace{5mm}
    \begin{subfigure}{0.341\textwidth}
    \centering
    {\tiny ResNet50, SD(txt+img)+SIF\par}
    \includegraphics[width=\textwidth]{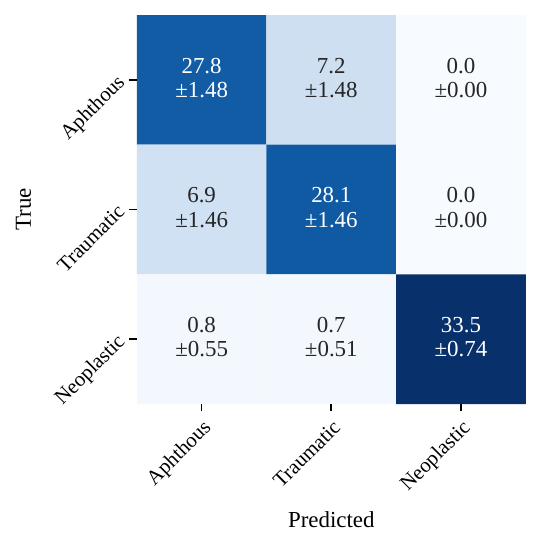}
    \end{subfigure}
    
    \begin{subfigure}{0.341\textwidth}
    \centering
    {\tiny ViT, Tr. aug.\par}
    \includegraphics[width=\textwidth]{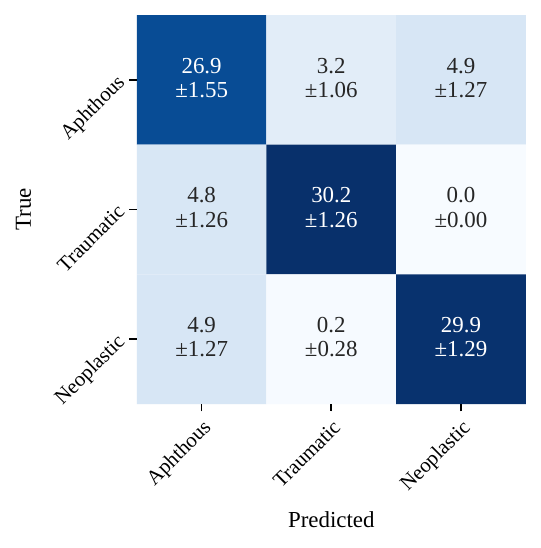}
    \end{subfigure}
    \hspace{5mm}
    \begin{subfigure}{0.341\textwidth}
    \centering
    {\tiny ViT, SD(txt+img)+SIF\par}
    \includegraphics[width=\textwidth]{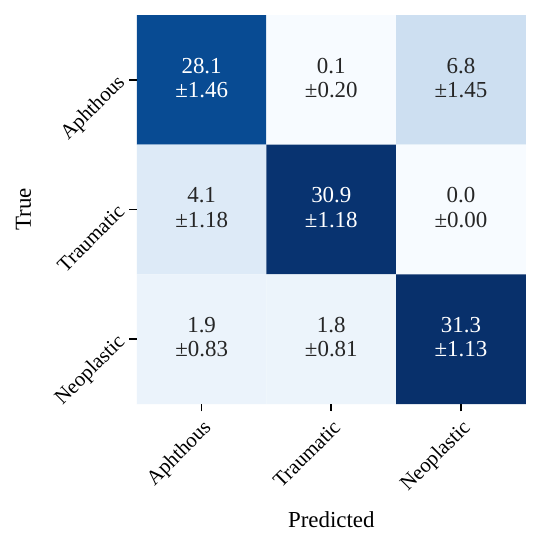}
    \end{subfigure}
\caption{Confusion matrices for PhotoMOCI comparing traditional augmentation and SD(txt+img)+SIF for ResNet50 and ViT.}
\label{fig:photomoci-confusion-matrices}
\end{figure*}

To complement the aggregate metrics reported in Table~\ref{tab:metriche_classificazione_PhotoMOCI}, Figure~\ref{fig:photomoci-confusion-matrices} reports confusion matrices on PhotoMOCI for the traditional augmentation baseline and for the best-performing synthetic generation setup, SD(txt+img)+SIF, using both ResNet50 and ViT. The confidence intervals reported in each cell correspond to the 95\% confidence intervals over the 10 runs. 
The confusion matrices confirm the trend in Table~\ref{tab:metriche_classificazione_PhotoMOCI}: SD(txt+img)+SIF improves neoplastic classification for both ResNet50 (31 to 33 correct predictions) and ViT (30 to 31), while also improving traumatic-lesion classification for ViT (30 to 31). Thus, the performance gain does not come at the cost of neoplastic recognition.

To complement the confidence intervals and assess the statistical significance of the observed improvements, we performed hypothesis tests on the same 10 runs used to compute Tables \ref{tab:metriche_classificazione_PhotoMOCI} and \ref{tab:metriche_classificazione_KOCD}. First, we performed overall non-parametric repeated-measures tests to assess the statistical significance of the observed improvements when introducing SIF. Specifically, for each dataset and classifier, we applied the Friedman test $\boldsymbol{\chi^2}$ across all training setups, considering the 10 repeated runs as paired blocks, following established recommendations for non-parametric comparison of learning methods \cite{demsar2006statistical,garcia2010advanced,derrac2011practical}. Kendall's $W$ was reported as effect size \cite{tomczak2014need}. From Table~\ref{tab:overall_friedman_tests}, we observe that the Friedman test showed a significant overall effect of the training setup on accuracy for both datasets and classifiers ($p<.0001$), with Kendall's $W$ ranging from 0.564 to 0.631, indicating a moderate-to-large effect.

\begin{table}[h]
\scriptsize
\centering
\caption{Overall statistical significance analysis using Friedman tests on accuracy.}
\label{tab:overall_friedman_tests}
{
\begin{tabular}{l l c c c}
\toprule
\textbf{Dataset} & \textbf{Model} & $\boldsymbol{\chi^2}$ & $\boldsymbol{p}$ & \textbf{Kendall's $\boldsymbol{W}$} \\
\midrule
KOCD & ResNet & 54.938 & $<0.0001$ & 0.610 \\
KOCD & ViT & 53.520 & $<0.0001$ & 0.595 \\
PhotoMOCI & ResNet & 56.749 & $<0.0001$ & 0.631 \\
PhotoMOCI & ViT & 50.787 & $<0.0001$ & 0.564 \\
\bottomrule
\end{tabular}
}
\end{table}

We then performed planned paired Wilcoxon signed-rank tests, following recommendations for paired non-parametric comparisons of learning methods \cite{demsar2006statistical,benavoli2016should}. Two comparison families were considered: (i) each synthetic generation setup combined with SIF was compared against traditional augmentation; and (ii) each SIF-filtered setup was compared against its corresponding unfiltered synthetic setup.

From Table~\ref{tab:planned_comparisons_traug_sif}, we observe that, against traditional augmentation, SD(txt+img)+SIF was the most consistent setup, improving accuracy in all four cases with significant effects: KOCD ResNet (+2.38 pp, $p=.0371$), KOCD ViT (+2.08 pp, $p=.0059$), PhotoMOCI ResNet (+1.73 pp, $p=.0273$), and PhotoMOCI ViT (+2.35 pp, $p=.0195$). SG3(lbl)+SIF also improved accuracy in all four cases, but reached significance only for KOCD ResNet (+1.72 pp, $p=.0488$), with a borderline effect on PhotoMOCI ViT (+1.79 pp, $p=.0515$). Conversely, SD(txt)+SIF and AC-SG3+SIF did not consistently surpass traditional augmentation, confirming that the benefit of SIF depends on the quality and conditioning of the synthetic generation setup.

\begin{table*}[t]
\footnotesize
\centering
\setlength{\tabcolsep}{4pt}
\caption{Planned Wilcoxon comparisons between synthetic+SIF setups and traditional augmentation on accuracy. Values report accuracy difference in percentage points and $p$-value.}
\label{tab:planned_comparisons_traug_sif}
{
\begin{tabular}{l cc cc cc cc}
\toprule
& \multicolumn{4}{c}{\textbf{KOCD}} & \multicolumn{4}{c}{\textbf{PhotoMOCI}} \\
\cmidrule(lr){2-5}\cmidrule(lr){6-9}
\textbf{\makecell[l]{Synthetic img.\\gen. setup}} &
\multicolumn{2}{c}{\textbf{ResNet}} &
\multicolumn{2}{c}{\textbf{ViT}} &
\multicolumn{2}{c}{\textbf{ResNet}} &
\multicolumn{2}{c}{\textbf{ViT}} \\
\cmidrule(lr){2-3}\cmidrule(lr){4-5}\cmidrule(lr){6-7}\cmidrule(lr){8-9}
& \textbf{$\Delta$ Acc.} & \textbf{$p$} & \textbf{$\Delta$ Acc.} & \textbf{$p$} & \textbf{$\Delta$ Acc.} & \textbf{$p$} & \textbf{$\Delta$ Acc.} & \textbf{$p$} \\
\midrule
SD(txt)+SIF & -1.49 & 0.0645 & -0.59 & 0.3750 & -2.44 & 0.0371 & -0.84 & 0.4316 \\
SD(txt+img)+SIF & +2.38 & 0.0371 & +2.08 & 0.0059 & +1.73 & 0.0273 & +2.35 & 0.0195 \\
SG3(lbl)+SIF & +1.72 & 0.0488 & +1.17 & 0.0934 & +1.08 & 0.0754 & +1.79 & 0.0515 \\
AC-SG3+SIF & +0.44 & 0.6953 & +0.21 & 0.6250 & -1.53 & 0.0195 & -0.17 & 0.6250 \\
\bottomrule
\end{tabular}
}
\end{table*}

Finally, Table~\ref{tab:planned_comparisons_nonsif_sif} compares each SIF-filtered setup with its corresponding unfiltered synthetic augmentation. SIF produced positive accuracy differences in all 16 planned comparisons. The largest gains were observed for SD(txt), with improvements of +3.05 pp and +3.58 pp on KOCD and +5.92 pp and +6.31 pp on PhotoMOCI, all with $p\leq.0039$. For SD(txt+img), the gains were smaller (+0.48 to +0.80 pp), with significant improvements for both ViT classifiers and borderline results for both ResNet classifiers ($p=.0566$). SG3(lbl) and AC-SG3 also showed mostly positive but sometimes borderline evidence; significant gains were observed for SG3(lbl) on PhotoMOCI ResNet (+0.32 pp, $p=.0250$) and for AC-SG3 on KOCD ViT (+0.91 pp, $p=.0223$), PhotoMOCI ResNet (+1.84 pp, $p=.0195$), and PhotoMOCI ViT (+1.33 pp, $p=.0324$). Overall, although several cases lie close to conventional significance thresholds, the direction of the paired differences is consistently favorable to SIF.

\begin{table*}[t]
\footnotesize
\centering
\setlength{\tabcolsep}{4pt}
\caption{Planned Wilcoxon comparisons between SIF-filtered and unfiltered synthetic augmentation on accuracy. Values report accuracy difference in percentage points and $p$-value.}
\label{tab:planned_comparisons_nonsif_sif}
{
\begin{tabular}{l cc cc cc cc}
\toprule
& \multicolumn{4}{c}{\textbf{KOCD}} & \multicolumn{4}{c}{\textbf{PhotoMOCI}} \\
\cmidrule(lr){2-5}\cmidrule(lr){6-9}
\textbf{\makecell[l]{Synthetic img.\\gen. setup}} &
\multicolumn{2}{c}{\textbf{ResNet}} &
\multicolumn{2}{c}{\textbf{ViT}} &
\multicolumn{2}{c}{\textbf{ResNet}} &
\multicolumn{2}{c}{\textbf{ViT}} \\
\cmidrule(lr){2-3}\cmidrule(lr){4-5}\cmidrule(lr){6-7}\cmidrule(lr){8-9}
& \textbf{$\Delta$ Acc.} & \textbf{$p$} & \textbf{$\Delta$ Acc.} & \textbf{$p$} & \textbf{$\Delta$ Acc.} & \textbf{$p$} & \textbf{$\Delta$ Acc.} & \textbf{$p$} \\
\midrule
SD(txt) & +3.05 & 0.0039 & +3.58 & 0.0020 & +5.92 & 0.0020 & +6.31 & 0.0020 \\
SD(txt+img) & +0.48 & 0.0566 & +0.71 & 0.0316 & +0.60 & 0.0566 & +0.80 & 0.0316 \\
SG3(lbl) & +0.81 & 0.0754 & +0.63 & 0.0750 & +0.32 & 0.0250 & +0.57 & 0.0695 \\
AC-SG3 & +0.74 & 0.0602 & +0.91 & 0.0223 & +1.84 & 0.0195 & +1.33 & 0.0324 \\
\bottomrule
\end{tabular}
}
\end{table*}

\subsection{Ablation}

In this section, we conduct two distinct ablation studies: (i) to evaluate the efficacy of Synthetic Image Filtering by isolating the individual contributions of Synthetic Proxy Classifier and the Synthetic Image Detector and to assess the SID threshold calibration.

\begin{figure*}[t]
    \begin{subfigure}{0.50\textwidth}
    \includegraphics[width=\textwidth]{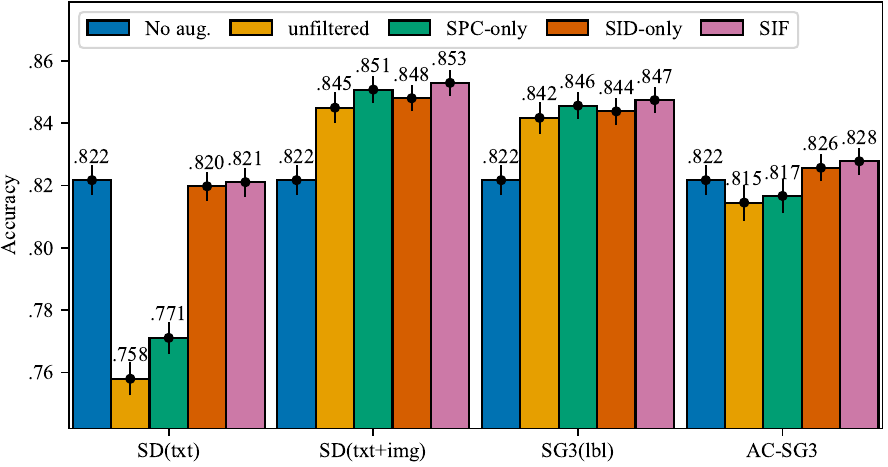}
      \subcaption{}
    \end{subfigure}
    \begin{subfigure}{0.50\textwidth}
    \includegraphics[width=\textwidth]{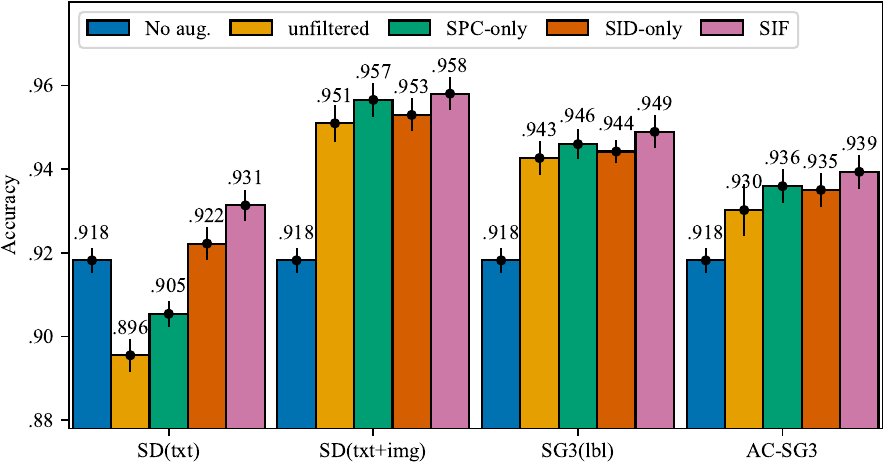}
      \subcaption{}
    \end{subfigure}
\vspace{-3mm}
\caption{Ablation study evaluating the impact of SPC and SID within the SIF design on (a) PhotoMOCI and (b) KOCD using ViT. The study compares training without augmentation (blue bar) to training with integrated synthetic data: unfiltered (ochre bar), filtered using SPC only (water-green bar), SID only (orange bar), and SIF (pink bar).
\label{fig:ablation}
}
\end{figure*}

\textbf{SIF component.} The results for the individual contributions of Synthetic Proxy Classifier and the Synthetic Image Detector are presented in Figure \ref{fig:ablation}(a) and \ref{fig:ablation}(b) for PhotoMOCI and KOCD, respectively.
For a comparative performance analysis, we established as a baseline the training on the small real dataset (\textcolor{RoyalBlue}{blue} bar). We then evaluated four distinct filtering augmentation configurations using ViT as oral lesion classifier, which demonstrated performance trends consistent with the ResNet architecture. The filtering configurations included: (i) an unfiltered dataset generated by completely omitting SIF (\textcolor{Tan}{ochre} bar); (ii) a filtered dataset utilizing only the SPC component (\textcolor{teal}{water-green} bar); (iii) a filtered dataset utilizing only the SID component (\textcolor{RedOrange}{orange} bar); (iv) the complete SIF pipeline integrating both auxiliary classifiers (\textcolor{CarnationPink}{pink} bar).

From the figures, we observe that the SD(txt+img) and SG3(lbl) configurations exhibit consistent behavior across both PhotoMOCI and KOCD. As previously discussed, the introduction of synthetic data in these settings provides a benefit even in the absence of SIF. Nevertheless, the adoption of SIF further yields marginal accuracy improvements, with SPC contributing more than SID. This indicates that, when synthetic images are generally of good quality, selecting samples through Synthetic Proxy Classifier that contribute to learning more discriminative class boundaries is crucial for improving classifier performance. Conversely, in scenarios where the introduction of synthetic data leads to performance degradation (SD(txt) on both PhotoMOCI and KOCD, and AC-SG3 on PhotoMOCI), an opposite trend emerges. In such cases, the primary gains are provided by SID, suggesting that low realism and visually implausible synthetic images are the main source of noise responsible for the observed drop in performance, and that their removal mitigates negative effects.

\textbf{SID threshold calibration.} To assess the sensitivity of SIF to the decision threshold used by the Synthetic Image Detector, we conducted an additional calibration experiment by varying the SID threshold $\tau$ in the interval $[0,1]$ with a step of $0.01$. The analysis was performed for both datasets, both classifiers (ResNet50 and ViT), and all synthetic data augmentation setups. For each threshold value, we derived the corresponding retained synthetic images and evaluated the final classification accuracy. The value $\tau=0$ corresponds to training without synthetic images, $\tau=0.5$ corresponds to the default SIF configuration, and $\tau=1$ corresponds to the unfiltered synthetic augmentation setting. Confidence bands report the 95\% confidence intervals estimated over the same 10 runs, consistent with all the other experiments where we repeated multiple runs.

\begin{figure*}[t]
\centering
    \begin{subfigure}{0.75\textwidth}
    \includegraphics[width=\textwidth]{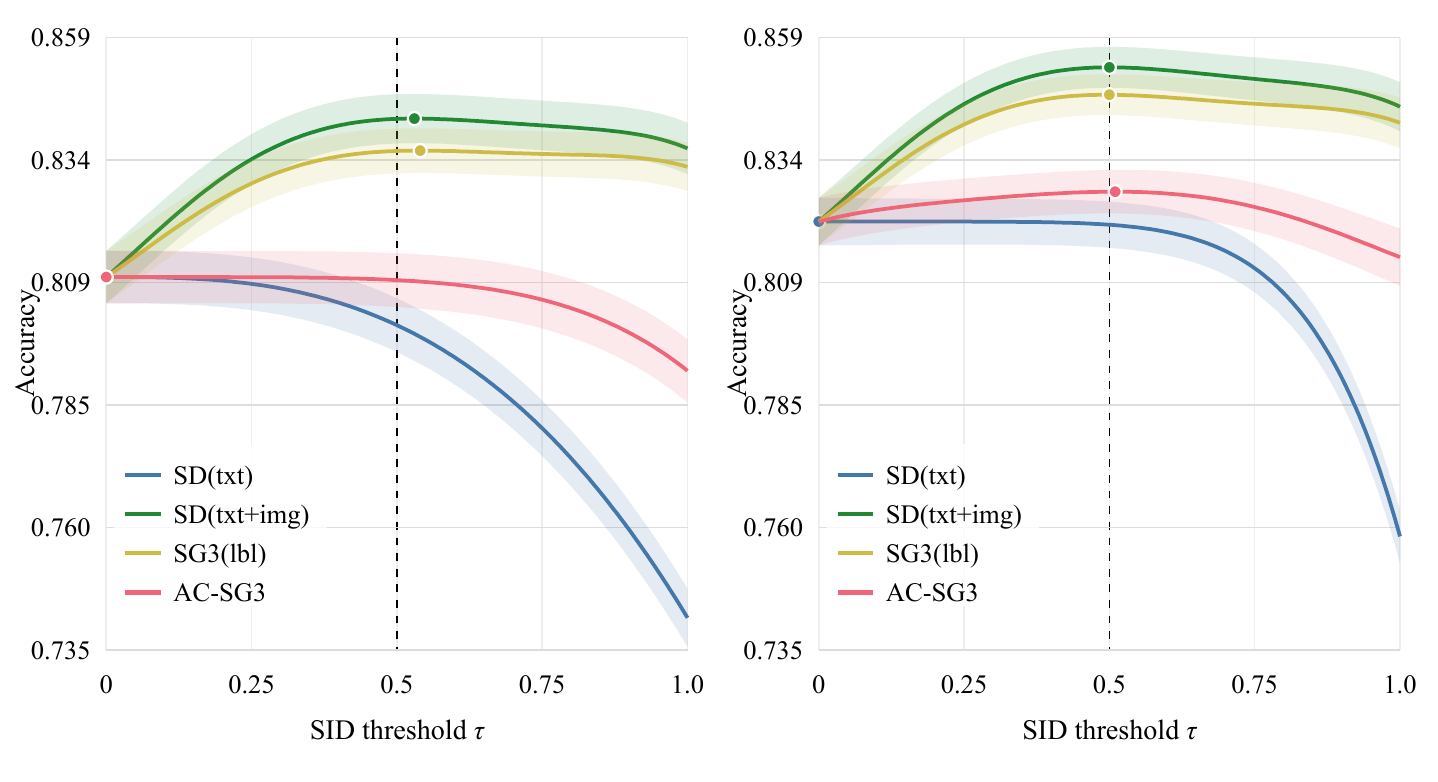}
      \subcaption{}
    \end{subfigure}
    
    \begin{subfigure}{0.75\textwidth}
    \includegraphics[width=\textwidth]{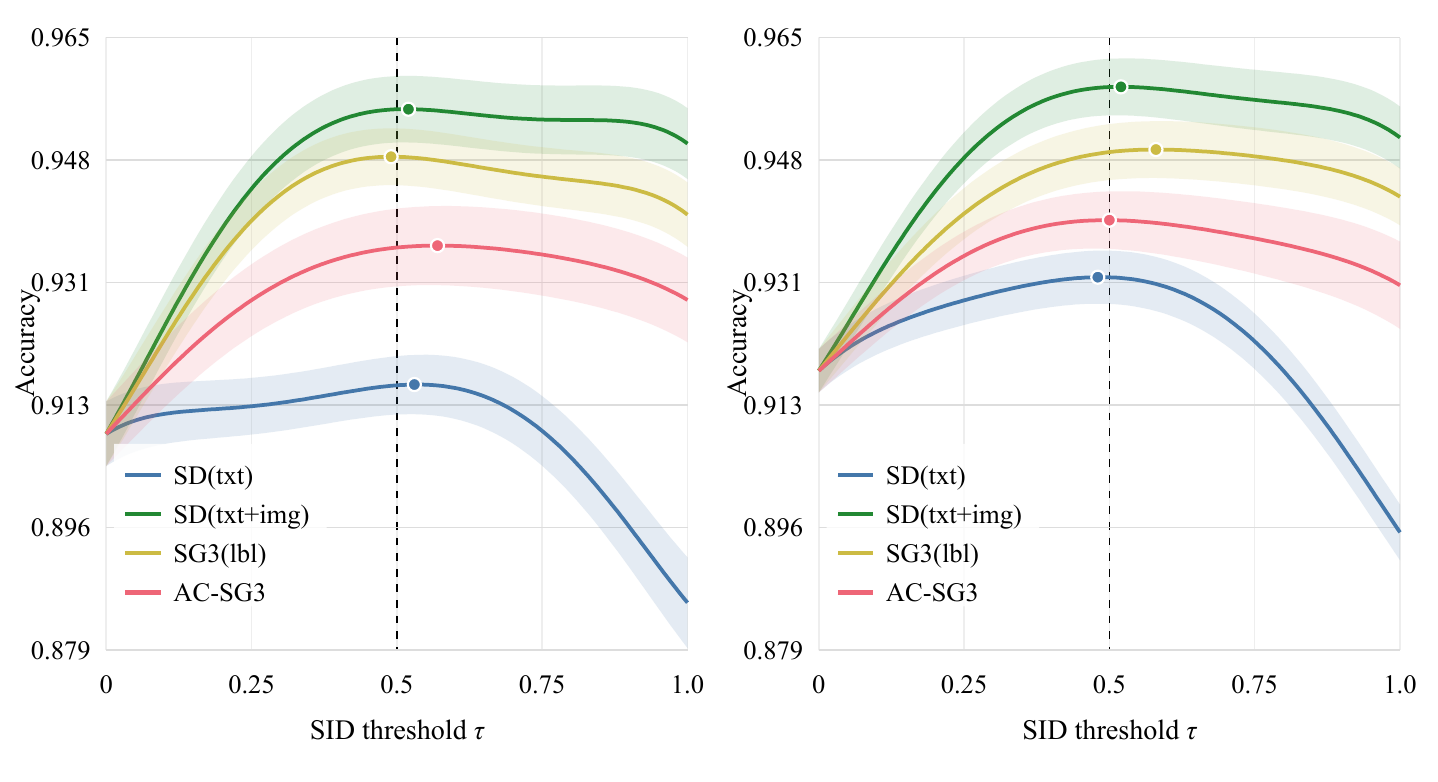}
      \subcaption{}
    \end{subfigure}

\caption{SID threshold calibration on (a) PhotoMOCI and (b) KOCD. Each plot reports accuracy as a function of the SID threshold $\tau$ for ResNet50 (left) and ViT (right), with shaded 95\% confidence intervals over 10 runs and a marker indicating the best threshold.}
\label{fig:sid-threshold-calibration}
\end{figure*}

Figure~\ref{fig:sid-threshold-calibration} summarizes the calibration analysis and shows that the final classification performance is stable around the threshold value $\tau=0.5$. 
Additionally, the limited overlap of confidence intervals among different synthetic data augmentation setups indicates that the relative differences among setups are preserved across the threshold sweep. 
In the few PhotoMOCI cases where the no-augmentation baseline is already higher than the corresponding SIF result, namely ResNet50 with SD(txt), ResNet50 with AC-SG3, and ViT with SD(txt), the maximum is attained at $\tau=0$, consistently with the values reported in Table~\ref{tab:metriche_classificazione_PhotoMOCI}.

\section{Discussion}\label{sec:discussion}

A preliminary consideration about current photographic oral cancer datasets concerns the practical relevance of binary cancer vs non-cancer (healthy) classifications for photographic screening. Although such datasets support controlled experimentation, especially for benchmarking, they do not reflect real clinical workflows when the negative class consists of healthy mucosa. Physicians would not capture images when no suspicious lesion is visible; rather, photographic assessment is typically relevant after a lesion has already been observed, and the clinical question becomes how to prioritize, characterize, or refer it. Consequently, binary datasets may overstate their utility and fail to represent the operational context in which computer-assisted tools are intended to function. In this sense, the three-class PhotoMOCI formulation, distinguishing aphthous, traumatic, and neoplastic lesions, better approximates a lesion-level triage scenario, where the model must discriminate among visible pathological findings instead of separating pathology from healthy tissue.

Summarizing the experiment results, traditional data augmentation consistently provides solid performance and remains a safe and reliable baseline strategy.
Regarding synthetic data augmentation strategies, the effectiveness of the proposed SIF strongly depends on the quality of the images generated during the earlier Synthetic Image Generation step of the proposed methodology. If the majority of the generated images are of low quality, SIF filters out most samples, resulting in a training setup that closely resembles training without augmentation on small real datasets. This behavior was observed in the ablation study, specifically, in the setup involving SD(txt) on the PhotoMOCI dataset.

Although the original datasets are characterized by a relatively balanced class, the analysis reported in Table~\ref{tab:SIF_retained} shows that the SIF preserves this balance after filtering, without favoring any particular pathology. More generally, class imbalance, whether given by the original dataset distributions or introduced during synthetic image generation, is not a limitation of the proposed pipeline. If required, sample discarding strategies during a next step can be adopted to increase the number of retained samples for under-represented categories.

Therefore, the adoption of generative models does not inherently guarantee performance improvements when synthetic samples are incorporated without further selection. The benefit of synthetic data augmentation depends primarily on two factors: (i) the architecture of the generative model and the adopted conditioning mechanism. In some cases, fine-tuning these models can be unstable or may lead to generators that fail to generalize properly, even when trained on limited datasets, as is typical for oral photographic datasets. This phenomenon is evident in the qualitative samples produced by SD(txt) and AC-SG3 in Figure~\ref{fig:SynthPhotoMOCI-examples}, as well as by SD(txt) in Figure~\ref{fig:SynthKOCD-examples}. (ii) The nature of the data, which may be highly granular, as in the case of the PhotoMOCI dataset. Such granularity increases the difficulty of generating realistic samples that faithfully capture the distinctive visual characteristics of the target classes.


Based on these considerations, we outline here some practical guidelines for the usage of the proposed methodology illustrated in Figure~\ref{fig:oral-synthetic-scema}. Specifically, this methodology should not be considered as a universal plug-and-play pipeline capable of returning an optimal oral lesion classifier. Instead, given the numerous factors affecting DL classifier training (such as the number of classes, class imbalance, single-center data collection, acquisition protocol, limited sample size, or class granularity), it should be considered as a set of guidelines for carefully evaluating each stage of the pipeline before proceeding to the next one. Indeed, these aspects may introduce dataset- and acquisition-specific biases, and should be considered when interpreting the results.

The proposed training pipeline presents some drawbacks compared to DL training based on traditional data augmentation techniques involving geometric and photometric transformations. These drawbacks mainly relate to increased computational requirements and additional human effort.
While traditional data augmentation relies on training a single classifier, where augmentations are typically applied at runtime and are well supported by modern DL libraries, the proposed methodology presents the following practical considerations:

\begin{itemize}
    \item It requires training two additional auxiliary classifiers, SPC and SID, which implement SIF, as well as fine-tuning a generative model, which represents the most computationally demanding component.
    \item About the development of the generative architecture, before proceeding to filtering and training the oral lesion classifier, it is essential to perform a preliminary evaluation of the generated images. This evaluation should ensure an appropriate balance between fidelity to the original image distribution (assessed using FID and KID metrics) and variability (measured through Precision, Recall, and Coverage), where generating samples that cover multiple patterns and cases facilitates the generalization of the final classifier.
    \item The development of SIF involves the training of both SPC and SID. In this context, achieving a balance between the high performance of SPC in addressing the same downstream classification task on synthetic images and the near-random performance of SID in failing to distinguish real from synthetic images is crucial. 
    As shown in Table~\ref{tab:SIF_metrics}, SD(txt) clearly exhibits difficulties in generating high-quality synthetic images. The same considerations are less intuitive if we compare SD(txt+img) and SG3(lbl) on PhotoMOCI. The first one gets higher SPC accuracy, while the second one achieves a better (lower) accuracy value on SID. 
    \item The methodology is more suitable for a two-phase training strategy, in which the Filtered Dataset is precomputed after the Synthetic Image Generation phase and the application of SIF. Generating synthetic samples and evaluating them with SIF on-the-fly during training would be particularly challenging, especially for diffusion-based models, whose inference process involves multiple diffusion steps.
    As reported in Table~\ref{tab:generation_runtime}, generating the complete SD(txt+img)-based image pool requires around 28 and 41 min for PhotoMOCI and KOCD, respectively, whereas ResNet50-based SIF filtering requires only 10.1 ms per image; thus, once generation is completed offline, filtering adds minimal overhead to dataset precomputation.
    \item Despite recent advances aimed at stabilizing training through regularization techniques and reducing computational costs using approaches such as LoRA \cite{yang2024low}, training a generative model remains challenging, particularly in limited-data scenarios. This often results in a time-consuming process, where substantial effort is devoted to hyperparameter tuning before obtaining a satisfactory generative model.
\end{itemize}

Under the assumption that sufficient computational resources are available and that multiple training trials, or preferably a systematic hyperparameter optimization procedure, can be conducted, the proposed methodology represents the most effective strategy among those evaluated in our benchmark for improving classification performance under the considered experimental setting, as indicated in Tables \ref{tab:metriche_classificazione_PhotoMOCI} and \ref{tab:metriche_classificazione_KOCD}.
However, when these conditions cannot be met, traditional data augmentation remains the preferred solution, as it is well supported by existing DL libraries, lightweight in terms of computational overhead, and can be efficiently applied at runtime during model training.

\section{Conclusion}\label{sec:conclusion}

This work addresses the challenges of data scarcity and investigates the synthetic data augmentation in photographic oral cancer. We introduced PhotoMOCI, a novel dataset designed to support multiple diagnostic tasks in the oral cancer domain, providing a publicly available resource for open research and benchmarking.
We also conduct a comprehensive evaluation of data augmentation strategies, comparing traditional transformation-based methods with generative approaches based on GANs and Diffusion Models. 
Our findings indicate that the direct use of synthetic images does not consistently improve classification performance and may even be counterproductive, emphasizing that synthetic data quality is critical in this setting.
To address this limitation, we proposed the \textbf{Synthetic Image Filter}, which selects synthetic samples based on machine-assessed visual realism and class representativeness using two auxiliary models: Synthetic Proxy Classifier and Synthetic Image Detector. Although qualitative inspection shows visual analogies between some retained synthetic images and real oral lesions, the term realism in this work refers specifically to machine indistinguishability from the real-image distribution, rather than to clinical realism or diagnostic plausibility. Experimental results show that this selective strategy effectively removes low-utility images and leads to more stable and improved downstream performance.

Overall, this study highlights the importance of careful dataset design and controlled use of generative augmentation in photographic oral cancer imaging, offering a comprehensive benchmark for future research on reliable data augmentation pipelines.
At the same time, the reported improvements should be regarded as dataset-specific experimental findings. The single-center origin of PhotoMOCI and the absence of an human expert assessment of the SIF-selected images represent a limitation; external and clinical validation would be strongly recommended before assessing its applicability in real-world screening workflows.


\subsection{Future work}

The proposed pipeline enables a multi-level evaluation of synthetic images and their impact when used as data augmentation for downstream tasks such as image classification. 
Specifically, (i) assessing generative image metrics at the Synthetic Image Generation stage, (ii) employing auxiliary classifiers assess image realism and class representativeness in the Synthetic Image Filter, while (iii) the final evaluation considers the effect of selected synthetic data on downstream performance. 
Given the importance of trustworthiness when deploying AI in high-risk domains such as healthcare, model transparency and alignment with human judgment are crucial \cite{longo2024explainable, parola2026human, rong2023towards}. 

\backmatter

\bibliography{sn-bibliography}

@misc{sagers2023augmentingmedicalimageclassifiers,
      title={Augmenting medical image classifiers with synthetic data from latent diffusion models}, 
      author={Luke W. Sagers and James A. Diao and Luke Melas-Kyriazi and Matthew Groh and Pranav Rajpurkar and Adewole S. Adamson and Veronica Rotemberg and Roxana Daneshjou and Arjun K. Manrai},
      year={2023},
      eprint={2308.12453},
      archivePrefix={arXiv},
      primaryClass={cs.CV},
      url={https://arxiv.org/abs/2308.12453}, 
}

@Article{alosaimi2022efficient,
AUTHOR = {Wael Alosaimi, M. Irfan Uddin},
TITLE = {Efficient Data Augmentation Techniques for Improved Classification in Limited Data Set of Oral Squamous Cell Carcinoma},
JOURNAL = {Computer Modeling in Engineering \& Sciences},
VOLUME = {131},
YEAR = {2022},
NUMBER = {3},
PAGES = {1387--1401},
URL = {http://www.techscience.com/CMES/v131n3/47382},
ISSN = {1526-1506},
DOI = {10.32604/cmes.2022.018433}
}

@article{oya2023diagnosis,
title = {Oral squamous cell carcinoma diagnosis in digitized histological images using convolutional neural network},
journal = {Journal of Dental Sciences},
volume = {18},
number = {1},
year = {2023},
issn = {1991-7902},
doi = {https://doi.org/10.1016/j.jds.2022.08.017},
url = {https://www.sciencedirect.com/science/article/pii/S1991790222002008},
author = {Kaori Oya and Kazuma Kokomoto and Kazunori Nozaki and Satoru Toyosawa},
}

@inproceedings{karras2019style,
  title={A style-based generator architecture for generative adversarial networks},
  author={Karras, Tero and Laine, Samuli and Aila, Timo},
  booktitle={Proceedings of the IEEE/CVF conference on computer vision and pattern recognition},
  year={2019}
}

@inproceedings{liu2020diverse,
  title={Diverse image generation via self-conditioned gans},
  author={Liu, Steven and Wang, Tongzhou and Bau, David and Zhu, Jun-Yan and Torralba, Antonio},
  booktitle={Proceedings of the IEEE/CVF conference on computer vision and pattern recognition},
  pages={14286--14295},
  year={2020}
}

@article{bourou2024gans,
  title={GANs Conditioning Methods: A Survey},
  author={Bourou, Anis and Mezger, Val{\'e}rie and Genovesio, Auguste},
  journal={arXiv preprint arXiv:2408.15640},
  year={2024}
}

@inproceedings{odena2017conditional,
  title={Conditional image synthesis with auxiliary classifier gans},
  author={Odena, Augustus and Olah, Christopher and Shlens, Jonathon},
  booktitle={International conference on machine learning},
  pages={2642--2651},
  year={2017},
  organization={PMLR}
}

@inproceedings{FujimotoAutomaticOralDiagnosis,
  author    = {Fujimoto, Taro and Fukuzawa, Eiji and Tatehara, Seiko and Satomura, Kazuhito and Ohya, Jun},
  booktitle = {2022 44th Annual International Conference of the IEEE Engineering in Medicine \& Biology Society (EMBC)},
  title     = {Automatic Diagnosis of Early-Stage Oral Cancer and Precancerous Lesions from ALA-PDD Images Using GAN and CNN},
  year      = {2022},
  pages     = {2161--2164},
  doi       = {10.1109/EMBC48229.2022.9871868}
}

@article{ribeiro2022assessment,
  title={Assessment of screening programs as a strategy for early detection of oral cancer: a systematic review},
  author={Ribeiro, Marcela Ferreira Abrahao and Oliveira, Maria Clara Moreira and Leite, Alice Carvalho and Bruzinga, Fabio Fernandes Borem and Mendes, Polianne Alves and Grossmann, Soraya de Mattos Camargo and de Ara{\'u}jo, V{\^a}nia Eloisa and Souto, Giovanna Ribeiro},
  journal={Oral Oncology},
  volume={130},
  year={2022},
  publisher={Elsevier}
}

@article{dossantos2023influence,
  author  = {Dos Santos, Dal{\'i} F. D. and de Faria, Paulo R. and Traven{\c c}olo, Bruno A. N. and do Nascimento, Marcelo Z.},
  year    = {2023},
  title   = {Influence of Data Augmentation Strategies on the Segmentation of Oral Histological Images Using Fully Convolutional Neural Networks},
  journal = {Journal of Digital Imaging},
  volume  = {36},
  number  = {4},
  doi     = {10.1007/s10278-023-00814-z}
}

@misc{qi2020saggansemisupervisedattentionguidedgans,
  title        = {SAG-GAN: Semi-Supervised Attention-Guided GANs for Data Augmentation on Medical Images},
  author       = {Chang Qi and Junyang Chen and Guizhi Xu and Zhenghua Xu and Thomas Lukasiewicz and Yang Liu},
  year         = {2020},
  eprint       = {2011.07534},
  archivePrefix= {arXiv},
  primaryClass = {eess.IV},
  url          = {https://arxiv.org/abs/2011.07534}
}

@article{Waheed_2020,
    title     = {CovidGAN: Data Augmentation Using Auxiliary Classifier GAN for Improved Covid-19 Detection},
    volume    = {8},
    ISSN      = {2169-3536},
    url       = {http://dx.doi.org/10.1109/ACCESS.2020.2994762},
    DOI       = {10.1109/access.2020.2994762},
    journal   = {IEEE Access},
    publisher = {Institute of Electrical and Electronics Engineers (IEEE)},
    author    = {Waheed, Abdul and Goyal, Muskan and Gupta, Deepak and Khanna, Ashish and Al-Turjman, Fadi and Pinheiro, Placido Rogerio},
    year      = {2020},
}

@inproceedings{neff2017gan,
  author    = {Neff, Thomas and Payer, Christian and {\v S}tern, Darko and Urschler, Martin},
  year      = {2017},
  month     = {05},
  pages     = {140--145},
  title     = {Generative Adversarial Network based Synthesis for Supervised Medical Image Segmentation},
  booktitle = {Proceedings of the OAGM \& ARW Joint Workshop on Vision, Automation and Robotics},
  doi       = {10.3217/978-3-85125-524-9-30},
  publisher = {Verlag der TU Graz}
}

@misc{qasim2021redganattackingclassimbalance,
      title        = {Red-GAN: Attacking class imbalance via conditioned generation. Yet another perspective on medical image synthesis for skin lesion dermoscopy and brain tumor MRI},
      author       = {Ahmad B Qasim and Ivan Ezhov and Suprosanna Shit and Oliver Schoppe and Johannes C Paetzold and Anjany Sekuboyina and Florian Kofler and Jana Lipkova and Hongwei Li and Bjoern Menze},
      year         = {2021},
      eprint       = {2004.10734},
      archivePrefix= {arXiv},
      primaryClass = {eess.IV},
      url          = {https://arxiv.org/abs/2004.10734}
}

@inproceedings{dhariwal2021diffusion,
 author = {Dhariwal, Prafulla and Nichol, Alexander},
 booktitle = {Advances in Neural Information Processing Systems},
 editor = {M. Ranzato and A. Beygelzimer and Y. Dauphin and P.S. Liang and J. Wortman Vaughan},
 pages = {8780--8794},
 publisher = {Curran Associates, Inc.},
 title = {Diffusion Models Beat GANs on Image Synthesis},
 url = {\url{https://proceedings.neurips.cc/paper_files/paper/2021/file/49ad23d1ec9fa4bd8d77d02681df5cfa-Paper.pdf}},
 volume = {34},
 year = {2021}
}

@article{Frid_Adar_2018,
   title={GAN-based synthetic medical image augmentation for increased CNN performance in liver lesion classification},
   volume={321},
   ISSN={0925-2312},
   url={http://dx.doi.org/10.1016/j.neucom.2018.09.013},
   DOI={10.1016/j.neucom.2018.09.013},
   journal={Neurocomputing},
   publisher={Elsevier BV},
   author={Frid-Adar, Maayan and Diamant, Idit and Klang, Eyal and Amitai, Michal and Goldberger, Jacob and Greenspan, Hayit},
   year={2018},
   month=dec, pages={321–331} }

@article{hinton2006reducing,
  author    = {Hinton, Geoffrey E. and Salakhutdinov, Ruslan R.},
  title     = {Reducing the dimensionality of data with neural networks},
  journal   = {Science},
  volume    = {313},
  number    = {5786},
  pages     = {},
  year      = {2006},
  doi       = {10.1126/science.1127647},
  publisher = {American Association for the Advancement of Science},
  url       = {https://doi.org/10.1126/science.1127647}
}

@misc{mirza2014conditionalgenerativeadversarialnets,
      title={Conditional Generative Adversarial Nets}, 
      author={Mehdi Mirza and Simon Osindero},
      year={2014},
      eprint={1411.1784},
      archivePrefix={arXiv},
      primaryClass={cs.LG},
      url={https://arxiv.org/abs/1411.1784}, 
}

@article{karras2021alias,
  title={Alias-free generative adversarial networks},
  author={Karras, Tero and Aittala, Miika and Laine, Samuli and H{\"a}rk{\"o}nen, Erik and Hellsten, Janne and Lehtinen, Jaakko and Aila, Timo},
  journal={Advances in neural information processing systems},
  volume={34},
  pages={852--863},
  year={2021}
}

@article{Johnson2019,
  author  = {Johnson, Justin M. and Khoshgoftaar, Taghi M.},
  title   = {Survey on deep learning with class imbalance},
  journal = {Journal of Big Data},
  volume  = {6},
  number  = {1},
  year    = {2019},
  pages   = {27},
  doi     = {10.1186/s40537-019-0192-5},
  url     = {https://doi.org/10.1186/s40537-019-0192-5}
}

@article{Zhang2023DLmedImages,
  author         = {Zhang, Huanhuan and Qie, Yufei},
  title          = {Applying Deep Learning to Medical Imaging: A Review},
  journal        = {Applied Sciences},
  volume         = {13},
  year           = {2023},
  number         = {18},
  article-number = {10521},
  url            = {https://www.mdpi.com/2076-3417/13/18/10521},
  issn           = {2076-3417},
  doi            = {10.3390/app131810521}
}

@misc{whang2022datacollectionqualitychallenges,
  title         = {Data Collection and Quality Challenges in Deep Learning: A Data-Centric AI Perspective},
  author        = {Steven Euijong Whang and Yuji Roh and Hwanjun Song and Jae-Gil Lee},
  year          = {2022},
  eprint        = {2112.06409},
  archiveprefix = {arXiv},
  primaryclass  = {cs.LG},
  url           = {https://arxiv.org/abs/2112.06409}
}

@article{alzubaiDataScarsity,
  author  = {Alzubaidi, Laith and Bai, Jinshuai and Al-Sabaawi, Aiman and Santamaría, Jose and Albahri, A.s and Al-dabbagh, Bashar and Fadhel, Mohammed and Manoufali, Mohamed and Zhang, Jinglan and Al-Timemy, Ali and Duan, Ye and Abdullah, Amjed and Farhan, Laith and Lu, Yi and Gupta, Ashish and Albu, Felix and Abbosh, Amin and Gu, Yuantong},
  year    = {2023},
  month   = {04},
  pages   = {},
  title   = {A survey on deep learning tools dealing with data scarcity: definitions, challenges, solutions, tips, and applications},
  volume  = {10},
  journal = {Journal of Big Data},
  doi     = {10.1186/s40537-023-00727-2}
}

@inproceedings{ronneberger2015u,
  author    = {Ronneberger, Olaf
               and Fischer, Philipp
               and Brox, Thomas},
  editor    = {Navab, Nassir
               and Hornegger, Joachim
               and Wells, William M.
               and Frangi, Alejandro F.},
  title     = {U-Net: Convolutional Networks for Biomedical Image Segmentation},
  booktitle = {Medical Image Computing and Computer-Assisted Intervention -- MICCAI 2015},
  year      = {2015},
  publisher = {Springer International Publishing},
  address   = {Cham},
  pages     = {234--241},
  isbn      = {978-3-319-24574-4}
}

@article{shin2016deep,
  author   = {Shin, Hoo-Chang and Roth, Holger R. and Gao, Mingchen and Lu, Le and Xu, Ziyue and Nogues, Isabella and Yao, Jianhua and Mollura, Daniel and Summers, Ronald M.},
  journal  = {IEEE Transactions on Medical Imaging},
  title    = {Deep Convolutional Neural Networks for Computer-Aided Detection: CNN Architectures, Dataset Characteristics and Transfer Learning},
  year     = {2016},
  volume   = {35},
  number   = {5},
  pages    = {1285-1298},
  doi      = {10.1109/TMI.2016.2528162}
}

@InProceedings{tortora2025gan,
    author="Mantegna, Massimiliano
    and Tronchin, Lorenzo
    and Tortora, Matteo
    and Soda, Paolo",
    editor="Palaiahnakote, Shivakumara
    and Schuckers, Stephanie
    and Ogier, Jean-Marc
    and Bhattacharya, Prabir
    and Pal, Umapada
    and Bhattacharya, Saumik",
    title="Benchmarking GAN-Based vs Classical Data Augmentation on Biomedical Images",
    booktitle="Pattern Recognition. ICPR 2024 International Workshops and Challenges",
    year="2025",
    publisher="Springer Nature Switzerland",
    address="Cham",
    pages="92--104",
    isbn="978-3-031-87660-8"
}

@article{rofena2025lesion,
  title={Lesion-Aware Generative Artificial Intelligence for Virtual Contrast-Enhanced Mammography in Breast Cancer},
  author={Rofena, Aurora and Manchia, Arianna and Piccolo, Claudia Lucia and Zobel, Bruno Beomonte and Soda, Paolo and Guarrasi, Valerio},
  journal={arXiv preprint arXiv:2505.03018},
  year={2025}
}

@article{rofena2025augmented,
  title={Augmented Intelligence for Multimodal Virtual Biopsy in Breast Cancer Using Generative Artificial Intelligence},
  author={Rofena, Aurora and Piccolo, Claudia Lucia and Zobel, Bruno Beomonte and Soda, Paolo and Guarrasi, Valerio},
  journal={preprint arXiv:2501.19176},
  year={2025}
}

@article{kim2020increasing,
  title     = {Increasing incidence and improving survival of oral tongue squamous cell carcinoma},
  author    = {Kim, Yi-Jun and Kim, Jin Ho},
  journal   = {Scientific reports},
  volume    = {10},
  number    = {1},
  pages     = {7877},
  year      = {2020},
  publisher = {Nature Publishing Group UK London}
}

@article{zhou2022ru,
  title     = {Ru (II)-modified TiO2 nanoparticles for hypoxia-adaptive photo-immunotherapy of oral squamous cell carcinoma},
  author    = {Zhou, Jia-Ying and Wang, Wen-Jin and Zhang, Chen-Yu and Ling, Yu-Yi and Hong, Xiao-Jing and Su, Qiao and Li, Wu-Guo and Mao, Zong-Wan and Cheng, Bin and Tan, Cai-Ping and others},
  journal   = {Biomaterials},
  volume    = {289},
  pages     = {121757},
  year      = {2022},
  publisher = {Elsevier},
  maxnames  = {2}
}

@article{bramati2021early,
  title     = {Early diagnosis of oral squamous cell carcinoma may ensure better prognosis: A case series},
  author    = {Bramati, Chiara and Abati, Silvio and Bondi, Stefano and Lissoni, Alessandra and Arrigoni, Gianluigi and Filipello, Federica and Trimarchi, Matteo},
  journal   = {Clinical Case Reports},
  volume    = {9},
  number    = {10},
  year      = {2021},
  publisher = {Wiley-Blackwell}
}

@article{rajpurkar2022ai,
  title     = {AI in health and medicine},
  author    = {Rajpurkar, Pranav and Chen, Emma and Banerjee, Oishi and Topol, Eric J},
  journal   = {Nature medicine},
  volume    = {28},
  number    = {1},
  pages     = {31--38},
  year      = {2022},
  publisher = {Nature Publishing Group US New York}
}

@inproceedings{lee2021survey,
  title     = {A Survey of Smart Healthcare for the Elderly based on User Requirements and Supply Accessibility},
  author    = {Lee, Ching Hung and Zhang, Zehao and Zhao, Xuejiao},
  booktitle = {5th International Conference on Crowd Science and Engineering},
  year      = {2021}
}

@article{di2024automated,
  title     = {Automated Detection of Oral Malignant Lesions Using Deep Learning: Scoping Review and Meta-Analysis},
  author    = {Di Fede, Olga and La Mantia, Gaetano and Parola, Marco and Maniscalco, Laura and Matranga, Domenica and Tozzo, Pietro and Campisi, Giuseppina and Cimino, Mario GCA},
  journal   = {Oral Diseases},
  year      = {2024},
  publisher = {Wiley Online Library}
}

@article{hu2021lora,
  author       = {Edward J. Hu and
                  Yelong Shen and
                  Phillip Wallis and
                  Zeyuan Allen{-}Zhu and
                  Yuanzhi Li and
                  Shean Wang and
                  Weizhu Chen},
  title        = {LoRA: Low-Rank Adaptation of Large Language Models},
  journal      = {CoRR},
  volume       = {abs/2106.09685},
  year         = {2021},
  url          = {https://arxiv.org/abs/2106.09685},
  eprinttype    = {arXiv},
  eprint       = {2106.09685},
  bibsource    = {dblp computer science bibliography, https://dblp.org}
}

@article{yang2024low,
  title={Low-rank adaptation for foundation models: A comprehensive review},
  author={Yang, Menglin and Chen, Jialin and Zhang, Yifei and Liu, Jiahong and Zhang, Jiasheng and Ma, Qiyao and Verma, Harshit and Zhang, Qianru and Zhou, Min and King, Irwin and others},
  journal={preprint arXiv:2501.00365},
  year={2024}
}

@article{zhan2024conditional,
  title={Conditional image synthesis with diffusion models: A survey},
  author={Zhan, Zheyuan and Chen, Defang and Mei, Jian-Ping and Zhao, Zhenghe and Chen, Jiawei and Chen, Chun and Lyu, Siwei and Wang, Can},
  journal={arXiv preprint arXiv:2409.19365},
  year={2024}
}

@inproceedings{krizhevsky2012imagenet,
  author    = {Alex Krizhevsky and Ilya Sutskever and Geoffrey E. Hinton},
  title     = {ImageNet Classification with Deep Convolutional Neural Networks},
  booktitle = {Advances in Neural Information Processing Systems (NeurIPS)},
  volume    = {25},
  year      = {2012},
  url       = {https://papers.nips.cc/paper/4824-imagenet-classification-with-deep-convolutional-neural-networks.pdf}
}

@article{kingma2013auto,
  author  = {Diederik P. Kingma and Max Welling},
  title   = {Auto-Encoding Variational Bayes},
  journal = {arXiv preprint arXiv:1312.6114},
  year    = {2013},
  url     = {https://arxiv.org/abs/1312.6114}
}

@article{goodfellow2014generative,
  author  = {Ian J. Goodfellow and others},
  title   = {Generative Adversarial Networks},
  journal = {arXiv preprint arXiv:1406.2661},
  year    = {2014},
  url     = {https://arxiv.org/abs/1406.2661}
}

@article{radford2015unsupervised,
  author  = {Alec Radford and Luke Metz and Soumith Chintala},
  title   = {Unsupervised Representation Learning with Deep Convolutional Generative Adversarial Networks},
  journal = {arXiv preprint arXiv:1511.06434},
  year    = {2015},
  url     = {https://arxiv.org/abs/1511.06434}
}

@inproceedings{sohl2015deep,
  author    = {Jascha Sohl-Dickstein and Eric A. Weiss and Niru Maheswaranathan and Surya Ganguli},
  title     = {Deep Unsupervised Learning using Nonequilibrium Thermodynamics},
  booktitle = {Proceedings of the 32nd International Conference on Machine Learning (ICML)},
  pages     = {2256--2265},
  volume    = {37},
  year      = {2015},
  url       = {https://arxiv.org/abs/1503.03585}
}

@inproceedings{arjovsky2017wasserstein,
  author    = {Martin Arjovsky and Soumith Chintala and L{\'e}on Bottou},
  title     = {Wasserstein Generative Adversarial Networks},
  booktitle = {Proceedings of the 34th International Conference on Machine Learning (ICML)},
  volume    = {70},
  pages     = {214--223},
  year      = {2017},
  url       = {https://proceedings.mlr.press/v70/arjovsky17a.html}
}

@inproceedings{ho2020denoising,
  author    = {Jonathan Ho and Ajay Jain and Pieter Abbeel},
  title     = {Denoising Diffusion Probabilistic Models},
  booktitle = {Advances in Neural Information Processing Systems (NeurIPS)},
  volume    = {33},
  pages     = {6840--6851},
  year      = {2020},
  url       = {https://arxiv.org/abs/2006.11239}
}

@inproceedings{karras2020analyzing,
  author    = {Tero Karras and others},
  title     = {Analyzing and Improving the Image Quality of StyleGAN},
  booktitle = {Proceedings of the IEEE/CVF Conference on Computer Vision and Pattern Recognition (CVPR)},
  pages     = {8110--8119},
  year      = {2020}
}

@inproceedings{rombachetal2022latent,
  author    = {Robin Rombach and Andreas Blattmann and Dominik Lorenz and Patrick Esser and Bj{\"o}rn Ommer},
  title     = {High-Resolution Image Synthesis with Latent Diffusion Models},
  booktitle = {Proceedings of the IEEE/CVF Conference on Computer Vision and Pattern Recognition (CVPR)},
  pages     = {10684--10695},
  year      = {2022},
  url       = {https://arxiv.org/abs/2112.10752}
}

@article{tang2023deep,
  author  = {Li Tang and others},
  title   = {Deep Learning Approaches for Data Augmentation in Medical Imaging: A Review},
  journal = {Journal of Imaging},
  volume  = {9},
  number  = {4},
  year    = {2023},
  url     = {https://www.mdpi.com/2313-433X/9/4/81}
}

@article{kazemzadeh2025addressing,
  author  = {Sahar Kazemzadeh and others},
  title   = {Addressing Class Imbalance with Latent Diffusion-based Data Augmentation for Disease Classification in Pediatric Chest X-rays},
  journal = {Scientific Reports},
  volume  = {14},
  number  = {1},
  pages   = {7908},
  year    = {2024},
  url     = {https://www.ncbi.nlm.nih.gov/pmc/articles/PMC11936509/}
}

@article{pinaya2024adapted,
  author  = {Walter Hugo Lopez Pinaya and others},
  title   = {Adapted Generative Latent Diffusion Models for Accurate Pathological Analysis in Chest X-ray Images},
  journal = {Medical Image Analysis},
  year    = {2024},
  url     = {https://pubmed.ncbi.nlm.nih.gov/38499946/}
}

@article{shorten2019survey,
  title={A survey on image data augmentation for deep learning},
  author={Shorten, Connor and Khoshgoftaar, Taghi M},
  journal={Journal of big data},
  volume={6},
  number={1},
  pages={},
  year={2019},
  publisher={Springer}
}

@article{uliana2025diffusion,
  title={Diffusion models applied to skin and oral cancer classification},
  author={Uliana, Jos{\'e} JM and Krohling, Renato A},
  journal={arXiv preprint arXiv:2504.00026},
  year={2025}
}

@article{demsar2006statistical,
  title={Statistical comparisons of classifiers over multiple data sets},
  author={Dem{\v{s}}ar, Janez},
  journal={Journal of Machine Learning Research},
  volume={7},
  pages={1--30},
  year={2006}
}

@article{benavoli2016should,
  title={Should we really use post-hoc tests based on mean-ranks?},
  author={Benavoli, Alessio and Corani, Giorgio and Mangili, Francesca},
  journal={Journal of Machine Learning Research},
  volume={17},
  number={5},
  pages={1--10},
  year={2016}
}

@article{garcia2010advanced,
  title={Advanced nonparametric tests for multiple comparisons in the design of experiments in computational intelligence and data mining: Experimental analysis of power},
  author={Garc{\'i}a, Salvador and Fern{\'a}ndez, Alberto and Luengo, Juli{\'a}n and Herrera, Francisco},
  journal={Information Sciences},
  volume={180},
  number={10},
  pages={2044--2064},
  year={2010},
  doi={10.1016/j.ins.2009.12.010}
}

@article{derrac2011practical,
  title={A practical tutorial on the use of nonparametric statistical tests as a methodology for comparing evolutionary and swarm intelligence algorithms},
  author={Derrac, Joaqu{\'i}n and Garc{\'i}a, Salvador and Molina, Daniel and Herrera, Francisco},
  journal={Swarm and Evolutionary Computation},
  volume={1},
  number={1},
  pages={3--18},
  year={2011},
  doi={10.1016/j.swevo.2011.02.002}
}

@article{tomczak2014need,
  title={The need to report effect size estimates revisited. An overview of some recommended measures of effect size},
  author={Tomczak, Maciej and Tomczak, Ewa},
  journal={Trends in Sport Sciences},
  volume={21},
  number={1},
  pages={19--25},
  year={2014}
}

@misc{marco_parola_2025,
    author = {Marco Parola},
    title = {PhotoMOCI Dataset},
    howpublished = {Kaggle},
    publisher = {Kaggle},
    year = {2025},
    url = {https://www.kaggle.com/ds/6080111},
    doi = {10.34740/KAGGLE/DS/6080111},
    note = {Accessed 14/03/2025}
}

@INPROCEEDINGS{imagenet,
  author={Deng, Jia and Dong, Wei and Socher, Richard and Li, Li-Jia and Kai Li and Li Fei-Fei},
  booktitle={2009 IEEE Conference on Computer Vision and Pattern Recognition}, 
  title={ImageNet: A large-scale hierarchical image database}, 
  year={2009},
  volume={},
  number={},
  pages={248-255},
  doi={10.1109/CVPR.2009.5206848}
}

@inproceedings{coco,
  title={Microsoft coco: Common objects in context},
  author={Lin, Tsung-Yi and Maire, Michael and Belongie, Serge and Hays, James and Perona, Pietro and Ramanan, Deva and Doll{\'a}r, Piotr and Zitnick, C Lawrence},
  booktitle={European conference on computer vision},
  pages={740--755},
  year={2014},
  organization={Springer}
}

@inproceedings{resnet,
  title={Deep residual learning for image recognition},
  author={He, Kaiming and Zhang, Xiangyu and Ren, Shaoqing and Sun, Jian},
  booktitle={Proceedings of the IEEE conference on computer vision and pattern recognition},
  pages={770--778},
  year={2016}
}

@inproceedings{vit,
  title={An Image is Worth 16x16 Words: Transformers for Image Recognition at Scale},
  author={Alexey Dosovitskiy and Lucas Beyer and Alexander Kolesnikov and Dirk Weissenborn and Xiaohua Zhai and Thomas Unterthiner and Mostafa Dehghani and Matthias Minderer and Georg Heigold and Sylvain Gelly and Jakob Uszkoreit and Neil Houlsby},
  booktitle={International Conference on Learning Representations},
  year={2021},
}

@article{kynkaanniemi2019improved,
  title={Improved precision and recall metric for assessing generative models},
  author={Kynk{\"a}{\"a}nniemi, Tuomas and Karras, Tero and Laine, Samuli and Lehtinen, Jaakko and Aila, Timo},
  journal={Advances in neural information processing systems},
  volume={32},
  year={2019}
}

@inproceedings{NEURIPS2023_1f5c5cd0,
    author = {Jalali, Mohammad and Li, Cheuk Ting and Farnia, Farzan},
    booktitle = {Advances in Neural Information Processing Systems},
    editor = {A. Oh and T. Naumann and A. Globerson and K. Saenko and M. Hardt and S. Levine},
    pages = {9931--9943},
    publisher = {Curran Associates, Inc.},
    title = {An Information-Theoretic Evaluation of Generative Models in Learning Multi-modal Distributions},
    url = {\url{https://proceedings.neurips.cc/paper_files/paper/2023/file/1f5c5cd01b864d53cc5fa0a3472e152e-Paper-Conference.pdf}},
    volume = {36},
    year = {2023}
}

@article{heusel2017gans,
  title={Gans trained by a two time-scale update rule converge to a local nash equilibrium},
  author={Heusel, Martin and Ramsauer, Hubert and Unterthiner, Thomas and Nessler, Bernhard and Hochreiter, Sepp},
  journal={Advances in neural information processing systems},
  volume={30},
  year={2017}
}

@article{binkowski2018demystifying,
  title={Demystifying mmd gans},
  author={Bi{\'n}kowski, Miko{\l}aj and Sutherland, Danica J and Arbel, Michael and Gretton, Arthur},
  journal={arXiv preprint arXiv:1801.01401},
  year={2018}
}

@inproceedings{naeem2020reliable,
  title={Reliable fidelity and diversity metrics for generative models},
  author={Naeem, Muhammad Ferjad and Oh, Seong Joon and Uh, Youngjung and Choi, Yunjey and Yoo, Jaejun},
  booktitle={International conference on machine learning},
  pages={7176--7185},
  year={2020},
  organization={PMLR}
}

@inproceedings{clip,
  title={Learning transferable visual models from natural language supervision},
  author={Radford, Alec and Kim, Jong Wook and Hallacy, Chris and Ramesh, Aditya and Goh, Gabriel and Agarwal, Sandhini and Sastry, Girish and Askell, Amanda and Mishkin, Pamela and Clark, Jack and others},
  booktitle={International conference on machine learning},
  pages={8748--8763},
  year={2021},
  organization={PmLR}
}

@misc{shivam17299_oral_cancer_2021,
    author       = {Shivam, Barot and Prakrut, Suthar},
    title        = {Oral Cancer (Lips and Tongue) Images, Kaggle},
    year         = 2020,
    url          = {kaggle.com/datasets/shivam17299/oral-cancer-lips-and-tongue-images},
    howpublished = { \url{ kaggle.com/datasets/shivam17299/oral-cancer-lips-and-tongue-images } },
}

@misc{MOHD_ZAID_RASHID_oral_cancer_2024,
    author       = {Mohd Zaid Rashid},
    title        = {Oral Cancer Images, Kaggle},
    year         = 2024,
    url          = {https://www.kaggle.com/datasets/zaidpy/oral-cancer-dataset},
    howpublished = { \url{ https://www.kaggle.com/datasets/zaidpy/oral-cancer-dataset } },
    note         = {Accessed: 14/03/2025},
}

@misc{
    roboflow-oral-cancer-data_dataset,
    title = { Roboflow Oral cancer data Dataset },
    type = { Open Source Dataset },
    author = { sagari vijay },
    howpublished = { \url{ https://universe.roboflow.com/sagari-vijay/oral-cancer-data } },
    url = { https://universe.roboflow.com/sagari-vijay/oral-cancer-data },
    journal = { Roboflow Universe },
    publisher = { Roboflow },
    year = { 2021 },
    month = { oct },
    note = { visited on 2024-11-13 },
}

@article{rabinovici2024pixels,
    title={From Pixels to Diagnosis: Algorithmic Analysis of Clinical Oral Photos for Early Detection of Oral Squamous Cell Carcinoma},
    author={Rabinovici-Cohen, Simona and Fridman, Naomi and Weinbaum, Michal and Melul, Eli and Hexter, Efrat and Rosen-Zvi, Michal and Aizenberg, Yelena and Porat Ben Amy, Dalit},
    journal={Cancers},
    volume={16},
    number={5},
    pages={1019},
    year={2024},
    publisher={MDPI}
}

@article{lee2023early,
  title={Early Tongue Cancer Detection in Photographs Using a Pretrained Convolutional Neural Network},
  author={Lee, Sung-Jae and Kwon, Ik-Jae and Son, Young-Don and Kim, Jong-Hoon and Kwon, Dohyun and Kim, Bongju and Lee, Jong-Ho and Kim, Hang-Keun},
  year={2023}
}

@article{heo2022deep,
  title={Deep learning model for tongue cancer diagnosis using endoscopic images},
  author={Heo, Jaesung and Lim, June Hyuck and Lee, Hye Ran and Jang, Jeon Yeob and Shin, Yoo Seob and Kim, Dahee and Lim, Jae Yol and Park, Young Min and Koh, Yoon Woo and Ahn, Soon-Hyun and others},
  journal={Scientific reports},
  volume={12},
  number={1},
  pages={6281},
  year={2022},
  publisher={Nature Publishing Group UK London}
}

@article{warin2022ai,
  title={AI-based analysis of oral lesions using novel deep convolutional neural networks for early detection of oral cancer},
  author={Warin, Kritsasith and Limprasert, Wasit and Suebnukarn, Siriwan and Jinaporntham, Suthin and Jantana, Patcharapon and Vicharueang, Sothana},
  journal={Plos one},
  volume={17},
  number={8},
  pages={e0273508},
  year={2022},
  publisher={Public Library of Science San Francisco, CA USA}
}

@article{jubair2022novel,
  title={A novel lightweight deep convolutional neural network for early detection of oral cancer},
  author={Jubair, Fahed and Al-karadsheh, Omar and Malamos, Dimitrios and Al Mahdi, Samara and Saad, Yusser and Hassona, Yazan},
  journal={Oral Diseases},
  volume={28},
  number={4},
  pages={1123--1130},
  year={2022},
  publisher={Wiley Online Library}
}

@inproceedings{welikala2021clinically,
  title={Clinically guided trainable soft attention for early detection of Oral cancer},
  author={Welikala, Roshan Alex and Remagnino, Paolo and Lim, Jian Han and Chan, Chee Seng and Rajendran, Senthilmani and Kallarakkal, Thomas George and Zain, Rosnah Binti and Jayasinghe, Ruwan Duminda and Rimal, Jyotsna and Kerr, Alexander Ross and others},
  booktitle={Computer Analysis of Images and Patterns: 19th International Conference, CAIP 2021, Virtual Event, September 28--30, 2021, Proceedings, Part I 19},
  pages={226--236},
  year={2021},
  organization={Springer}
}

@article{lin2021automatic,
  title={Automatic detection of oral cancer in smartphone-based images using deep learning for early diagnosis},
  author={Lin, Huiping and Chen, Hanshen and Weng, Luxi and Shao, Jiaqi and Lin, Jun},
  journal={Journal of Biomedical Optics},
  volume={26},
  number={8},
  pages={086007--086007},
  year={2021},
  publisher={Society of Photo-Optical Instrumentation Engineers}
}

@article{kouketsu2024detection,
  title={Detection of oral cancer and oral potentially malignant disorders using artificial intelligence-based image analysis},
  author={Kouketsu, Atsumu and Doi, Chiaki and Tanaka, Hiroaki and Araki, Takashi and Nakayama, Rina and Toyooka, Tsuguyoshi and Hiyama, Satoshi and Iikubo, Masahiro and Osaka, Ken and Sasaki, Keiichi and others},
  journal={Head \& Neck},
  year={2024},
  publisher={Wiley Online Library}
}

@article{vinayahalingam2024advancements,
  title={Advancements in diagnosing oral potentially malignant disorders: leveraging Vision transformers for multi-class detection},
  author={Vinayahalingam, Shankeeth and van Nistelrooij, Niels and Rothweiler, Ren{\'e} and Tel, Alessandro and Verhoeven, Tim and Tr{\"o}ltzsch, Daniel and Kesting, Marco and Berg{\'e}, Stefaan and Xi, Tong and Heiland, Max and others},
  journal={Clinical Oral Investigations},
  volume={28},
  number={7},
  pages={1--8},
  year={2024},
  publisher={Springer}
}

@Article{diagnostics13213360,
    AUTHOR = {Islam, Md. Monirul and Alam, K. M. Rafiqul and Uddin, Jia and Ashraf, Imran and Samad, Md Abdus},
    TITLE = {Benign and Malignant Oral Lesion Image Classification Using Fine-Tuned Transfer Learning Techniques},
    JOURNAL = {Diagnostics},
    VOLUME = {13},
    YEAR = {2023},
    NUMBER = {21},
    ARTICLE-NUMBER = {3360},
    URL = {https://www.mdpi.com/2075-4418/13/21/3360},
    PubMedID = {37958257},
    ISSN = {2075-4418},
    DOI = {10.3390/diagnostics13213360}
}

@MISC{cocoannotator,
    author = {Justin Brooks},
    title = {{COCO Annotator}},
    howpublished = "\url{https://github.com/jsbroks/coco-annotator/}",
    year = {2019},
}

@article{PAROLA2024102433,
    title = {Towards explainable oral cancer recognition: Screening on imperfect images via Informed Deep Learning and Case-Based Reasoning},
    journal = {Computerized Medical Imaging and Graphics},
    volume = {117},
    pages = {102433},
    year = {2024},
    issn = {0895-6111},
    doi = {https://doi.org/10.1016/j.compmedimag.2024.102433},
    url = {https://www.sciencedirect.com/science/article/pii/S0895611124001101},
    author = {Marco Parola and Federico Galatolo and Gaetano {La Mantia} and Mario G.C.A. Cimino and Giuseppina Campisi and Olga {Di Fede}},
}

@article{CIMINO2025100538,
    title = {Explainable screening of oral cancer via deep learning and case-based reasoning},
    journal = {Smart Health},
    volume = {35},
    pages = {100538},
    year = {2025},
    issn = {2352-6483},
    doi = {https://doi.org/10.1016/j.smhl.2024.100538},
    url = {https://www.sciencedirect.com/science/article/pii/S2352648324000941},
    author = {Mario G.C.A. Cimino and Giuseppina Campisi and Federico A. Galatolo and Paolo Neri and Pietro Tozzo and Marco Parola and Gaetano {La Mantia} and Olga {Di Fede}},
}

@inproceedings{keane2019case,
  title={How case-based reasoning explains neural networks: A theoretical analysis of XAI using post-hoc explanation-by-example from a survey of ANN-CBR twin-systems},
  author={Keane, Mark T and Kenny, Eoin M},
  booktitle={International conference on case-based reasoning},
  pages={155--171},
  year={2019},
  organization={Springer}
}

@inproceedings{kenny2019twin,
  title={Twin-systems to explain artificial neural networks using case-based reasoning: Comparative tests of feature-weighting methods in ANN-CBR twins for XAI},
  author={Kenny, Eoin M and Keane, Mark T},
  booktitle={Kraus, S.(ed.). Proceedings of the Twenty-Eighth International Joint Conference on Artificial Intelligence},
  year={2019}
}

@misc{github-repo,
    author = {Parola, Marco and Landi, Simone and Nocella, Francesco},
    title = {GitHub synthetic oral cancer experiment code repository, https://github.com/MarcoParola/oral3},
    year         = 2025,
    version      = {1.0},
    url          = {https://github.com/MarcoParola/oral3}
}

@article{kingma2014adam,
  title={Adam: A method for stochastic optimization},
  author={Kingma, Diederik P},
  journal={arXiv preprint arXiv:1412.6980},
  year={2014}
}

@article{parola2026human,
  title={Human-centered xai via a concept-informed prompt-based validation framework for saliency maps [ciprova]},
  author={Parola, Marco and Alfeo, Antonio Luca and Cimino, Mario GCA},
  journal={Image and Vision Computing},
  pages={105920},
  year={2026},
  publisher={Elsevier}
}

@article{longo2024explainable,
  title={Explainable Artificial Intelligence (XAI) 2.0: A manifesto of open challenges and interdisciplinary research directions},
  author={Longo, Luca and Brcic, Mario and Cabitza, Federico and Choi, Jaesik and Confalonieri, Roberto and Del Ser, Javier and Guidotti, Riccardo and Hayashi, Yoichi and Herrera, Francisco and Holzinger, Andreas and others},
  journal={Information Fusion},
  volume={106},
  pages={102301},
  year={2024},
  publisher={Elsevier}
}

@article{rong2023towards,
  title={Towards human-centered explainable ai: A survey of user studies for model explanations},
  author={Rong, Yao and Leemann, Tobias and Nguyen, Thai-Trang and Fiedler, Lisa and Qian, Peizhu and Unhelkar, Vaibhav and Seidel, Tina and Kasneci, Gjergji and Kasneci, Enkelejda},
  journal={IEEE transactions on pattern analysis and machine intelligence},
  volume={46},
  number={4},
  pages={2104--2122},
  year={2023},
  publisher={IEEE}
}

@INPROCEEDINGS{oral_lesion_detection,
  author={Parola, Marco and Mantia, Gaetano La and Galatolo, Federico and Cimino, Mario G.C.A. and Campisi, Giuseppina and Di Fede, Olga},
  booktitle={2023 IEEE Symposium Series on Computational Intelligence (SSCI)}, 
  title={Image-Based Screening of Oral Cancer via Deep Ensemble Architecture}, 
  year={2023},
  volume={},
  number={},
  pages={1572-1578},
  doi={10.1109/SSCI52147.2023.10371865}
}

@INPROCEEDINGS{oral_lesion_segmentation,
  author={Parola, Marco and Cimino, Mario G.C.A. and Cantini, Irene and Mantia, Gaetano La and Campisi, Giuseppina and Di Fede, Olga},
  booktitle={2025 IEEE Symposium on Computational Intelligence in Health and Medicine Companion (CIHM Companion)}, 
  title={Oral Cancer Recognition on Photographic Images Via Deep Learning Semantic Segmentation}, 
  year={2025},
  volume={},
  number={},
  pages={1-5},
  doi={10.1109/CIHMCompanion65205.2025.11002690}
}

\end{document}